\documentclass{article} % For LaTeX2e
\usepackage{iclr2023_conference,times}
\usepackage{amsmath,amsfonts,bm}

\def\eqref#1{equation~\ref{#1}}
\def\1{\bm{1}}

\def\vb{{\bm{b}}}

\def\ve{{\bm{e}}}

\def\vh{{\bm{h}}}

\def\vq{{\bm{q}}}

\def\vs{{\bm{s}}}

\def\vx{{\bm{x}}}
\def\vy{{\bm{y}}}
\def\vz{{\bm{z}}}

\def\mA{{\bm{A}}}
\def\mB{{\bm{B}}}

\def\mD{{\bm{D}}}
\def\mE{{\bm{E}}}

\def\mG{{\bm{G}}}
\def\mH{{\bm{H}}}

\def\mS{{\bm{S}}}

\def\mW{{\bm{W}}}
\def\mX{{\bm{X}}}
\def\mY{{\bm{Y}}}
\def\mZ{{\bm{Z}}}

\DeclareMathAlphabet{\mathsfit}{\encodingdefault}{\sfdefault}{m}{sl}
\SetMathAlphabet{\mathsfit}{bold}{\encodingdefault}{\sfdefault}{bx}{n}

\def\gG{{\mathcal{G}}}

\usepackage{caption}
\usepackage{hyperref}
\usepackage{url}
\usepackage{graphicx}  
\usepackage{epstopdf}
\usepackage{subfigure}
\graphicspath{{figures/}}

\usepackage{amsmath}
\usepackage{cases}
\usepackage{algorithm}
\usepackage{hyperref}
\usepackage{algorithmic}
\usepackage{lineno}
\usepackage{booktabs}
\usepackage{multirow}
\usepackage{multicol}
\usepackage{float}
\usepackage{soul}
\soulregister\cite7 % 针对\cite命令
\soulregister\citep7 % 针对\citep命令
\soulregister\citet7 % 针对\citet命令
\soulregister\ref7 % 针对\ref命令
\soulregister\pageref7 % 针对\pageref命令
\usepackage{xcolor}

\newfloat{figtab}{htb}{fgtb}
\makeatletter
  \newcommand\figcaption{\def\@captype{figure}\caption}
  \newcommand\tabcaption{\def\@captype{table}\caption}
\makeatother

\title{$\rm A^2Q$: Aggregation-Aware Quantization for \\ Graph Neural Networks}

\author{Zeyu Zhu$^{1,2}$ \quad Fanrong Li$^2$ \quad Zitao Mo$^2$ \quad Qinghao Hu$^2$ \quad
Gang Li$^3$ \quad Zejian Liu$^2$ \\ \textbf{Xiaoyao Liang$^3$ \quad Jian Cheng$^2$\thanks{Corresponding author}}\\
$^1$School of Future Technology, University of Chinese Academy of Sciences\\
$^2$Institute of Automation, Chinese Academy of Sciences\\
$^3$Shanghai Jiao Tong University\\
\texttt{\{zhuzeyu2021, lifanrong2017, mozitao2017\}@ia.ac.cn},
\\
\texttt{\{huqinghao2014, liuzejian2018\}@ia.ac.cn}, \texttt{\{gliaca\}@sjtu.edu.cn}\\
\texttt{\{liang-xy\}@cs.sjtu.edu.cn}\\
\texttt{\{jcheng\}@nlpr.ia.ac.cn}
}

\iclrfinalcopy
\begin{document}
% \linenumbers
\maketitle

\begin{abstract}
   % As graph data size increases, the latency and memory usage in the inference process pose 
   % a significant challenge to the deployment of Graph Neural Networks (GNNs).
   % Quantization is a promising 
   % method for efficient neural network inference. However, many previous quantization methods
   % for GNNs suffer from severe accuracy degradation due to the underutilization 
   % of the characteristic of GNNs.
   % In this paper, we analyze the aggregated nodes features and find that 
   % the topology of graph leads to significant difference between nodes. 
   % This motivates us to 
   % propose the aggregation-aware mixed-precision quantization ($\rm A^2Q$) method which 
   % learns appropriate quantization parameters for different nodes. 
   % Moreover, considering the gradients for features are almost zero in semi-supervised tasks
   % due to the sparse connections 
   % between nodes, 
   % we propose a Local Gradient method to introduce quantization error as supervised information
   % to train the quantization parameters.
   % Finally,  
   % we develop a Nearest Neighbor Strategy to generalize to unseen graphs, 
   % which assigns quantization parameters to unseen nodes. 
   % Extensive experiments on eight public node-level and graph-level datasets 
   % demonstrate the generality 
   % and robustness of our proposed method. 
   % Compared to the state-of-the-art quantization method, 
   % our method can obtain up to \textbf{11.4\%} and \textbf{8.7\%} 
   % accuracy gains on the node-level and graph-level tasks, respectively, 
   % while simultaneously achieving up to a \textbf{2.00x} speedup in our hardware simulator.
   As graph data size increases, the vast latency and memory consumption during inference pose a 
   significant challenge to the real-world deployment of Graph Neural Networks (GNNs). 
   While quantization is a powerful approach to reducing GNNs complexity, 
   most previous works on GNNs quantization
   fail to exploit the unique characteristics of GNNs, suffering from severe accuracy degradation. 
   Through an in-depth analysis of the topology of GNNs, we observe that the 
   topology of the graph leads to significant differences between nodes, and most of the nodes in a graph appear to 
   have a small aggregation value. Motivated by this, in this paper, we propose the Aggregation-Aware 
   mixed-precision Quantization ($\rm A^2Q$) for GNNs, where an appropriate bitwidth is automatically 
   learned and assigned to each node in the graph. To mitigate the vanishing gradient problem 
   caused by sparse connections between nodes, we propose a Local Gradient method to serve the 
   quantization error of the node features as the supervision during training. We also develop a 
   Nearest Neighbor Strategy to deal with the generalization on unseen graphs. Extensive 
   experiments on eight public node-level and graph-level datasets demonstrate the generality and 
   robustness of our proposed method. 
   Compared to the FP32 models, our method can achieve up to a 
   18.6x (i.e., 1.70bit) compression ratio with negligible accuracy degradation.
   Morever, compared to the state-of-the-art quantization method, our 
   method can achieve up to 11.4\% and 9.5\% accuracy improvements on the node-level and 
   graph-level tasks, respectively, and up to 2x speedup on a dedicated hardware accelerator.
\end{abstract}
\section{Introduction}
Recently, Graph Neural Networks (GNNs) have attracted much attention 
% from both the industrial and academic
% communities 
due to their superior learning and representing ability for non-Euclidean geometric data. 
A number of GNNs have been widely used in real-world applications, 
such as recommendation system \citep{jin2020multi}, 
% e-commerce analysis \citep{yang2019aligraph}, 
and social network analysis \citep{lerer2019pytorch}, etc. 
% and molecular interactions \citep{de2018molgan}.
Many of these tasks put forward high requirements for low-latency inference. 
However, the real-world graphs are often extremely 
large and irregular, such as Reddit with 232,965 nodes, which needs 19G floating-point operations (FLOPs) 
to be processed by 
a 2-layer Graph Convolutional Network (GCN) 
with only 81KB parameters \citep{tailor2020degree}, while ResNet-50, a 50-layer DNN,
only takes 8G FLOPs to process an image \citep{canziani2016analysis}.
What is worse, it requires a huge amount of memory access for GNNs inference, 
e.g., the nodes features size of Reddit is up to 534MB, leading to high latency. 
% which may cause a higher latency than computation \citep{you2022gcod,sze2017efficient}.
Therefore, the aforementioned problems pose a challenge to realize efficient inference
of GNNs.

Neural network quantization can reduce the model size and accelerate inference 
without modifying the model architecture, 
which has become a promising method to solve this problem in recent years.
Unfortunately, there remain some issues in the existing works on GNNs quantization. 
\citet{feng2020sgquant} only quantizes the node feature and keeps floating point calculations during inference.  
\citet{tailor2020degree} proposes a degree-quant training strategy to 
quantize GNNs to the low-bit fixed point but causes a large accuracy drop, 
e.g., 11.1\% accuracy drops when quantizing to 4bits.
Moreover, some works~\citep{wang2021bi,bahri2021binary,wang2021binarized,jing2021meta} 
quantize GNNs into 1-bit and compute with XNOR and bit count operations.
% \verb+popcount+. 
However, these 1-bit quantization methods are either restricted to the node-level tasks or can not generalize 
well to other GNNs.

\begin{figure}[t]
   \vspace{-0.3cm}
   \subfigure[]{
      \begin{minipage}[t]{0.48\linewidth}
         \centering
         \includegraphics[width=\linewidth,height=0.7\textwidth]{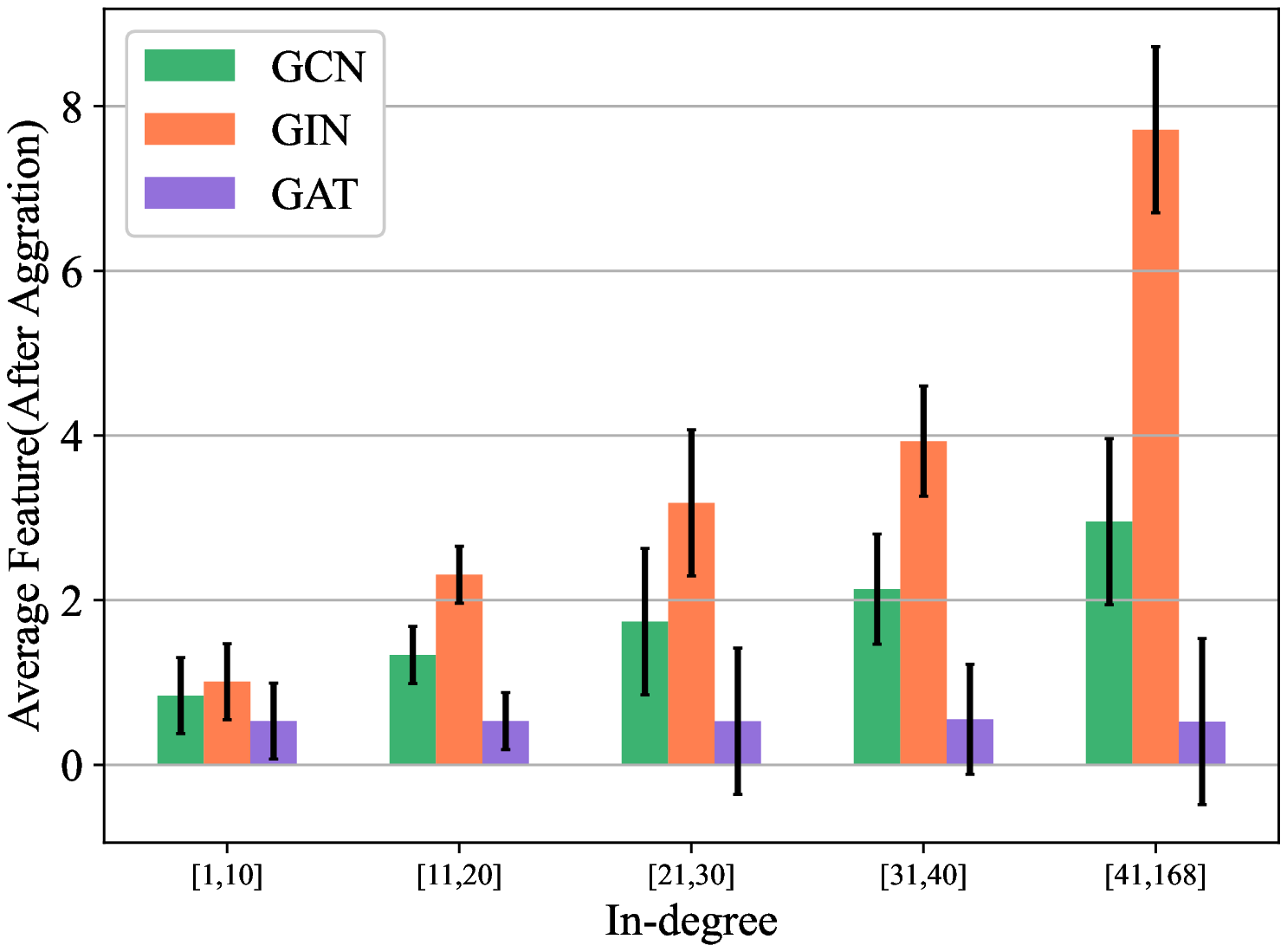}
         \label{nc_fea_var_deg}
      \end{minipage}
   }
   \subfigure[]{
      \begin{minipage}[t]{0.48\linewidth}
         \centering
         \includegraphics[width=\linewidth,height=0.7\linewidth]{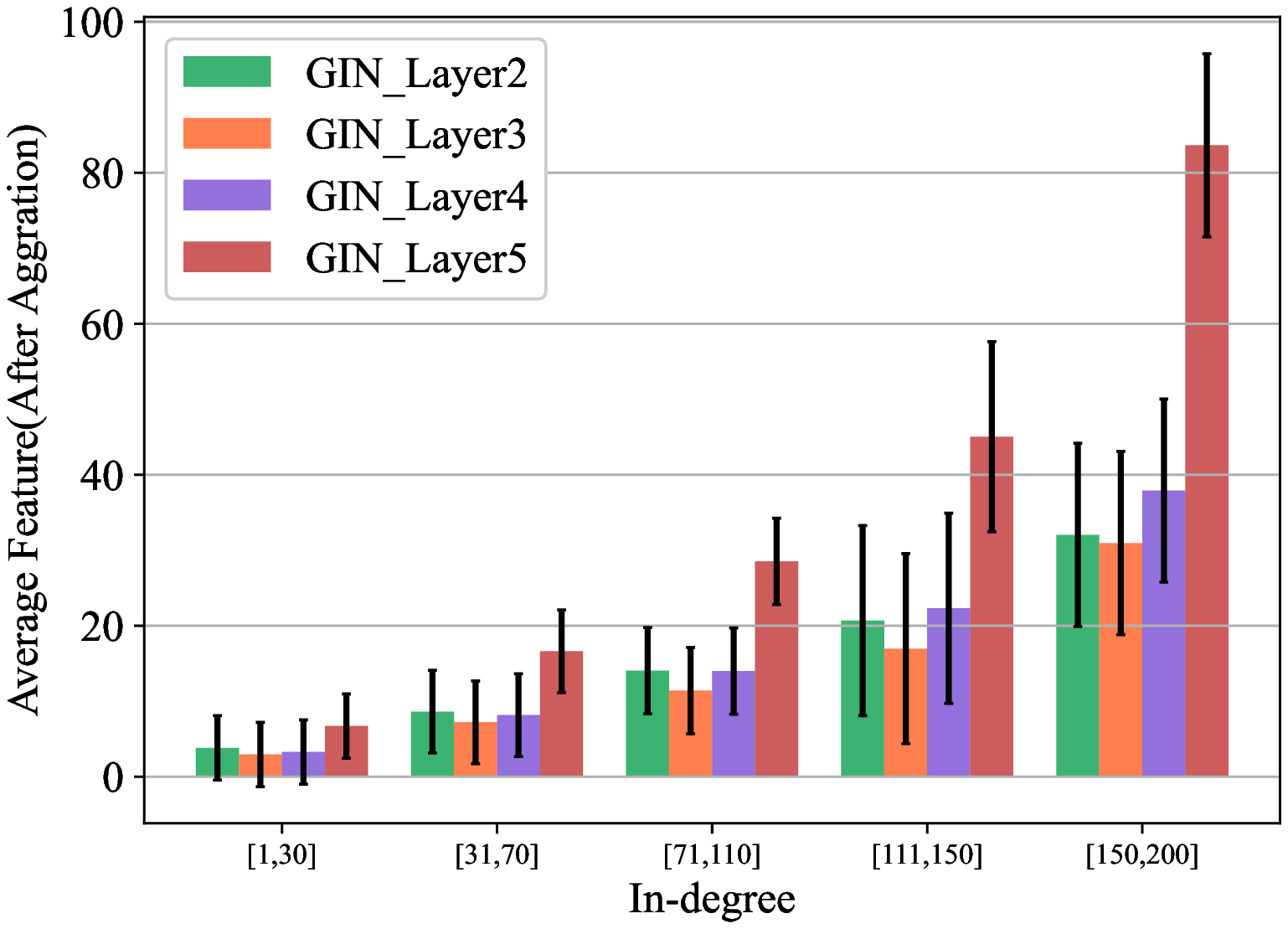}
         \label{gc_fea_var_deg}
      \end{minipage}
   }
   \caption{The analysis of the average aggerated node features 
   in different in-degrees node groups
   on various tasks. (a) The values at the final layer for GNNs trained on Cora. (b) The values at the 2-5 layer 
   of GIN trained on REDDIT-BINARY. The average values are all generated from 10 runs.}
   \label{fea_var_deg}
   % \begin{minipage}[t]{0.5\linewidth}
   %    \centering
   %    \includegraphics[width=1\linewidth,height=0.83\textwidth]{first_Figureeps}
   %    \caption{Comparisons of accuracy on Cora and REDDIT-BINARY for GNNs, e.g. GCN-Cora represent the task of GCN model on Cora dataset.
   %    $\rm{A^2Q-\#bit}$ where \# is the average bit of the node feature in the model quantized by our method.}
   %    \label{result_comparison}
   % \end{minipage}
\end{figure}

Most of the above methods do not make 
full use of the property of GNNs and graph data, resulting in severe accuracy degradation 
or poor generalization.
% Due to the underutilization of the property of GNNs and graph data, the above methods tend to have 
% severe accuracy degradation or poor generalization.
As presented in 
MPNN framework \citep{gilmer2017neural}, 
GNNs processing is divided into two phase: First, in the aggregation phase, 
a node collects information from 
neighboring nodes and uses the aggregation function 
to generate hidden features; second, in the update phase, the hidden features are transformed
into new features by an update function.
We analyze 
the nodes features after aggregation in Figure \ref{fea_var_deg} and find that 
the higher the in-degree is, the larger the node features 
tend to be after aggregation. And the features vary significantly between nodes 
with different in-degrees, which represent the
topology of a graph. 
Moreover, according to \citet{xie2014distributed,aiello2001random}, 
the degrees of nodes in most real-world graph data often follow the 
power-law distribution, i.e., nodes with a low degree account for the majority of graph data. 
% and the features of these nodes do not require high bitwidths to 
% quantize. 
Therefore, specially quantizing the nodes features according to the topology of the graphs will
be beneficial to reduce the quantization error while achieving a higher compression ratio. 
% Moreover, since the number of input nodes varies in graph-level tasks, 
% few works can generalize well on unseen graphs since 
% the number of quantization parameters is fixed.
% However, manually assigning the quantization parameters is challenging due to the large number of 
% nodes in a graph (e.g. 169,343 nodes in ogbn-arxiv), 
% which results in time-consuming process of finding an assignment strategy and the 
% assignment strategy based on prior knowledge is often sub-optimal.
% for the following
% reasons: 1. The assignment strategy is often based on prior knowledge without considering the 
% actual conditions, leading to the strategy being sub-optimal. 2. The combinations of
% quantization parameters explode as the nodes increase, and there are often thousands or 
% even hundreds of thousands
% of nodes in a graph, resulting in the time-consuming process of finding an assignment strategy.  
% it is unreasonable to assign the same bitwidth to all nodes as previous works, 
% leading to excessive memory consumption on nodes with low in-degrees and incur 
% severe quantization error 
% on nodes with high in-degrees. 

% The degree is an essential topology property of the nodes in graph data, 
% In order to further compress GNNs without severe accuracy degradation, we 
% focus on the mixed precision quantization. An intuitive reason is 
% that the nodes retain a different amount of information after aggregation. 
In this paper, 
we propose the \textbf{Aggregation-Aware Quantization} ($\rm A^2Q$) method, 
which quantizes different nodes features with different learnable quantization parameters, 
including bitwidth and step size. These parameters can be adaptively learned during 
training and are constrained by a penalty on memory size to improve the compression ratio.
However, when quantizing the model in 
semi-supervised tasks, the gradients for most quantization parameters are zero due to 
the sparse connections
between nodes, 
% while the labeled nodes only account for a tiny fraction, 
which makes the training non-trivial. We propose the \textbf{Local Gradient} method to solve this problem by 
introducing quantization error as supervised information.
% removing the vanish term. 
Finally, to generalize our method to unseen graphs in which the number of the nodes varies,
we develop the \textbf{Nearest Neighbor Strategy} which  
assigns the learned quantization parameters to the unseen graph nodes. 
To the best of our knowledge, we are the first to introduce the mixed-precision quantization to the GNNs.
Compared with the previous works, our proposed methods can significantly 
compress GNNs with negligible accuracy drop.    
% Due to the sparsity of the connections
% between nodes in a graph and the labeled nodes only account for a small fraction, the gradients for 
% Figure \ref{result_comparison} shows our method is powerful in both accuracy and compression ratio
% compared with the DQ-INT4, which is a state-of-the-art quantization method.

% Based on our analysis and the problems in past studies, the contributions of this paper are:
In summary, the key contributions of this paper are as follows:
\begin{itemize}
   \item [1)] 
   We propose the Aggregation-Aware mixed-precision Quantization ($\rm A^2Q$) 
   method to enable an adaptive learning 
   of quantization parameters. Our learning method is powerful by fully utilizing the characteristic 
   of GNNs, and the learned bitwidth is strongly related to the topology of the graph.
   \item [2)]
   A \textbf{Local Gradient} method is proposed to train the quantization parameters in 
   semi-supervised learning tasks. 
   Furthermore, to generalize our method to the unseen graphs
   in which the number of input nodes is variable, 
   we develop the \textbf{Nearest Neighbor Strategy} to select 
   quantization parameters for the nodes of the unseen graphs.
   \item [3)]
   Experiments demonstrate that we can achieve a compression ratio up to 
   18.6x with negligible accuracy degradation compared to the full-precision 
   (FP32) models. Moreover, the 
   model trained with our $\rm A^2Q$ method
   outperforms the state-of-the-art (SOTA) method up to 11.4\% with a speedup up to 2.00x 
   in semi-supervised tasks, and obtains
   up to 9.5\% gains with a 1.16x speedup in graph-level tasks. We provide our 
   code at this URL: \href{https://github.com/weihai-98/A-2Q}{https://github.com/weihai-98/$\rm{A^2Q}$}.
 \end{itemize}

\section{Related Work}
\label{background}

\qquad\textbf{\textit{Graph Neural Networks:}}
The concept of the graph neural network was first proposed in \citet{scarselli2008graph}, which 
attempted to generalize neural networks to model non-Euclidean data. 
% and in 
In the following years, various GNN models were 
proposed. For example, Graph Convolution Network (GCN) \citep{kipf2016semi} 
uses a layer-wise propagation rule that is based on a first-order 
approximation of spectral convolutions on graphs, Graph Isomorphism Network (GIN) \citep{xu2018powerful}
designed a provably maximally powerful GNN under the MPNN framework, 
and Graph Attention Network (GAT) \citep{velivckovic2017graph} introduces the attention mechanism to 
graph processing.
Although GNNs have encouraging performance in a wide range of domains \citep{jin2020multi,yang2019aligraph},
the huge amount of float-point operations
and memory access in 
process pose a challenge to efficient inference, which hinder the 
applications of GNNs.

% In this paper, we select three typical GNN models that can be represented by the MPNN framework, 
% Graph Convolution Network (GCN) \citep{kipf2016semi}, Graph Isomorphism Network (GIN) \citep{xu2018powerful}, 
% and Graph Attention Network (GAT) \citep{velivckovic2017graph}. 
% The detailed information about these GNNs is  
% shown in Appendix \ref{gnn_formula}.
% The detailed information about them is given in Table~\ref{gnn_formula}. 
% where $d_i$ denotes the degree of node $i$ in GCN, 
% the $\varepsilon $  denotes a learnable constant in GIN, and $\alpha$ represent 
% attention coefficients in GAT.

\qquad\textbf{\textit{Quantized GNNs:}}
% As a promising method to reduce the model size and accelerate the inference process, quantization
% is also applied to GNNs. \citet{feng2020sgquant} proposed a 
% quantization strategy to assign bits for different levels in GNNs.
% In \citet{tailor2020degree}, a degree-quant training
% strategy was proposed to train the quantized GNNs. 
% This method quantizes the GNNs 
% % to 4bits or 8bits 
% by probabilistically generating masks on nodes during training
% and only quantizes nodes that are not masked. 
% % When quantizing GNNs to 8bit, this method maintains the model performance compared to the model at FP32. However, the compression ratio is relatively small. 
% In addition, there are some works that 
% quantize GNNs into 1-bit, i.e., to represent the weights and features by \{+1, -1\} and then compute them
% by \verb+popcount+ \citep{wang2021bi,bahri2021binary,wang2021binarized,jing2021meta}. 
% The all existing
% works do not take full advantage of the characteristics of GNNs
% and graph data, resulting in high accuracy degradation and poor generalization.
As a promising method to reduce the model size and accelerate the inference process, quantization
is also applied to GNNs.
Some works quantize features and weights in GNNs to low bitwidths 
\citep{feng2020sgquant,tailor2020degree}
or even 1-bit \citep{wang2021bi,bahri2021binary,wang2021binarized,jing2021meta}, i.e., 
use fixed-point numbers instead of floating-point numbers for computation. 
% But these works do not achieve a good trade-off between compression ratio and accuracy, 
But when the compression ratio is high (e.g., $<$4bit), the performance degradation of these works 
is significant, and the generalization of 1-bit method is limited. 
There are also some works on vector quantization (VQ), which use the vectors in a codebook obtained 
during the training process 
instead of the original features \citep{ding2021vq,huang2022epquant}. However, 
searching for vectors in the codebook is computationally complex.
% The all existing
% works do not take full advantage of the characteristics of GNNs
% and graph data, resulting in high accuracy degradation or poor generalization.

\qquad\textbf{\textit{Mixed-Precision Quantization:}}
Based on the idea that different layers have different sensitivities to quantization, 
mixed-precision quantization is proposed in CNNs 
to quantize different layers to different bitwidths for better model compression.
Early works \citep{wang2019haq,lou2019autoq} proposed reinforcement learning (RL) based methods to 
search bitwidth 
for different layers, but they often require large computational resources, 
which limits the exploration of the search space. Another important class of mixed-precision method 
is the criteria-based method, they use the specific criteria to represent the quantization sensitivity, 
e.g., \citep{dong2019hawq,dong2020hawq,chen2021towards}quantize different layers 
with different bitwidths based on the trace of the Hessian. 
Recently, there are some other methods to learn the bitwidth during training 
\citep{uhlich2019mixed,esser2019learned,jain2020trained}.
However, due to the huge difference between GNNs and CNNs, 
it is difficult to use these methods on GNNs directly, 
and our $\rm A^2Q$ is the first method to introduce the mixed-precision quantization to GNNs, 
further improving the inference efficiency of GNNs.

% Some works designed the quantization criteria based on the sensitivity or Hessian information, then 
% higher bitwidths will
% be assigned to the more sensitive layer \citep{dong2019hawq}. Other works proposed differentiable approaches
% to learn the optimal quantization parameters \citep{uhlich2019mixed}. Moreover, other works 
% utilize neural architecture search (NAS). For example, \citet{wu2019fbnet} combines 
% all possible quantization parameters into a
% ``stochastic super net'' and approximate the optimal scheme via sampling. 
% These works produce the SOTA results in CNNs.
% However, it is sitll a blank area that  
% introduce the mixed-precision quantization method to further improve the inference efficiency of GNNs.  

\section{Method}
\label{method}
% In Section \ref{aaq},
% we propose an Aggregation-Aware Quantization method to fully utilize the property of GNNs and graph data. 
% However,
% The method does not perform as well as we expect when used in semi-supervised tasks. 
% We analyze the reasons and introduce a 
% Local Gradient method, which solves this problem well in Section \ref{local gradient}. 
% Finally, to generalize our method to the 
% unseen graphs, we develop the 
% Nearest Neighbor Strategy, which can select quantization parameters for unseen graphs in Section \ref{nearest neighbor strategy}.
% Our method is an end-to-end training method that can be easily applied to other GNN models.
In this section, we describe our proposed Aggregation-Aware Quantization 
in detail. Firstly, we present the formulation of the mixed-precision quantization for GNNs, 
which fully utilizes the property
of GNNs and graph data. 
Secondly, we introduce the Local Gradient method to address the gradient vanishing problem during training. 
Finally, we detail the Nearest Neighbor Strategy, which is used for generalizing our 
approach to the unseen graphs.

\subsection{Aggregation-Aware Quantization}
\label{aaq}
We assume a graph data with $N$ nodes and the node features are $F$-dimensional, i.e., the feature map is 
$\mX\in \mathbb{R}^{N\times F}$ and $\vx_i$ is the features of node $i$. 
We use the learnable parameters step size $\alpha_i\in \mathbb{R}_+ $ and 
bitwidth $b_i\in \mathbb{R}_{+} $
to quantize the features of the $i$-th node as:
\begin{gather}
   \bar{\vx}_i=sign(\vx_i) \begin{cases}
      \lfloor \frac{\left| \vx_i \right|}{\alpha_i}+0.5 \rfloor \text{,} & \left| \vx \right|<\alpha_i(2^{[b_i]-1}-1) \\
      \\
      2^{[b_i]-1}-1 \text{,} & \left| \vx_i \right| \geq \alpha_i(2^{[b_i]-1}-1)  
   \label{quantize_1}
   \end{cases}
   \ \text{,} 
   % \vx_q=s\cdot \bar{\vx} \label{quantize_2}
\end{gather}
where $\lfloor \cdot \rfloor $ is the floor function, and $[\cdot ]$ is the round function to 
ensure the bitwidth
used to quantize is an integer.
The learnable parameters are
$\vs_X=(\alpha_1,\alpha_2,...,\alpha_N)$, 
and $\vb_X=(b_1,b_2,...,b_N)$.
Then we can obtain the fixed-point feature map $\bar{\mX}$, 
and the original feature can be represented as $\mX_q=\mS_X\cdot \bar{\mX}$, 
where $\mS_X=diag(\alpha_1,\alpha_2,...,\alpha_N)$. Note that we use $[b]+1$ as the quantization bitwidth
for the features after ReLU because the values are all non-negative.
% After quantizing the feature map, the aggregation phase can be performed by integer operations to reduce 
% the computational overhead. Note that we use $[b]+1$ as the quantization bitwidth
% for the features after ReLU, because the values are all non-negative.
% \vspace{-1cm}
\begin{figure}[t]
   \centering
   \begin{minipage}[t]{0.48\textwidth}
      \centering
      \includegraphics[scale=0.42]{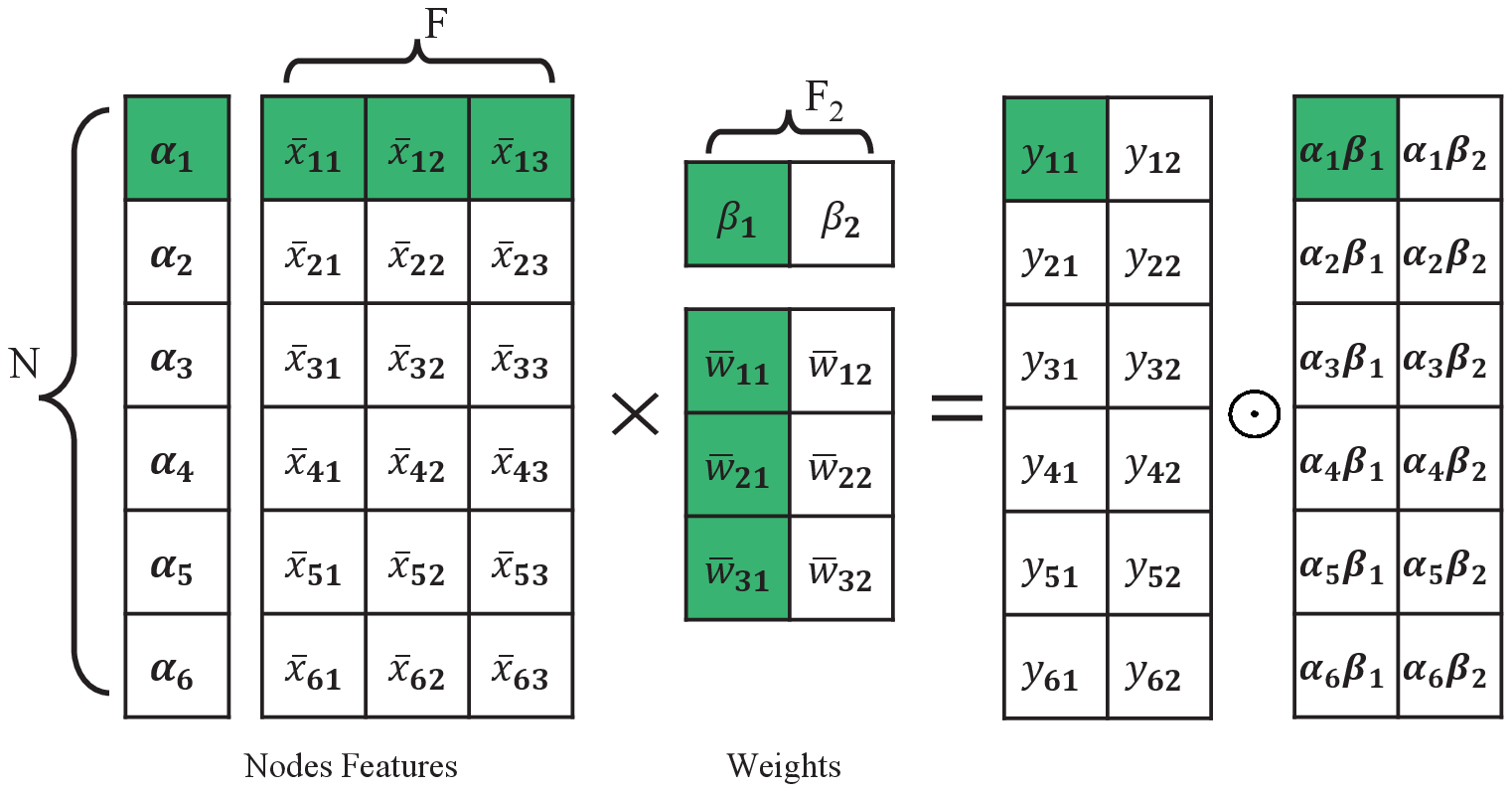}
      \caption{Perform matrix multiplication by the integer represented. $\bar{x}$ and $\bar{w}$ are both integers.}
      \label{quant_mat}
   \end{minipage}
   \quad
   \begin{minipage}[t]{0.48\textwidth}
      \centering
      \includegraphics[scale=0.3]{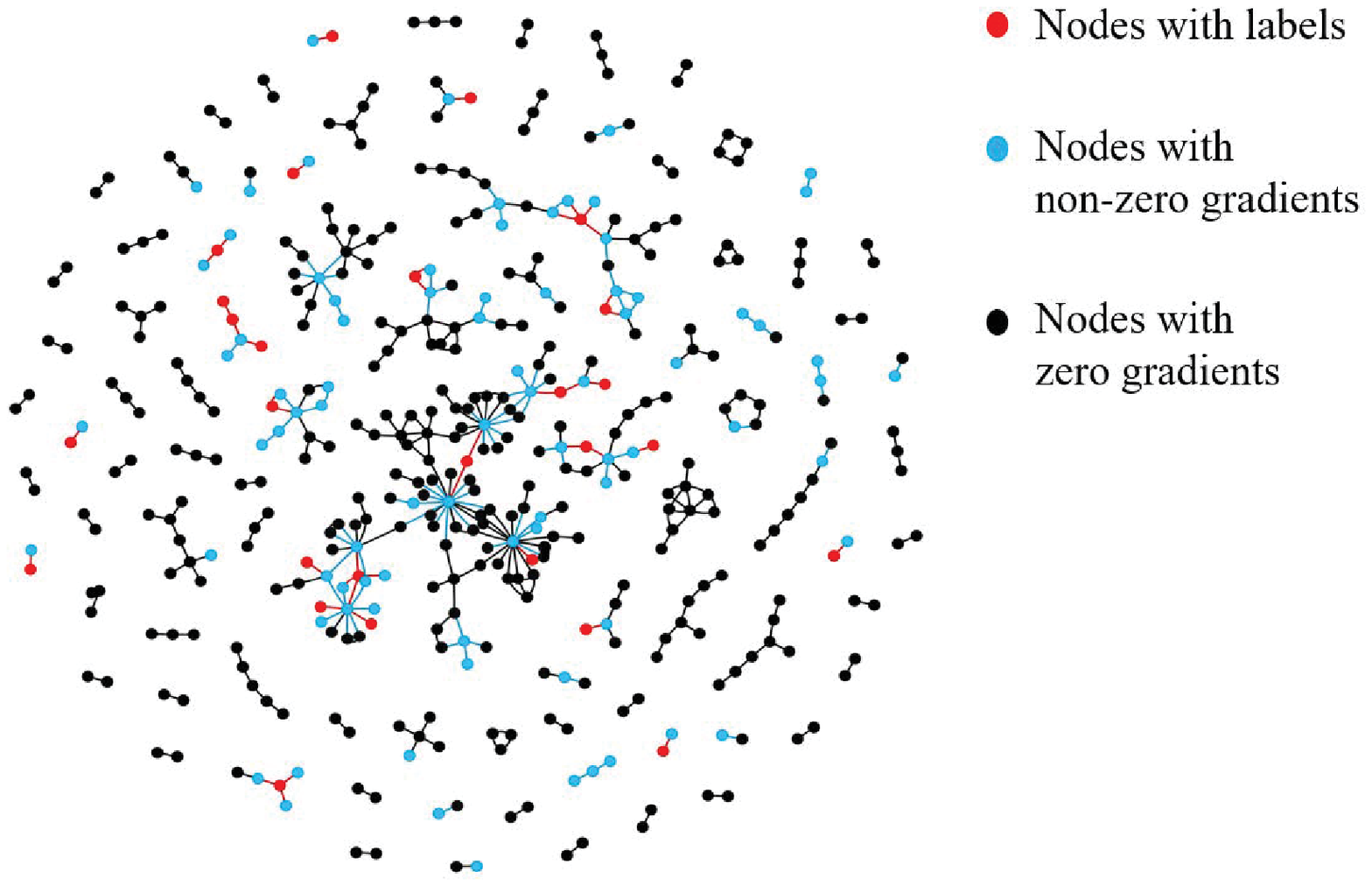}
      \caption{The gradients to $x_q$ in GCN trained on Cora by sampling
      400 nodes.}
      \label{grad_zero}
   \end{minipage}
\end{figure}

In the update phase, the node features are often transformed with a 
linear mapping or an MLP in which matrix multiplication $\mX\mW$ is the main computation, 
and the transformed node features are the input to the next layer in GNNs. 
In order to accelerate the update phase, we also quantize $\mW$. Due to 
the fact that $\mW$ in a certain layer is shared by
all nodes,  
we quantize $\mW$ to the same bitwidth of 4bits for all GNNs in this paper. However, 
each column of $\mW$ has its learnable quantization
step size, i.e.,  $\vs_W=(\beta_1,\beta_2,..,\beta_{F_2})$, where $F_2$ is the 
output-dimension of the node features in current layer 
and $\beta_i$ is the quantization step size for the $i$-th column of $\mW$, 
and we also use Eq. \ref{quantize_1} to
quantize $\mW$.
We can obtain the integer representation $\bar{\mW}$ 
and the quantized representation $\mW_q=\bar{\mW}\cdot \mS_W$,
% \begin{gather}
%    \label{w_quant}
%    \mW_q=\bar{\mW}\cdot \mS_W
% \end{gather}
where $\mS_W=diag(\beta_1,\beta_2,...,\beta_{F_2})$.  
The float-point matrix multiplication in the update phase can be reformulated as follow:
% \begin{minipage}{1\textwidth}
% \vspace{-10pt}
% \setlength{\abovedisplayskip}{-10pt}
\begin{equation}
   % \begin{split}
   \label{mat_mul}
   \mX\cdot \mW\approx \mX_q\cdot \mW_q = (\mS_X\cdot \bar{\mX})\cdot (\bar{\mW}\cdot \mS_W)= (\bar{\mX}\cdot \bar{\mW}) \odot (\vs_X \otimes \vs_W)\ \text{,}
                                       %  &= \mS_X\cdot (\bar{\mX}\cdot \bar{\mW})\cdot \mS_W\\
                                       %  &= (\bar{\mX}\cdot \bar{\mW}) \odot (\vs_X \otimes \vs_W)
   % \end{split}
\end{equation}
% \setlength{\belowdisplayskip}{10cm}
% \end{minipage}
where $\odot$ denotes an element-wise multiplication, and $\otimes$ denotes the outer product. 
After training, we can obtain 
$\vs_X$ and $\vs_W$ so that the outer product can be pre-processed before inference.
An example is illustrated in Figure \ref{quant_mat}. 
For the aggregation phase, i.e., $\mA\mX$, $\mA$ is the adjacency matrix and 
$\mA\in\{0,1\}^{N\times N}$,
we quantize the $\mX$ as the quantization way of $\mW$ because
the nodes features involved in the aggregation process come from the update phase, in which 
the features lose the topology information of graphs. 
Then the aggregation
phase can be performed by integer operations to reduce the computational overhead.

The quantization parameters $(s,b)$ are trained by the backpropagation algorithm.
Since the floor and round 
functions used in the quantization process are not differentiable, 
we use the straight-through estimator \citep{bengio2013estimating}
to approximate the gradient through these functions, and the
gradients of the quantization parameters can be calculated by: 
% \begin{flalign}
%    \label{loss_grad_s}
%    \frac{\partial L}{\partial s}=\sum\limits_{i=1}^{d}{\frac{\partial L}{\partial x_{q}^{i}}\cdot \frac{\partial x_{q}^{i}}{\partial s}}\\
%    \frac{\partial L}{\partial b}=\sum\limits_{i=1}^{d}{\frac{\partial L}{\partial x_{q}^{i}}\cdot \frac{\partial x_{q}^{i}}{\partial b}} \label{loss_grad_b}
% \end{flalign} 

\begin{minipage}{0.48\textwidth}
   \begin{equation}
      \label{loss_grad_s}
      \frac{\partial L}{\partial s}=\sum\limits_{i=1}^{d}{\frac{\partial L}{\partial x_{q}^{i}}\cdot \frac{\partial x_{q}^{i}}{\partial s}} \ \text{,}
   \end{equation}
   \end{minipage}
   \begin{minipage}{0.48\textwidth}
   \begin{equation}
      \label{loss_grad_b}
      \frac{\partial L}{\partial b}=\sum\limits_{i=1}^{d}{\frac{\partial L}{\partial x_{q}^{i}}\cdot \frac{\partial x_{q}^{i}}{\partial b}} \ \text{,}
   \end{equation}
   \end{minipage}
where $d$ is the dimension of the vector $\vx$, $(s,b)$ are the quantization parameters for $\vx$,
% to be quantized by parameters $(s,b)$,  
and $x_{q}^{i}$ is the value of $i$-th dimension in $\vx_q$.
Detailed information about quantization process
and the backpropagation are shown in 
Appendix \ref{more uniform quantization} and \ref{proof2} \textbf{Proof 2 and 3}.
% The gradient of $x_q^{i}$ with respect to $(s,b)$ is in Appendix \ref{more uniform quantization}.
% \begin{gather}
%    \begin{bmatrix}
%       \frac{\partial x_{q}^{i}}{\partial s}\\
%       \\
%       \frac{\partial x_{q}^{i}}{\partial b}
%    \end{bmatrix}=\begin{cases}
%       \begin{bmatrix}
%          \frac{1}{s}\left( x_{q}^{i}-{{x}^{i}} \right)\\
%          0
%       \end{bmatrix} \text{,} & \left| \vx \right|<s(2^{b-1}-1)\\
%       sign(x)\begin{bmatrix}
%          \left( {{2}^{b-1}}-1 \right)\\
%          {{2}^{b-1}}\ln \left( 2 \right)s
%       \end{bmatrix} \text{,} & \left| \vx \right| \geq s(2^{b-1}-1)
%    \end{cases}
%    \ \text{.}
% \end{gather}

In order to improve the compression ratio of the node features, 
we introduce a penalty term on the memory size:
\begin{equation}
   {{L}_{memory}}={{( \frac{1}{\eta }\cdot \sum\limits_{l=1}^{L}{\sum\limits_{i=1}^{N}{\dim^{l}\cdot b_{i}^{l}}}-{{M}_{target}})^{2}}} \ \text{,}
\end{equation}
where $L$ is the number of layers in the GNNs, $N$ is the total number of nodes, $\dim^l$ is the 
length of the node features in $l$-th layer,
$b_i^{l}$ is the quantization bitwidth for node $i$ in $l$-th layer, $M_{target}$ is the target memory size 
on the total node features memory size,
and $\eta = 8*1024$, which is a constant to convert the unit of memory size to $\rm KB$. 
Then the model and quantization parameters can be trained by the loss function:
% \vspace{-1cm}
\begin{equation}
   L_{total}={{L}_{task}}+\lambda \cdot {L}_{memory} \ \text{,}
\end{equation} 
where $L_{task}$ is the task-related loss function and 
$\lambda$ is a penalty factor on $L_{memory}$.
% \vspace{-1cm}

\subsection{Local Gradient}
\label{local gradient}
Although the above end-to-end learning method is concise and straightforward,
the gradients for the quantization parameters of nodes features, i.e.,
$\frac{\partial {L_{task}} }{\partial s}$ and 
$\frac{\partial {L_{task}} }{\partial b}$, are almost zero 
during the training process of semi-supervised tasks, which poses a significant challenge
to train the quantization parameters for nodes features. 
We analyze the property of GNNs and graph data, and find that
two reasons lead to this phenomenon: 1. The extreme sparsity of the connections between nodes in graph data.
2. Only a tiny fraction of nodes with labels are used for training in semi-supervised tasks 
(e.g., $0.30\%$ in PubMed
dataset).
Therefore, $\frac{\partial {L_{task}} }{\partial {x_{q}}}$ for most node features are zero 
(detailed proof in Appendix \ref{proof}), which 
results in that the gradients for quantization parameters of these nodes vanish according 
to Eq. \ref{loss_grad_s} and Eq. \ref{loss_grad_b}. To clarify, we visualize
the $\frac{\partial {L_{task}} }{\partial {x_{q}}}$ in the second layer of GCN trained on Cora. 
As shown in Figure \ref{grad_zero}, most gradients for the nodes features are 
zero.
% Therefore, we propose the \textbf{\textit{Local Gradient}} method to solve this problem.

The gradients of the $L_{task}$ w.r.t. quantized nodes features can be viewed as the 
supervised information from the labeled nodes 
which 
enable the training of the quantization parameters for nodes features. However, 
this supervised information is missing due to zero gradients.
Considering the quantization error is related to the $L_{task}$, 
% Suppose $\vx$ is the feature before quantization and $\vx_q$ is the features after quantization.
% Therefore, 
we introduce
the quantization error $E =\frac{1}{d}\left| {{\vx}_{q}}-\vx \right|_1$ 
as the supervised information for the quantization parameters of nodes features,
where $\vx$ is the features before quantization, $\vx_q$ is the features after quantization
and $|\cdot |_1$ denotes the L1 norm.
We refer to this method as \textbf{Local Gradient} because the gradients 
are computed by the local quantization errors instead of back-propagated task-related gradients.
Then the quantization parameters for node features can be trained by gradients
from $E$:
% which is essentially a local loss function of the quantization error for node features. The gradients from $\Delta$ are:
% \begin{gather}
%    \label{quant_loss_s}
%    \frac{\partial \Delta }{\partial s}=\sum\limits_{i=1}^{d}{sign\left( x_{q}^{\left( i \right)}-{{x}^{\left( i \right)}} \right)\cdot \frac{\partial x_{q}^{\left( i \right)}}{\partial s}} \\
%    \frac{\partial \Delta }{\partial b}=\sum\limits_{i=1}^{d}{sign\left( x_{q}^{\left( i \right)}-{{x}^{\left( i \right)}} \right)\cdot \frac{\partial x_{q}^{\left( i \right)}}{\partial b}} \label{quant_loss_b}
% \end{gather}

\begin{minipage}{0.48\textwidth}
   \begin{equation}
      \label{quant_loss_s}
      \frac{\partial E }{\partial s}=\frac{1}{d}\sum\limits_{i=1}^{d}{sign( x_{q}^{i}-{{x}^{i}} )\cdot \frac{\partial x_{q}^{i}}{\partial s}} \ \text{,}
   \end{equation}
   \end{minipage}
   \begin{minipage}{0.48\textwidth}
   \begin{equation}
      \label{quant_loss_b}
      \frac{\partial E }{\partial b}=\frac{1}{d}\sum\limits_{i=1}^{d}{sign( x_{q}^{i}-{{x}^{i}} )\cdot \frac{\partial x_{q}^{i}}{\partial b}} \ \text{.}
   \end{equation}
   \end{minipage}
Note that the quantization parameters of $\mW$ are still trained by utilizing the gradients in Eq. \ref{loss_grad_s}.
\subsection{Nearest Neighbor Strategy}
\label{nearest neighbor strategy}
In graph-level tasks, the quantized GNNs are required to generalize to unseen graphs. 
In such a scenario, 
the number 
of input nodes may vary during training or inference. However, 
the learnable method can only train 
a fixed number of $(s,b)$ pairs which are the same as the number of input nodes, 
so it is challenging to 
learn the $s$ and $b$ for every node in graph-level tasks. 
To solve this problem, we propose the \textbf{Nearest Neighbor Strategy}, 
which 
allows learning of a 
fixed number of quantization parameters and select quantization parameters for the unseen graphs.

% We heuristically argue that the quantization error of the maximum value in a vector tends to contribute 
% the most to the overall quantization error,  
% and therefore minimizing the quantization error of the maximum value in a feature should be first considered
% when selecting the quantization parameters.
The proposed strategy is shown in Algorithm \ref{alg1}.
To ensure the numerical range of $\vx_q$ is as close as to $\vx$ at FP32, 
a simple way is to keep the maximum quantization
value equal to the maximum absolute value of $\vx$.
Based on this idea, we first initialize $m$ groups of quantization parameters, 
then we calculate the maximum quantization value for every group, i.e., 
$q_{max}=s(2^{[b]-1}-1)$. When quantizing the features of node $i$, 
the feature with the largest absolute value $f_i$ in the 
node features $\vx_i$ is  first selected, and then we find the nearest $q_{max}$ and 
quantize the node features with the $(s,b)$ corresponding to this $q_{max}$.
\begin{algorithm}[t]
	%\textsl{}\setstretch{1.8}
	\renewcommand{\algorithmicrequire}{\textbf{Input:}}
	\renewcommand{\algorithmicensure}{\textbf{Output:}}
	\caption{Nearest Neighbor Strategy}
	\label{alg1}
	\begin{algorithmic}[1]
      \STATE \textbf{ForwardPass} ($\mX=(\vx_1,\vx_2,...,\vx_N)^{T}$):
      \STATE \qquad Initialize$(\vs,\vb), \vs\in \mathbb{R}_+^{m\times 1},\vb\in \mathbb{R}_+^{m\times 1} $ before training
      \STATE \qquad Calculate $\vq_{max}=\vs\odot (2^{\vb-1}-1)$
      \STATE \qquad Calculate the maximum absolute value in the features of each node: ${{f}_{i}}=\underset{j}{\mathop{\max }}\,abs( \vx_{i}^{(j)} )$
      \STATE \qquad Search the index of quantization parameters for each node: $inde{{x}_{i}}=\arg \underset{k}{\mathop{\min }}\,| {{f}_{i}}-q_{\max }^{k} |$
      \STATE \qquad Quantize the $i$-th node features using $(s_{index_i},b_{index_i})$
      \STATE \qquad return $\mX_q$
      \STATE \textbf{end}
      % \STATE \textbf{BackwardPass} ($grad_{\bar{\vs}},grad_{\bar{\vb}},index$):
      % \STATE \qquad $grad_s=scatter\_add(grad_{\bar{s}},index)$
      % \STATE \qquad $grad_b=scatter\_add(grad_{\bar{b}},index)$
      % \STATE \qquad return $grad_s,grad_b$
      % \STATE \textbf{end}
   \end{algorithmic}  
\end{algorithm}
When performing backpropagation, we first calculate the gradients of the loss function w.r.t. quantization parameters 
according to Eq. \ref{loss_grad_s} and Eq. \ref{loss_grad_b}. 
For a specific set of quantization parameters $(s_j, b_j)$, we
collect the gradients from the nodes that have used them and 
add these gradients together.
After the model has been trained, we obtain the quantization parameters $(\vs,\vb)$. Since $q_{max}$ can be calculated and 
sorted in advance, searching the nearest $q_{max}$ can be implemented by binary searching. 
% As an illustrative example, 
% the inference time is 122.60ms for GIN trained on REDDIT-BINARY with our 
% Nearest Neighbor Strategy, which only incurs 0.95\% overhead 
% compared with the 121.45ms of the inference without the selection process. 
Usually,
we set $m=1000$ for 
all graph-level tasks in our paper and the overhead introduced to inference time is negligible.
% Therefore, we calculate and
% sort the $q_{max}$ which makes the process of searching the nearest $q_{max}$ a binary search during inference
% to reduce the overhead 
% introduced by searching.
\begin{table}[t]
   % \vspace{-0.2cm}
   \caption{The results comparison on node-level tasks. The average bits are counted for each task when the best 
   results are achieved.}
   \label{node-level-results}
   \begin{center}
   % \addvbuffer[0 -3pt]{
   \begin{tabular}{clllll}
   
   \hline \toprule[2pt]
   Dataset                                & Model                & Accuracy             & Average bits                                            & Compression Ratio                                            & Speedup                                             \\  \midrule[1pt]
   \multirow{6}{*}{\textbf{Cora}}         & GCN(FP32)            & 81.5±0.7\%           & 32                                                      & 1x                                                           & ---                                                  \\
                                          & GCN(DQ  )            & 78.3±1.7\%           & 4                                                       & 8x                                                           & 1x                                                \\
                                          & GCN(ours)            & \textbf{80.9±0.6\%} & \textbf{1.70}                                           & \textbf{18.6x}                                               & \textbf{2.00x}                                       \\ \cline{2-6} 
                                          % & GIN(FP32)            & 77.6±1.1\%           & 32                                                      & 1x                                                           & ---                                                  \\
                                          % & GIN(DQ  )            & 69.9±3.4\%           & 4                                                       & 8x                                                           & 1x                                               \\
                                          % & GIN(ours)            & \textbf{77.8±1.6\%}  & \textbf{2.37}                                           & \textbf{13.4x}                                               & \textbf{1.41x}                                       \\ \cline{2-6} 
                                          & GAT(FP32)            & 83.1±0.4\%           & 32                                                      & 1x                                                           & ---                                                  \\
                                          & GAT(DQ  )            & 71.2±2.9\%           & 4                                                       & 8x                                                           & 1x                                                \\
                                          & GAT(ours)            & \textbf{82.6±0.6\%}  & \textbf{2.03}                                           & \textbf{15.4x}                                               & \textbf{1.49x}                                       \\ \hline
   \multirow{6}{*}{\textbf{CiteSeer}}     & GCN(FP32)            & 71.1±0.7\%           & 32                                                      & 1x                                                           & ---                                                  \\
                                          & GCN(DQ  )            & 66.9±2.4\%           & 4                                                       & 8x                                                           & 1x                                                \\
                                          & GCN(ours)            & \textbf{70.6±1.1\%}  & \textbf{1.87}                                           & \textbf{17.0x}                                               & \textbf{1.91x}                                       \\ \cline{2-6} 
                                          & GIN(FP32)            & 66.1±0.9\%           & 32                                                      & 1x                                                           & ---                                                  \\
                                          & GIN(DQ  )            & 60.8±2.1\%           & 4                                                       & 8x                                                           & 1x                                               \\
                                          & GIN(ours)            & \textbf{65.1±1.7\%}  & \textbf{2.54}                                           & \textbf{12.6x}                                               & \textbf{1.37x}                                       \\ \hline 
                                          % & GAT(FP32)            & 72.5±0.7\%           & 32                                                      & 1x                                                           & ---                                                  \\
                                          % & GAT(DQ  )            & 67.6±1.5\%           & 4                                                       & 8x                                                           & 1x                                                \\
                                          % & GAT(ours)            & \textbf{71.9±0.7\%}  & \textbf{1.94}                                           & \textbf{16.2x}                                               & \textbf{1.45x}                                       \\ \hline
   \multirow{3}{*}{\textbf{PubMed}}       & GAT(FP32)            & 79.0±0.3\%           & 32                                                      & 1x                                                           & --- \\          
                                          & GAT(DQ)              & 70.6±12.5\%           & 4                                                       & 8x                                                           & 1x  \\
                                          & GAT(ours)            & \textbf{78.8±0.4\%}  & \textbf{2.12}                                           & \textbf{15.1x}                                               & \textbf{1.38x} \\ \hline
   \multirow{3}{*}{\textbf{ogbn-arxiv}}   & GCN(FP32)            & 71.7±0.3\%           & 32                                                      & 1x                                                           & --- \\
                                          & GCN(DQ)              & 65.4±3.9\%           & 4                                                       & 8x                                                           & 1x  \\
                                          & GCN(ours)            & \textbf{71.1±0.3\%}  & \textbf{2.65}                                           & \textbf{12.1x}                                               & \textbf{1.28x} \\ \bottomrule[2pt]
   
      \end{tabular}
   \end{center}
   \end{table}
\section{Experiments}
\label{experiments}
\subsection{Experimental Settings}
In this section, we evaluate our method on three typical GNN models, i.e., GCN, GIN, and GAT.
And we compare our method with the FP32 GNN model and DQ-INT4 \citep{tailor2020degree} 
on eight datasets, including four 
node-level semi-learning tasks (Cora, CiteSeer, PubMed, ogbn-arxiv) \citep{hu2020open,yang2016revisiting} 
and four graph-level tasks (REDDIT-BINARY, MNIST, CIFAR10, ZINC) \citep{yanardag2015deep,dwivedi2020benchmarking},
to demonstrate the generality and robustness of 
our method. Among these datasets, 
ZINC is a dataset for regression tasks, 
which uses regression loss as the metric of the model performance,
while others are all for classification tasks.
% In this section, we select the model at FP32 and quantized by DQ-INT4 \citep{tailor2020degree} as baselines to compare with 
% our method $\rm A^2Q$ on GCN, GIN, and GAT. 

For a fair comparison, we set the quantization bitwidth of $\mW$ for all GNNs to 4bits 
as DQ-INT4. 
We count the average bitwidths for nodes features in all layers of the overall model 
and list them in our results, 
denoted by ``Average bits''.
Since today's CPUs and GPUs can not support mixed-precision operations well, 
we implement a precision-scalable hardware accelerator to perform the
overall inference process for GNN.
The accelerator employs massive bit-serial multipliers \cite{judd2016stripes}, therefore, 
the latency of the integer multiplications is determined by the bitwidth of the 
node features. To evaluate the performance gains of our method over DQ-INT4, 
we develop a cycle-accurate simulator for our accelerator. 
More details about accelerator architecture are shown in Appendix \ref{architecture}.
Moreover, we show the compression ratio of quantized GNNs 
compared to the FP32 models in terms of overall memory size.
% , which only considers the node features, 
% because the node features account for more than 90\%
% of the overall memory size in various tasks.
For simplicity, we use GNN(DQ) to represent the GNNs quantized by 
DQ-INT4 and GNN-dataset to represent the task
in which we run the experiment, e.g., GCN-Cora represents the GCN model trained on Cora.  
Detailed information about datasets and settings is in 
Appendix \ref{datsets} and Appendix \ref{experimental setup}. 
% We provide our 
% code in the supplementary material.
\begin{table}[t]
   \caption{The results comparison on graph-level tasks.}
   \label{graph_level_result_1}
   \begin{center}
   \begin{tabular}{clllll}
   \hline \toprule[2pt]
   Dataset                                 & Model          & Accuracy (Loss↓)            & Average bits                                             & Compression ratio                                            & Speedup         \\ \midrule[1pt]
   \multirow{6}{*}{\textbf{MNIST}}         & GCN(FP32) & 90.1±0.2\%                 & 32                                                       & 1x                                                           & ---              \\
                                           & GCN(DQ)   & 84.4±1.3\%                  & 4                                                        & 8x                                                           & 1x          \\
                                           & GCN(ours) & \textbf{89.9±0.8\%}         & \textbf{3.50}                                            & \textbf{9.12x}                                                        & \textbf{1.17x}  \\ \cline{2-6} 
                                           & GIN(FP32) & 96.4±0.4\%                 & 32                                                       & 1x                                                           & ---              \\
                                           & GIN(DQ)   & 95.5±0.4\%                  & 4                                                        & 8x                                                           & 1x         \\
                                           & GIN(ours) & \textbf{95.7±0.2\%}         & \textbf{3.75}                                            & \textbf{8.52x}                                              & \textbf{1.07x}  \\ \hline 
                                          %  & GAT(FP32) & 95.6±0.1\%                  & 32                                                       & 1x                                                           & ---              \\
                                          %  & GAT(DQ)   & 93.1±0.3\%                  & 4                                                        & 8x                                                           & 1x          \\
                                          %  & GAT(ours) & \textbf{93.9±1.0\%}         & \textbf{3.86}                                            & \textbf{8.28x}                                               & \textbf{1.13x} \\ \hline
   \multirow{6}{*}{\textbf{CIFAR10}}       & GCN(FP32) & 55.9±0.4\%                  & 32                                                       & 1x                                                           & ---              \\
                                           & GCN(DQ)   & 51.1±0.7\%                  & 4                                                        & 8x                                                           & 1x          \\
                                           & GCN(ours) & \textbf{52.5±0.8\%}         & \textbf{3.32}                                            & \textbf{9.62x}                                               & \textbf{1.25x} \\ \cline{2-6} 
                                          %  & GIN(FP32) & 57.5±0.7\%                  & 32                                                       & 1x                                                           & ---              \\
                                          %  & GIN(DQ)   & 50.7±1.6\%                  & 4                                                        & 8x                                                           & 1x          \\
                                          %  & GIN(ours) & \textbf{54.9±1.5\%}         & \textbf{3.53}                                            & \textbf{9.06x}                                               & \textbf{1.14x} \\ \cline{2-6} 
                                           & GAT(FP32) & 65.4±0.4\%                  & 32                                                       & 1x                                                           & ---              \\
                                           & GAT(DQ)   & 56.5±0.6\%                  & 4                                                        & 8x                                                           & 1x          \\
                                           & GAT(ours) & \textbf{64.7±2.8\%}         & \textbf{3.73}                                            & \textbf{8.57x}                                               & \textbf{1.12x} \\ \hline
   \multirow{3}{*}{\textbf{ZINC}}          & GCN(FP32) & 0.450±0.008                  & 32                                                       & 1x                                                           & ---              \\
                                           & GCN(DQ)   & 0.536±0.011                  & 4                                                        & 8x                                                           & 1x          \\
                                           & GCN(ours) & \textbf{0.492±0.056}         & \textbf{3.68}                                             & \textbf{8.68x}                                                & \textbf{1.08x} \\ \hline
   \multirow{3}{*}{\begin{tabular}[c]{@{}l@{}} \textbf{REDDIT-}\\ \textbf{BINARY}\end{tabular}} & GIN(FP32) & 92.2±2.3\%                  & 32                                                       & 1x                                                           & ---              \\
                                           & GIN(DQ)   & 81.3±4.4\%                  & 4                                                        & 8x                                                           & 1x          \\
                                           & GIN(ours) & \textbf{90.8±1.8\%}         & \textbf{3.50}                                             & \textbf{9.14x}                                                & \textbf{1.16x} \\ \bottomrule[2pt]
   \end{tabular}
   % \vspace{-1.5cm}
   \end{center}
   \end{table} 
\subsection{Node-Level Tasks}
Table \ref{node-level-results} shows the experimental results on three GNN architectures trained on four node-level 
datasets. 
Compared with DQ-INT4, our method can achieve significantly better accuracy on each task, even with a higher 
compression ratio, improving the inference performance with 1.28x to 2.00x speedups.
On almost all node-level tasks, our proposed $\rm A^2Q$ has 
negligible accuracy drop compared to the FP32 baselines while achieving
12.1x-18.6x compression ratio.
% While the two types of networks, GIN and GAT, 
% involve more complex computations compared to GCN, 
% such as the calculating attention coefficients in GAT, which makes it more difficult to 
% quantize these two models. 
% Therefore, DQ method on these two types of models is relatively poor,
Since both GIN and GAT involve more complex computations, such as the calculation of attention 
coefficients in GAT, 
it is more challenging to quantize those models, and DQ performs poorly on these two models. 
However, our method can overcome this problem and maintain comparable accuracy compared 
with the FP32 models. 
% This is because the $\rm A^2Q$ can learn the 
% appropriate quantization bitwidth for 
% each node features and therefore achieves a higher compression ratio while maintaining the accuracy. 
Our method can outperform the DQ-INT4 by 11.4\% on the GAT-Cora task with 
a smaller bitwidth (2.03 v.s. 4).
% For the dataset ogbn-arxiv with 169,343 nodes, 
% our method still works, only a 0.84\% degradation while achieving the 20.4x compression ratio compared with FP32 baseline, which 
% proves that our method is robust.
Even on ogbn-arxiv, which has a large number of
nodes, $\rm A^2Q$ can achieve a 12.1x compression ratio compared with FP32 baseline with comparable accuracy, 
which demonstrates the robustness of our method. 
% To further illustrate the superiority
% of our method, we also present a comparison with the binary quantization method 
% in Appendix \ref{comp_binary}. 
Moreover, to demonstrate the generality of our method, we also evaluate our method 
on heterogeneous graphs and the inductive learning tasks  
and compare with more related works in Appendix \ref{appendix_node_level_results}.

\vspace{-2pt}

% On Cora dataset, we increases the number of layers in GCN and count the average bits of node features for each layer except for the first layer in which the
% features $X\in \{0,1\}^{N\times F}$, the results are shown in Figure \ref{aloss_storge} which shows that the shallow layers 
% tend to use lower bits to quantize. We believe that this phenomenon is caused by the over-smooth in the deeper layers, in which 
% the features of a node will be very similar to its neighbor's. Therefore, to distinguish the different nodes correctly, the 
% nodes features in deep layers need to use higer bits to quantize.
\vspace{-0.cm}
\subsection{Graph-Level Tasks}
% \vspace{-0.5cm}
Table \ref{graph_level_result_1} presents the comparison results 
on the graph-level tasks. Our method can obtain better results on 
all tasks than DQ-INT4 with 
higher compression and a considerable speedup. Especially on the GIN-REDDIT-BINARY task, 
our method outperforms DQ-INT4 by 9.5\% while achieving a 1.16x speedup. 
Even for graph datasets with similar in-degrees, such as
MNIST and CIFAR10, our method also learns the appropriate bitwidths for higher 
compression ratio and better accuracy. Although on GIN-MINST task, 
the improvement of our method is  
relatively small due to the similarity of the in-degrees between different nodes, 
our method can achieve comparable accuracy with smaller bitwidth (3.75 v.s. 4).
% \textbf{Analysis:}
\begin{figure}[t]
\vspace{-0.4cm}

   \flushleft
   \subfigure[GCN-CiteSeer]{
      \begin{minipage}[t]{0.18\textwidth}
         \centering
         \includegraphics[width=1.13\linewidth,height=1.10001\linewidth]{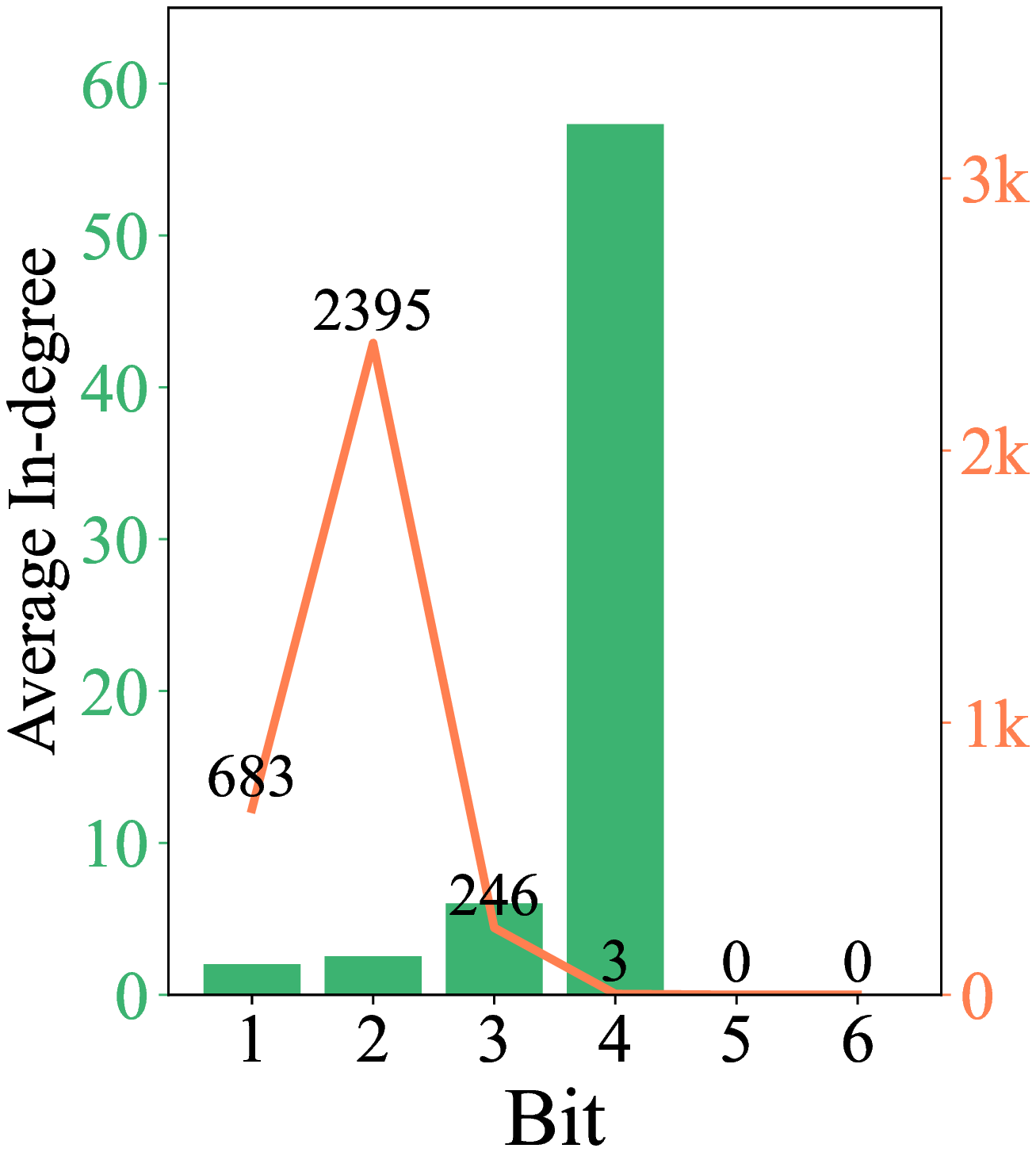}
         % \caption{GCN-CiteSeer}
         \label{gcn_cite_bit_deg}
      \end{minipage}
   }
   \subfigure[GIN-CiteSeer]{
      \begin{minipage}[t]{0.18\textwidth}
         \centering
         \includegraphics[width=1.08\linewidth,height=1.1\linewidth]{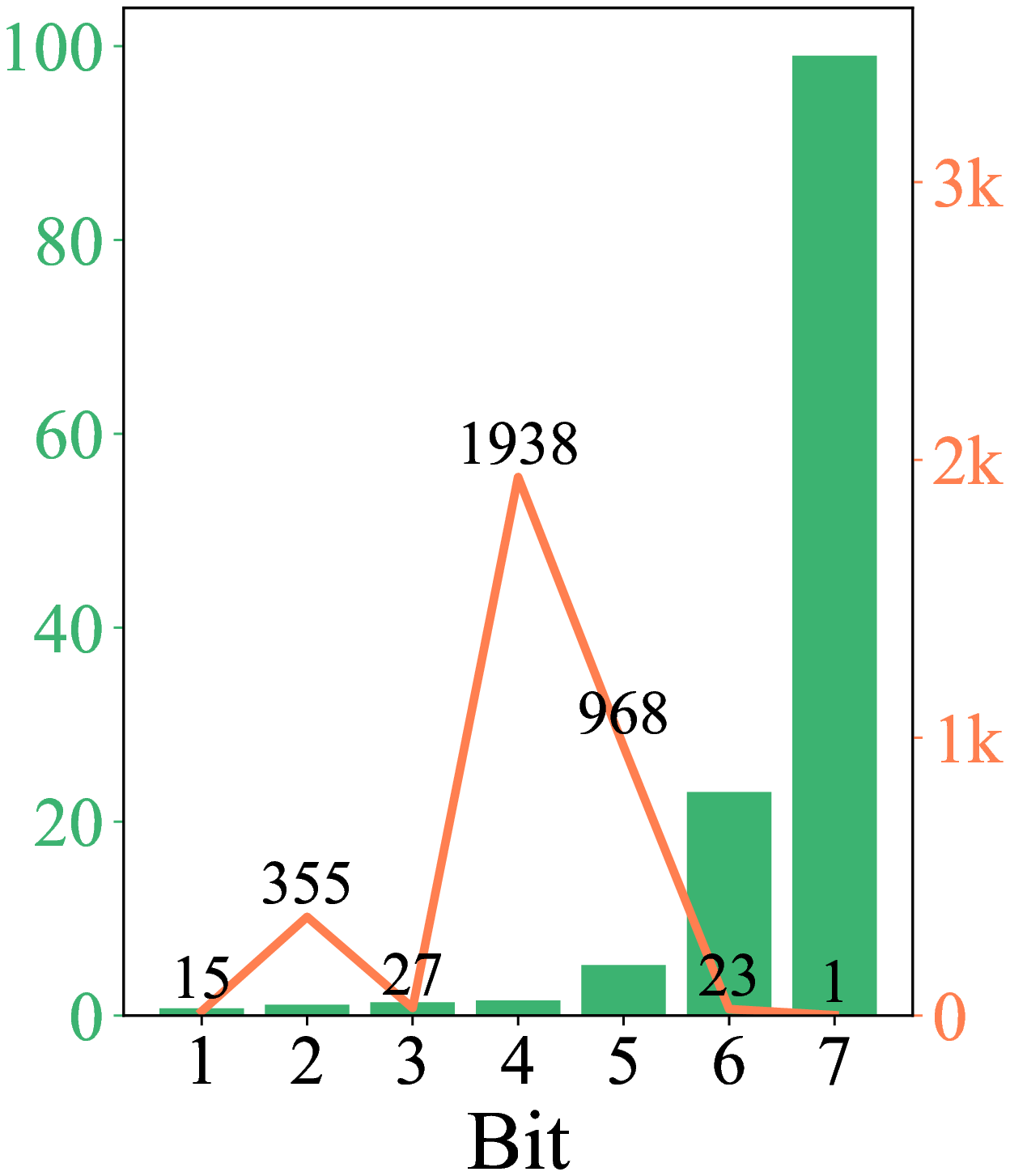}
         % \caption{GIN-CiteSeer}
         \label{gin_cite_bit_deg}
      \end{minipage}
   }
   \subfigure[GAT-CiteSeer]{
      \begin{minipage}[t]{0.18\textwidth}
         \centering
         \includegraphics[width=1.08\linewidth,height=1.1\linewidth]{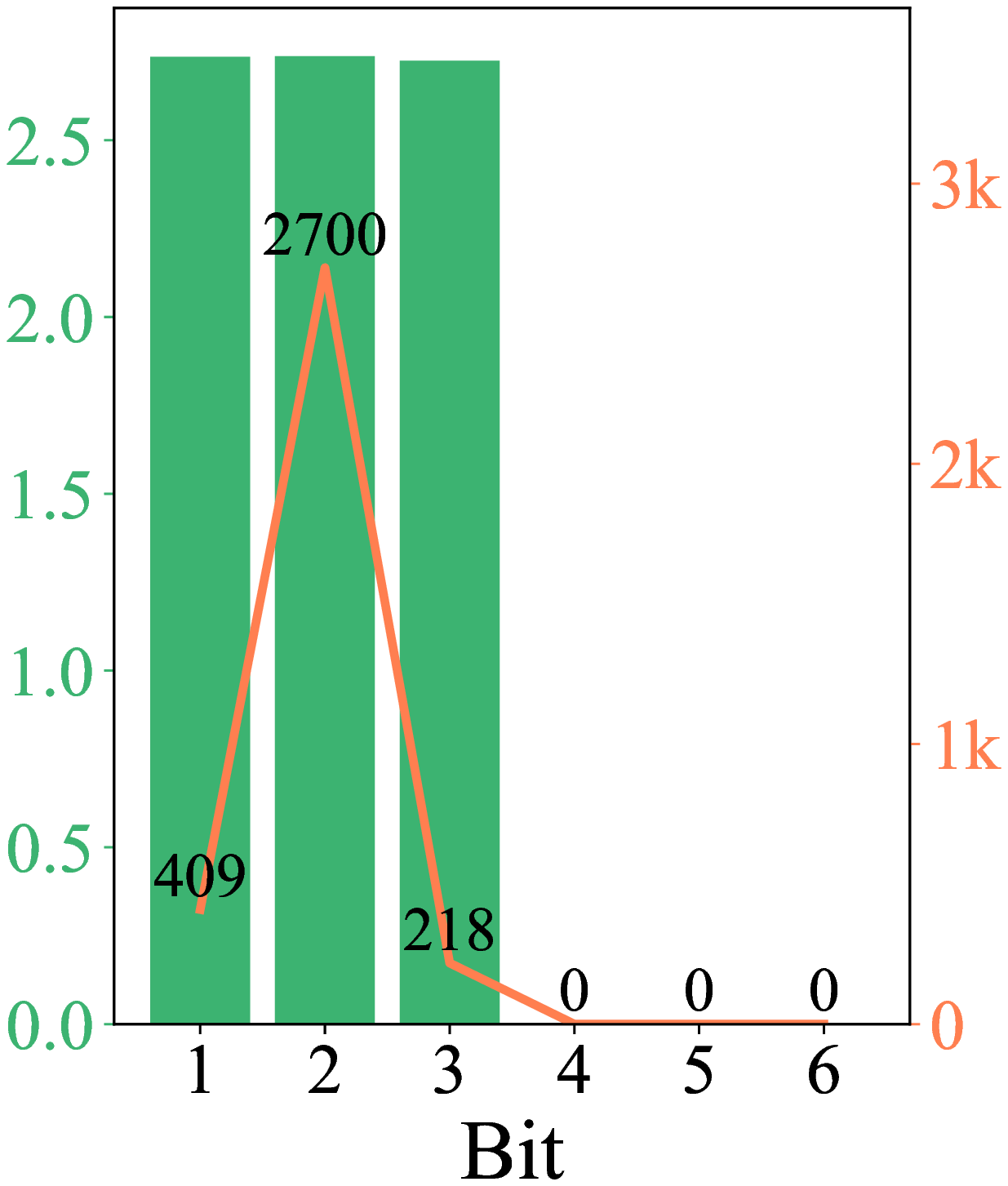}
         % \caption{GAT-CiteSeer}
         \label{gat_cite_bit_deg}
      \end{minipage}
   }
   \subfigure[The first layer]{
      \begin{minipage}[t]{0.18\textwidth}
         \centering
         \includegraphics[width=1.08\linewidth,height=1.1\linewidth]{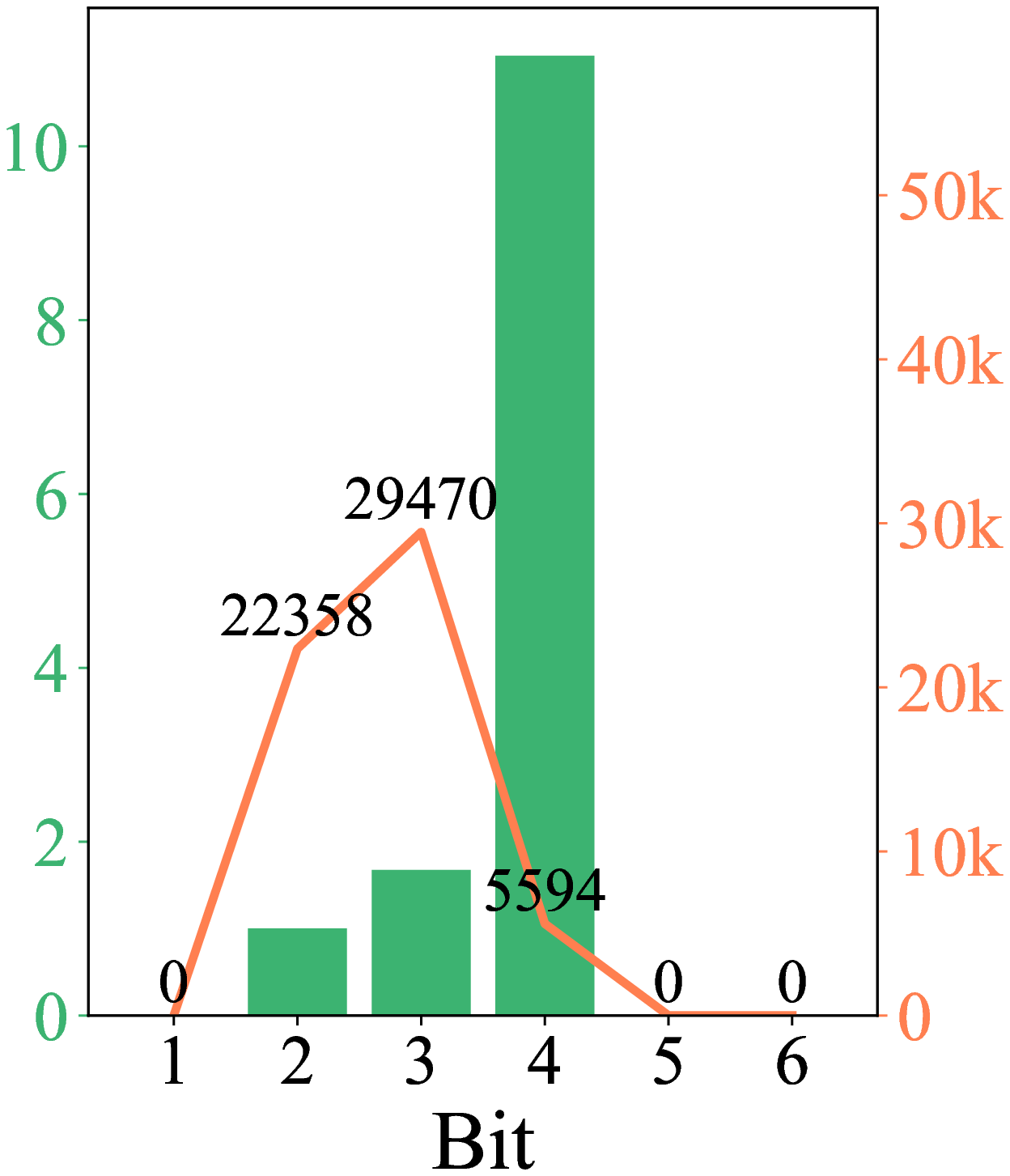}
         % \caption{GIN-CiteSeer}
         \label{gin_reddit_1}
      \end{minipage}
   }
   \subfigure[The second layer]{
      \begin{minipage}[t]{0.18\textwidth}
         \centering
         \includegraphics[width=1.13\linewidth,height=1.1\linewidth]{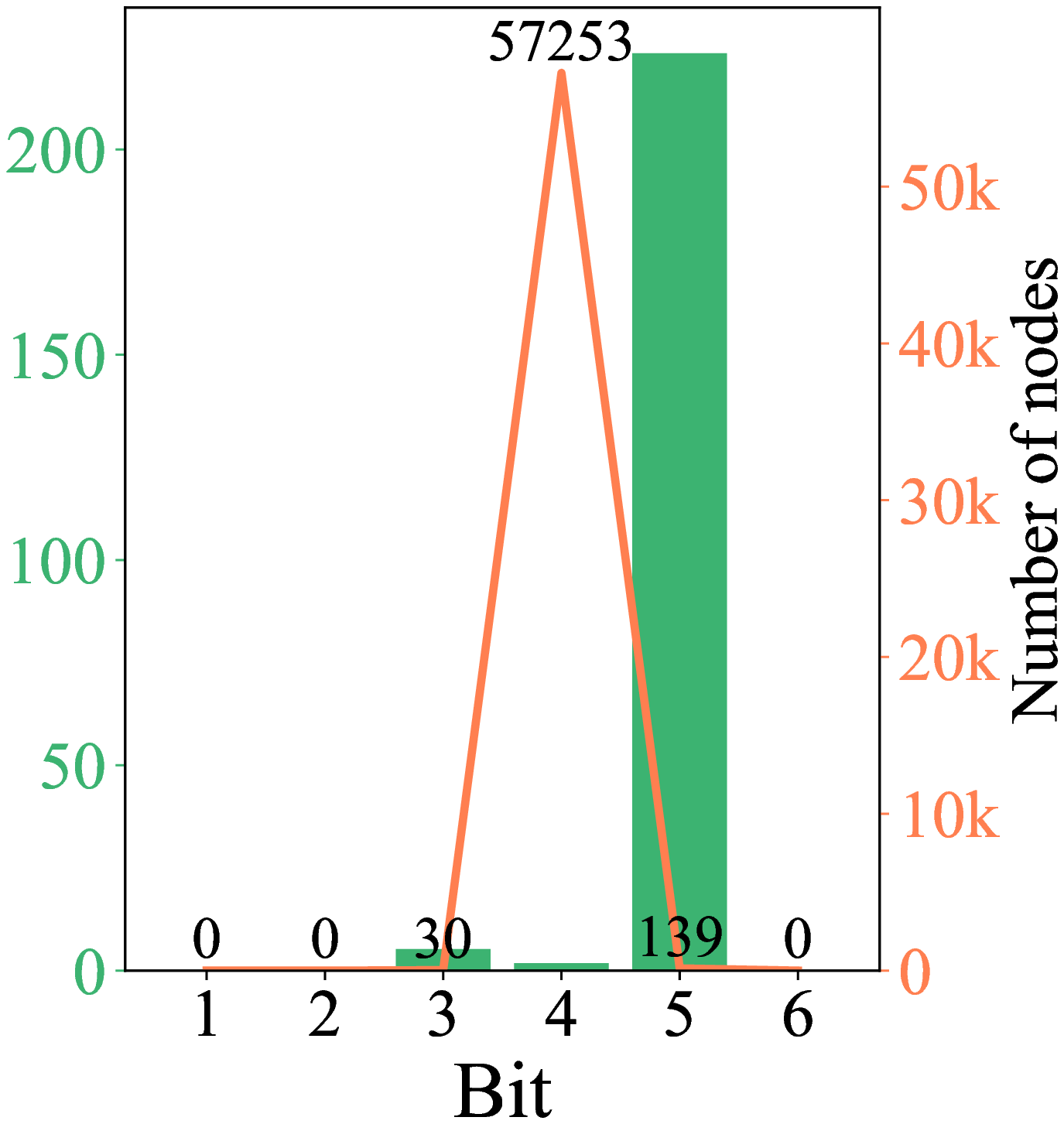}
         % \caption{GAT-CiteSeer}
         \label{gin_reddit_2}
      \end{minipage}
   }
   \caption{The relationship between quantized bitwidth and average in-degrees of nodes. 
   (a), (b) and (c) represent the results of three GNN models trained on CiteSeer. 
(d) and (e) are results 
about
the first
and the second layer of an MLP, which is the update function of GIN trained on REDDIT-BINARY.
The green bars represent the average in-degrees for the certain bitwidth used by nodes and
the orange polylines represent the number of the nodes that use this certain bitwidth.}
   \label{node_level_bit_deg}
   % \setlength{belowcaption}{-1cm}%
   % \begin{minipage}[t]{0.5\linewidth}
   %    \centering
   %    \includegraphics[width=1\linewidth,height=0.83\textwidth]{first_Figureeps}
   %    \caption{Comparisons of accuracy on Cora and REDDIT-BINARY for GNNs, e.g. GCN-Cora represent the task of GCN model on Cora dataset.
   %    $\rm{A^2Q-\#bit}$ where \# is the average bit of the node feature in the model quantized by our method.}
   %    \label{result_comparison}
   % \end{minipage}
\end{figure}
\subsection{Analysis}
To understand why our approach works, we analyze the relationship between the learned bitwidths and the 
topology of the graph.
% To demonstrate that our method is aggregation-aware and fully utilizes the property of GNNs and graph data,
% we analyze the relationship between the learned bitwidths and the topology of the graph.
Figure \ref{gcn_cite_bit_deg} and \ref{gin_cite_bit_deg} reveal that 
the bitwidth learned by $\rm A^2Q$ is strongly related to 
the topology of graph data in the node-level tasks. As the bitwidth increases, 
the average in-degrees of nodes become larger. In other words, $\rm A^2Q$ method tends to 
learn higher bitwidth for nodes with higher in-degrees. However, in GAT, as shown in 
Figure \ref{gat_cite_bit_deg}, the learned bits are irregular. 
This is because the features aggregated 
in GAT are topology-free. However, our method can still learn appropriate quantization 
bitwidths for 
different nodes, 
which improves accuracy while reducing memory usage.
In addition, Figure \ref{node_level_bit_deg} also shows the node 
distribution for different bitwidths and the result is consistent with power-law distribution. 
Since nodes in graph data mainly have low in-degrees, most of the nodes are 
quantized to low bitwidth ($\leq 4$), compressing the GNNs as much as possible. 
And there are also some high in-degree nodes quantized to high bitwidth, 
which can help to maintain the accuracy of the GNN models. As a result, 
the average bitwidth of the entire graph features is low, and the accuracy degradation is negligible.

For the graph-level tasks in which the number of nodes varies, 
our method is also aggregation-aware.
We select a layer of GIN trained on REDDIT-BINARY and analyze
the relationship 
between bitwidth and average in-degrees of nodes using the 
corresponding bitwidth to quantize in  
Figure \ref{gin_reddit_1} and \ref{gin_reddit_2}. 
It can be seen that the bitwidth 
learned for nodes features input to 
the second layer of MLP, which is the update function in GIN for graph-level tasks,
does not present a correlation with the topology of graph. 
We analyze the reason and find that the node features before 
the second layer is the result mapped by the first layer of 
MLP and is activated by the activation function, e.g., ReLU, which results
% \vspace{-0.3cm} 
in the 
node features losing the topology information.
We present more experiment results in Appendix \ref{more_results}.
to demonstrate 
that our method is generally applicable.

% \begin{figure}
%    \centering
%    \includegraphics[width=1\textwidth]{all_no_axis.eps}
%    % \caption{GAT-CiteSeer}
%    \label{gin_reddit_2}
% \end{figure}

\section{Ablation Study}

\begin{figure}[t]
   \begin{minipage}[]{0.48\textwidth}
      \vspace{-0.5cm}
      \captionof{table}{Ablation Study.}
      \label{leran_quantization_p}
      \begin{tabular}{ccll}
         \hline \toprule[2pt]
         Model                               & Config & Accuracy            & \begin{tabular}[c]{@{}l@{}}Average \\ bits\end{tabular}          \\ \midrule[1pt] 
         \multirow{5}{*}{\textbf{GIN-Cora}} & no-lr                   & 33.7±4.1\%          & 4                                                                \\
                                             & no-lr-b            & 75.6±0.2\%          & 4                                                                \\
                                             & no-lr-s           & 56.1±4.9\%          & 3.85                                                                \\
                                             & lr-all               & \textbf{77.8±1.6\%} & \textbf{2.37}                                                     \\ \hline
                                          %   & lr-bit               & \textbf{73.1±0.5\%} & \textbf{2.75}                                                    \\ \hline
         \multirow{3}{*}{\textbf{\begin{tabular}[c]{@{}c@{}}GCN-\\ CiteSeer\end{tabular}}} & FP32                   & 71.1±0.7\%          & 32                                                                \\
                                             & Global               & 56.8±6.7\%          & 3                                                                \\
                                             & Local                & \textbf{70.6±1.1\%} & \textbf{1.87}                                                    \\ \bottomrule[2pt] 
         \end{tabular}
   \end{minipage}%
   \qquad
   \
   \begin{minipage}[]{0.45\textwidth}
   \centering
   \includegraphics[scale=0.34]{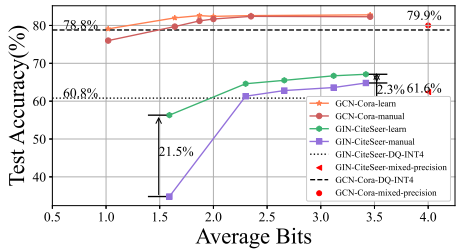}
   \caption{The comparison between learning bitwidth and assign manually.}
   \label{bit_hand_lr_comp}
   \end{minipage}
\end{figure}

\textbf{The advantage of learning-based mixed-precision quantization:}
In Figure \ref{bit_hand_lr_comp}, we compare our $\rm A^2Q$ with the manual mixed-precision method, 
which manually assigns high-bit to those nodes with high in-degrees and low-bit to those nodes 
with low in-degrees.
In the figure, the postfix ``learn'' denotes that using $\rm A^2Q$ method, 
``manual'' denotes that we assign bits to nodes and the model only learns the stepsize, and
``mixed-precision'' denotes that the model uses the same quantization method as 
DQ-INT4 but assigning different bitwidths
to nodes. For the ``mixed-precision'', we assign 5bits to 
those nodes with 50\% top 
in-degrees and assign 3bits to others.
The implications are two-fold. 
First, compared with the DQ-INT4, which uses the same quantization
bitwidth, the mixed-precision method obtains 1.1\% gains
on GCN-Cora tasks demonstrating that the mixed-precision method is more effective.
Second, the results of the learning method
outperform the manual method on all tasks.
Especially for the models with a high compression ratio, 
on GIN-CiteSeer task, learning method can achieve 21.5\% higher accuracy. 
This demonstrates that our learning method can perform better 
than the
assignment method according to prior knowledge
for mixed-precision 
quantization of GNNs.

\textbf{The power of learning the quantization parameters:} 
Ablations of two quantization parameters $(\vs, \vb)$ on the GIN-Cora task 
are reported in the first row of Table \ref{leran_quantization_p}. 
The ``no-lr'' denotes that do not use learning method,
``no-lr-b'' denotes 
that only learn the step size $\vs$
, ``no-lr-s'' denotes 
that only learn the bitwidths $\vb$, and ``lr-all'' denotes 
that learn the bitwidth and step size simultaneously. We can see that 
learning the 
step size can significantly increase the accuracy and 
even the ``no-lr-bit'' model can
outperform the DQ-INT4 at the same compression ratio.
When learning the bitwidth and step
size simultaneously, the model can achieve higher accuracy with a higher compression ratio.
This is because our method learns 
lower bitwidths for most nodes with low in-degrees and higher bitwidths for a tiny fraction of
nodes with high in-degrees, which can improve the compression ratio while achieving higher accuracy.

\textbf{Local Gradient v.s. Global Gradient:} 
To demonstrate the effectiveness of our 
Local Gradient method, 
we compare the 
models trained with and without it on the GCN-CiteSeer task in the last row of 
Table \ref{leran_quantization_p}. 
The ``Global'' denotes
that the model is trained with Eq. \ref{loss_grad_s} and Eq. \ref{loss_grad_b}. 
The model trained with the local method outperforms
the global method by 13.8\% with a higher compression ratio. 
This is because the Local Gradient method can learn quantization 
parameters for all nodes, while only quantization parameters 
for a part of nodes can be updated with the Global Gradient method 
due to the extreme sparse connection in the graph on the node-level semi-supervised tasks.

\textbf{The overhead of Nearest Neighbor Strategy:} 
We evaluate the real inference time of the GIN model on the 2080ti GPU. 
On REDDIT-BINARY task, the model without the selection process requires 121.45ms, 
while it takes 122.60ms for the model with our Nearest Neighbor Strategy, 
which only introduces 0.95\% overhead. But with the help of the 
Nearest Neighbor Strategy, our model can obtain 19.3\% accuracy gains for quantized
GIN on REDDIT-BINARY.
\vspace{-0.2cm}
\section{Conclusion}
This paper proposes $\rm A^2Q$, an aggregation-aware mixed-precision quantization method for 
GNNs, and
introduces the Local Gradient and Nearest Neighbor Strategy to generalize $\rm A^2Q$ to 
the node-level and graph-level
tasks, respectively. Our method can learn the quantization parameters for different nodes
by fully utilizing the property of GNNs and graph data.
The model quantized by our $\rm A^2Q$ can achieve up to a 18.6x compression ratio, and the accuracy 
degradation is negligible 
compared with the FP32 baseline. Compared with the prior SOTA, DQ-INT4, our method can 
significantly improve 11.4\% accuracy with up to a 2.00x speedup on different tasks.
Our work provides a 
general, robust and feasible solution to speed up the inference of GNNs. 
% which is friendly to the resource-constrained environment. 

\bibliography{iclr2023_conference}
\bibliographystyle{iclr2023_conference}

\newpage
\appendix
\section{Appendix}
\subsection{Uniform Quantization}
\label{more uniform quantization}
In this section, we will give a detailed introduction to the content related to quantification.
\subsubsection{Quantization Process}
\begin{figure}[H]
   \centering
   \includegraphics[scale=0.35]{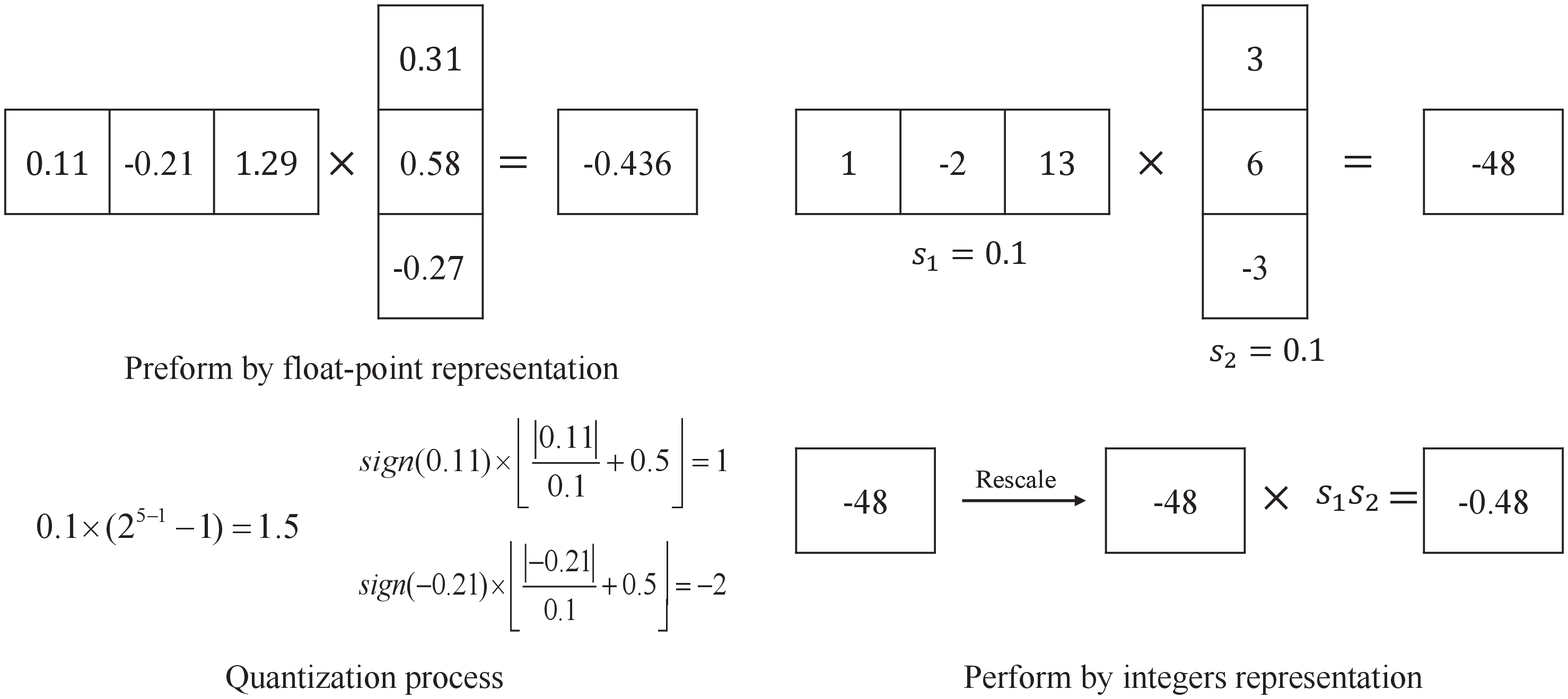}
   \caption{An example of performing inner product by integers representation.}
   \label{quant_cal_vector}
\end{figure}

For a vector $\vx$, the $\vx_q$ is a quantized representation. Given the 
quantization step size $s$, $s\in \mathbb{R_+}$, and the quantization 
bitwidth $b$, $b\in \mathbb{N_+} $, then the uniform quantization
is implemented as:
\begin{gather}
   \bar{\vx}=sign(\vx) \begin{cases}
      \lfloor \frac{\left| \vx \right|}{s}+0.5 \rfloor \text{,} & \left| \vx \right|<s(2^{b-1}-1) \\
      \\
      2^{b-1}-1 \text{,} & \left| \vx \right| \geq s(2^{b-1}-1)  
   \label{quantize_appendix}
   \end{cases}
   \ \text{.} 
   % \vx_q=s\cdot \bar{\vx} \label{quantize_2}
\end{gather}
The $\vx$ at 32bits is mapped to the integer number set $\{-2^{b-1}+1,...,0,...,2^{b-1}-1\}$ 
where the bitwidth is $\#b$ bits,  
and the quantized representation can be calculated as $\vx_q=s\cdot \bar{\vx}$.
For inference, $\bar{\vx}$ can be used to compute matrix multiplication in the update phase or 
perform other computations 
in GNNs layers 
% e.g., aggregation by integer operations which is much faster than float-point operations, 
and the output of these computations then are rescaled by
the corresponding $s$ using a relatively lower cost scalar-vector multiplication.
As an illustrative example, for vectors $\vx \in \mathbb{R} ^{3\times 1}$ and 
$\vy \in \mathbb{R} ^{3\times 1}$, 
the quantization 
parameters are both $s=0.1$, $b=5$, the process of inner product between these two vectors by integers is shown
in Figure \ref{quant_cal_vector}. When the values in a vector are all non-negative, we do 
not need
to represent the sign bit in the fixed-point representation. Therefore, the value can use 
$\#b$ bits to quantize instead of using the first bit to represent the sign bit. Then the
quantization range of uniform quantization is $[-s(2^{b}-1),s(2^b-1)]$.
\subsubsection{Gradient in Backpropagation} 
Due to the floor function used in the quantization process is not differentiable, 
the gradient of $x_q$ with respect to $x$ vanishes almost everywhere, 
which makes it
impossible to train the model by the backpropagation algorithm. Therefore, 
we use the straight-through estimator \citep{bengio2013estimating} to
approximate the gradient through the floor function, i.e., 
$\frac{\partial L}{\partial x}=\frac{\partial L}{\partial x_q}\mathbb{I}_{|x|\leq s(2^b-1)}$, where 
$\mathbb{I}_{|x|\leq s(2^b-1)}$ is a indicator function, whose value is $1$
when $|x|\leq s(2^b-1)$, and vice versa.
In our paper, the quantification parameters $(s,b)$ are learnable, the gradients of $x_q$ 
w.r.t.
$(s,b)$ used in Eq. \ref{loss_grad_s} and Eq. \ref{loss_grad_b} are:
\begin{gather}
   \begin{bmatrix}
      \frac{\partial x_{q}}{\partial s}\\
      \\
      \frac{\partial x_{q}}{\partial b}
   \end{bmatrix}=\begin{cases}
      \begin{bmatrix}
         \frac{1}{s}\left( x_{q}-{{x}} \right)\\
         0
      \end{bmatrix} \text{,} & \left| \vx \right|<s(2^{b-1}-1)\\
      sign(x)\begin{bmatrix}
         \left( {{2}^{b-1}}-1 \right)\\
         {{2}^{b-1}}\ln \left( 2 \right)s
      \end{bmatrix} \text{,} & \left| \vx \right| \geq s(2^{b-1}-1)
   \end{cases}
   \ \text{.}
\end{gather}

\linespread{1.5}
\begin{table}[H]
   \caption{The aggregation functions and update functions for GNNs used in this paper, $d_i$ denotes the degree of node $i$, 
   the $\varepsilon $ denotes a learnable constant, and $\alpha$ represent 
   attention coefficients.}
   \label{gnn_formula}
   \renewcommand\arraystretch{1.5}
   \begin{center}
   \begin{tabular}{lllll}
   \toprule[2pt]
   \multicolumn{1}{c}{Model} & \multicolumn{2}{c}{Aggregation function} & \multicolumn{2}{c}{Update function}      \\ \midrule[1pt]
   GCN                       & \multicolumn{2}{c}{$\vh_{i}^{( l )}=\sum\limits_{j\in \mathcal{N} ( i )\cup \{i\}}{\frac{1}{\sqrt{{{d}_{i}}}\sqrt{{{d}_{j}}}}\vx_{j}^{\left( l-1 \right)}}$} & \multicolumn{2}{c}{$\vx_{i}^{\left( l \right)}=ReLU( {{\mW}^{\left( l \right)}}\vh_{i}^{\left( l \right)}+{{\vb}^{\left( l \right)}} )$} \\ \hline
   GIN                       & \multicolumn{2}{c}{$\vh_{i}^{( l )}=( 1+{{\varepsilon }^{\left( l \right)}})\vx_{i}^{\left( l-1 \right)}+\sum\limits_{j\in \mathcal{N} \left( i \right)}{\vx_{j}^{\left( l-1 \right)}}$} & \multicolumn{2}{c}{$\vx_{i}^{\left( l \right)}=ML{{P}^{( l )}}( \vh_{i}^{\left( l \right)},{{\mW}^{\left( l \right)}},{{\vb}^{\left( l \right)}} )$} \\ \hline
   GAT                       & \multicolumn{2}{c}{$\vh_{i}^{( l )}=\sum\limits_{j\in \mathcal{N} \left( i \right)\cup \left\{ i \right\}}{\alpha _{i,j}^{\left( l \right)}}\vx_{j}^{\left( l-1 \right)}$} & \multicolumn{2}{c}{$\vx_{i}^{\left( l \right)}={{\mW}^{\left( l \right)}}\vh_{i}^{l}+{{\vb}^{\left( l \right)}}$} \\ \bottomrule[2pt]
   \end{tabular}
   \end{center}
   \end{table}
\begin{table}[H]
   \caption{The statistics for density of adjacency matrix and the labeled nodes in four node-level datasets.}
   \label{sparsity of a}
   \begin{center}
         \begin{tabular}{lllll}
            \toprule[2pt]
                        & Cora    & CiteSeer & PubMed  & ogbn-arxiv \\ \midrule[1pt]
            Density of A & 0.144\% & 0.112\%  & 0.028\% & 0.008\%    \\
            Labled nodes & 5.17\%  & 3.61\%   & 0.30\%  & 53.70\%    \\ \bottomrule[2pt]
            \end{tabular}
   \end{center}
\end{table}
\subsection{More about Graph Neural Networks}
\label{more gnn}
In this section, we first give detailed information about the MPNN framework \citep{gilmer2017neural}, and
then provide a detailed examination of the three GNNs used in our papers.

A graph $\gG=(\mathcal{V}, \mathcal{E} )$ consist of nodes $\mathcal{V}=\{1,...,N\}$ and edges 
$\mathcal{E} \subseteq \mathcal{V}\times \mathcal{V}$ has node features $\mX \in \mathbb{R}^{N\times F}$ and 
optionally H-dimensional
edge features $\mE \in \mathbb{R}^{E\times H}$. The MPNN framework can be formulated by 
$\vx_{i}^{(l)}={{\gamma }^{(l)}}( \vx_{i}^{( l-1)},
\underset{j\in \mathcal{N} ( i )}{\mathop{\square }}\,
{{\phi }^{( l )}}( \vx_{i}^{( l-1 )},\vx_{j}^{( l-1 )},
\ve_{ij}^{( l-1 )} ))$, where ${\phi}$ is a differentiable kernel function, $\mathop{\square}$ is the aggregation 
function 
which is permutation-invariant, and the $\gamma$
is a learnable update function,
$\vx_i$ is the features of node $i$ and $\ve_{ij}$ is 
the features of edge between node $i$ and $j$, $\mathcal{N}(i)=\{j:(i,j)\in \mathcal{E} \}$, and 
$l$ represents the $l$-th layer of the GNNs.

In this paper, we focus on three typical GNN models whose forwardpass all can be represented by 
the MPNN framework, 
Graph Convolution Network (GCN) \citep{kipf2016semi}, Graph Isomorphism Network (GIN) \citep{xu2018powerful}, 
and Graph Attention Network (GAT) \citep{velivckovic2017graph}.
the detailed information is shown in Table \ref{gnn_formula}.
\subsection{Proofs of Theoretical Results}
\label{proof2}
This section provides formal proof of the theoretical results of our paper.
\subsubsection{Notations}
Here, we define the notations utilized in our proof. $A=\{0,1\}^{N\times N}$ is the adjacency matrix that
indicates whether there is an edge between each pair of nodes, e.g., if there is an edge between node $i$ and
node $j$, then $a_{ij}=1$, otherwise, $a_{ij}=0$. Then, $\tilde{A}=A+I$ is the adjacency matrix for a graph that
is added to the self-loops. The degree matrix $D=diag({d_1,d_2,...,d_n})$, where 
${{d}_{i}}=\sum\nolimits_{j}{{{a}_{ij}}}$ and the degree matrix for the graph having self-loops is 
$\tilde{D}=({\tilde{d}_1,\tilde{d}_2,...,\tilde{d}_n})$, where $\tilde{d}_i=\sum\nolimits_{j}{{{\tilde{a}}_{ij}}}$.
\subsubsection{Proofs}
\label{proof}
\textbf{Proof 1.} \textit{The gradients of the loss 
function with respect to the node features in semi-supervised tasks are most zero.}

Without loss of generality, we use the GCN model as an example. From the Table \ref{gnn_formula}, the graph
convolution operation can be described as
\begin{gather}
    {{\mX}^{(l+1)}}=\sigma (\hat{\mA}{{\mX}^{(l)}}\mW^{(l)}) \text{,}
\end{gather}
where $\hat{\mA}=\tilde{\mD}^{-\frac{1}{2}} \tilde{\mA}\tilde{\mD}^{-\frac{1}{2}}$, is the normalized adjacency matrix, 
$\mW^{(l)}\in \mathbb{R} ^{F_{in}\times F_{out}}$ 
is a learnable weight matrix in the $l$-th layer of GCN. $\mX^{(l)}$ is the input of the $l$-th layer and the 
output of the $(l-1)$-th layer in GCN. $\sigma$ is the non-linear activation function, e.g., ReLU. Note that
the $\hat{\mA}$ is an extreme sparse matrix for node-level datasets in
our paper.

In our training process of the model, we use $nll\_loss$ as our task loss function $L$. Only the nodes
in the train set $T$ have labels. 
For the last layer of GCN, we get the node features to be classified by $\mH^{(l+1)}=softmax(\mX^{(l+1)})$.
Then the gradient of $L$ with respect to $\mX^{(l+1)}$ is
\begin{gather}
    \mG^1={{\nabla }_{{{\mX}^{(l+1)}}}}L=\frac{\partial L}{\partial \mH^{(l+1)}}\cdot \frac{\partial \mH^{(l+1)}}{\partial \mX^{(l+1)}}=
    [{{l}_{ij}}]\in {{\mathbb{R}}^{N\times {{F}_{out}}}}\text{,}
\end{gather}
where only the $\mG_{i,:}^1,\ i\in T$ is not zero, otherwise, $\mG_{i,:}^1=0$.
Then, the gradient of the loss function with respect to $\mX^{(l)}$ is
\begin{gather}
    \mG^2={{\nabla }_{{{\mX}^{(l)}}}}L={\hat{\mA}^{T}}({{{\nabla }_{{{\mX}^{(l+1)}}}}L}\odot {{\sigma}'(\hat{\mA}{{\mX}^{(l)}}\mW^{(l)})}){(\mW^{(l)})^T}\text{.}
\end{gather}
For node $j$ do not have an edge with the node $i$, $i\in T$, $\mG_{j,:}^2=0$. 
Table \ref{sparsity of a} lists the density of the adjacency matrix $\mA$ and the 
percentage of the labeled nodes in four node-level datasets.
Because the sparsity property of adjacency
matrix and the nodes with trained labels only account for a tiny fraction of the graph, 
the gradients from the loss function for 
most node features are zero. 

\textbf{Proof 2.} \textit{The normalized adjacency matrix $\hat{\mA}$ is not 
needed to be quantized for the GCN model.}

We take the process of $\mX\mW\rightarrow \mA(\mX\mW)$ as an illustrative example, which
represents first calculate the $\mB=\mX\mW$ and then calculate $\mA\mB$.
For the $l$-th layer of FP32 models, the first stage is $\mB_l=\mX_l\mW_l$, 
and then calculate the $X_{l+1}=\hat{\mA}\mB_l$, where 
$\mX_l\in \mathbb{R}^{N\times F_1}$, $\mW_l \in \mathbb{R}^{F_1\times F_2}$ and
$\mA\in\mathbb{R}^{N\times N}$.
The step-size for $\mB_l$, $\mX_l$ and $\mW_l$ 
is $\mS_{B_l}$, $\mS_{X_l}$ and $\mS_{W_l}$, respectively. 
And they are all diagonal matrices. 
% The 
% more specific form of the step-size matrices can be found in lines 123-138 of our main text.
The integer representations are calculated as $\mB_{l}=\mB_{l\_q}\mS_{B_l}$, 
$\mX_{l}=\mS_{X_l}\mX_{l\_q}$ and $\mW_{l}=\mW_{l\_q}\mS_{W_l}$.
Note that for the node-level tasks, we can obtain the 
$\mS_{B_l}$, $\mS_{X_l}$ and $\mS_{W_l}$
in advance. And for the graph-level tasks, we can obtain them through one more 
element-wise multiplication whose overhead is negligible, as the comparison in Table \ref{float_int_comp}.
Then the first stage is: 
\begin{equation}
    \begin{split}
        \mB_l &= \mX_l\cdot \mW_l \\
              &= (\mS_{X_l}\cdot {\mX_{l\_q}})\cdot (\mW_{l\_q}\cdot \mS_{W_l})
    \end{split} \ \text{,}
\end{equation}
and 
there exists $\mB_l = \mB_{l\_q}\mS_{B_l}$.
Therefore, the integers representation for the next stage can be calculated as:
\begin{equation}
   \label{rebuttal_b_fuse}
   \begin{split}
      \mB_{l\_q} &=  \mB_l \mS^{-1}_{B_l} \\
            &= (\mS_{X_l}\cdot {\mX_{l\_q}})\cdot (\mW_{l\_q}\cdot \mS_{W_l})\mS^{-1}_{B_l}\\
            &= (\mS_{X_l}\cdot {\mX_{l\_q}})\cdot (\mW_{l\_q}\cdot (\mS_{W_l}\mS^{-1}_{B_l}))\\
            &= (\mS_{X_l}\otimes (\mS_{W_l}\mS^{-1}_{B_l})) \odot ({\mX_{l\_q}}\cdot {\mW_{l\_q}})
   \end{split}\ \text{,}
\end{equation}
where the $(\mS_{X_l}\otimes (\mS_{W_l}\mS^{-1}_{B_l}))$ can be calculated offline.
Then we obtain the fixed-point representation $B_{l\_q}$ for the next stage and do not introduce
overhead.

The process of node degree normalization after the aggregation process can be represented 
as 
$\mX_{l+1} = \sigma(\hat{\mA}\mB_l)$, where 
$\hat{\mA}=\mD^{-\frac{1}{2}} \tilde{\mA}\mD^{-\frac{1}{2}}$ is 
the normalized adjacency matrix,
and 
$\sigma$ is the non-linear activation function. 
$\mD^{-\frac{1}{2}}$ at the right side of $\tilde{\mA}$ can be fused into 
the $\mS_{X_l}$ and then calculate $\mB_{l\_q}$ as Eq. \ref{rebuttal_b_fuse}.

Then the features of the $(l+1)$-th layer $\mX_{l+1}$ can be obtained as 
$\mX_{l+1} = \sigma(\mD^{-\frac{1}{2}} \tilde{\mA}\mB_{l\_q})$. 
And there exits
$\mX_{l+1} = \mS_{X_{l+1}}\mX_{(l+1)\_q}$. Therefore, the $\mX_{(l+1)\_q}$ can 
be obtained as:
\begin{equation}
    \begin{split}
        \mX_{(l+1)\_q} &= \mS_{X_{l+1}}^{-1}\mX_{l+1}\\
                       &= \mS_{X_{l+1}}^{-1}\sigma(\mD^{-\frac{1}{2}}\tilde{\mA}\mB_{l\_q})
    \end{split}\ \text{.}
\end{equation}
Note that the elements in diagonal matrix $\mS_{X_{l+1}}$ are all positive because
this matrix is made up of step-size, which is always positive.
Then we can obtain
$\mX_{(l+1)\_q}=\sigma(\mS_{X_{l+1}}^{-1}\mD^{-\frac{1}{2}}\tilde{\mA}\mB_{l\_q})$
, where $\mS_{X_{l+1}}^{-1}\mD^{-\frac{1}{2}}$ can be obtained before inference and 
$\tilde{\mA}\in \{0,1\}^{N\times N}$. The computation of  $\tilde{\mA}\mB_{l\_q}$ 
only has addition operations and the $\mS_{X_{l+1}}^{-1}\mD^{-\frac{1}{2}}$ can be 
obtained before inference for node-level tasks or introduce only once more element-wise
multiplication to calculate for the graph-level tasks.

The $\mD^{-\frac{1}{2}}$ at the left side is fused into
the element-wise multiplication performed by the next layer and 
the $\mD^{-\frac{1}{2}}$ at the right side
is fused into the element-wise multiplication performed by the current layer and the 
element-wise multiplication is a necessary stage in the quantized model.
Therefore, we can perform the node degree normalization 
using fixed point addition operation instead of quantizing the normalized adjacency matrix which
may introduce more quantization error.

\textbf{Proof 3.} \textit{The quantization process can be fused with Batch Normalization operations.}

When GNNs have Batch Normalization (BN) Layers, the calculation process
is as follows (Note that we have fused the mean and standard-deviation with
the learned parameters in BN):
\begin{equation}
   \begin{split}
      \mX_{l+1} &= BN(\sigma(\hat{\mA}\mB_{l\_q}))\\
                &= \sigma(\hat{\mA}\mB_{l\_q})\mY  + \mZ 
   \end{split} \ \text{,}
\end{equation}
where $\mY = diag(y_1,y_2,...,y_{F_2})\in \mathbb{R}^{F_2\times F_2}$, 
$\mZ = (\vz_1, \vz_2,...,\vz_{F_2})\in \mathbb{R}^{N\times F_2}$ and 
$\vz_i=(\theta_i,\theta_i,...,\theta_i)^T\in \mathbb{R}^N$ among which $y_i$ and 
$\theta_i$ are the BN parameters for the i-th dimension feature of the nodes features.
And there exits that $\mX_{l+1}=\mS_{X_{l+1}}\mX_{l+1\_q}$. Therefore, 
\begin{equation}
   \label{bn_fuse}
   \begin{split}
      \mX_{l+1\_q} &= \mS_{X_{l+1}}^{-1}\mX_{l+1}\\
                  &= \mS_{X_{l+1}}^{-1}(\sigma(\hat{\mA}\mB_{l\_q})\mY  + \mZ )\\
                  &= (\mS_{X_{l+1}}^{-1} \otimes \mY) \odot (\sigma(\hat{\mA}\mB_{l\_q})) 
                  + \mS_{X_{l+1}}^{-1}\mZ 
   \end{split}\ \text{.}
\end{equation}
Through Eq. \ref{bn_fuse},
we can fuse the quantization of the next layer into the BN operation of the current layer, which will 
not introduce
overhead because the BN layer itself requires floating point operations. Note that the
float point operations are also element-wise.

\begin{figure}[t]
   \centering
   \includegraphics[scale=0.35]{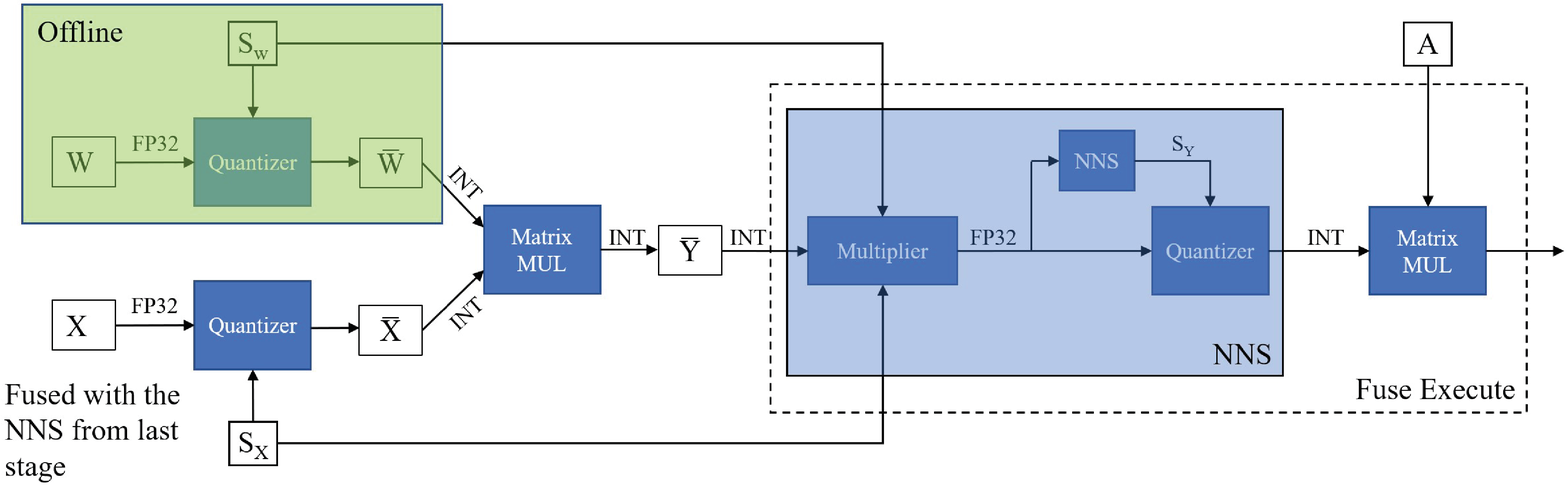}
   \caption{The pipeline of the quantization process on our accelerator.}
   \label{pipeline}
\end{figure}

\begin{table}[H]
   \caption{The comparison between fixed-point operations and float-point operations for some tasks using the 
   Nearest Neighbor Strategy.}
   \label{float_int_comp}
   \begin{center}
       \begin{tabular}{lllll}
           \hline \toprule[2pt]
           Task     & GIN-RE-IB & GCN-MNIST & GAT-CIFAR10 & GCN-ZINC \\ \midrule[1pt]
           Fixed-point(M)   & 936.96    & 455.69    & 1387.98     & 504.62   \\
           Float-point(M) & 7.35      & 2.06      & 13.71       & 1.74     \\
           Ratio    & 0.78\%    & 0.45\%    & 0.98\%      & 0.34\%   \\ \bottomrule[2pt]
           \end{tabular}
   \end{center}
   \end{table}

\subsection{The Overhead Analysis of Nearest Neighbor Strategy}

Through our dedicated hardware and the 
optimized pipeline, we reduce the overhead introduced by the Nearest Neighbor
Strategy (NNS) as much as possible.
As the pipeline is shown in Figure \ref{pipeline}, we fuse the (NNS) with the 
following operations. The fixed-point results produced by the previous stage
are used to first multiply the corresponding step-size from
the previous stage (an element-wise float point multiplication) and 
then execute the NNS process. After getting the step-size, 
these features are quantized immediately (an element-wise float point multiplication). 
Therefore, through this fusion way, 
we do not need the extra memory to store
a copy of
FP32 features. 

In addition, the overhead of the NNS is from one more 
element-wise float point multiplication and the search process. 
We provide a comparison of the number of float-point operations and fixed-point 
operations for different graph-level tasks in Table \ref{float_int_comp}, where 
`Fixed-point' denotes the fixed-point operation, `Float-point' denotes the float-point operation and
the `Ratio' denotes the percentage of the float-point operations in the overall 
process. The extra float-point operations introduced by NNS is only a tiny fraction of the fixed-point
operations.
On the other hand, through our optimized pipeline and the comparator array 
used in our accelerator 
the latency introduced by the 
search process of the NNS can be overlapped.
Therefore, the overhead introduced by NNS is negligible.

\subsection{Datasets}
\label{datsets}
\begin{figure}[ht]
   \centering
   \subfigure[Cora]{
      \begin{minipage}[t]{0.4\textwidth}
         \centering
         \includegraphics[scale=0.38]{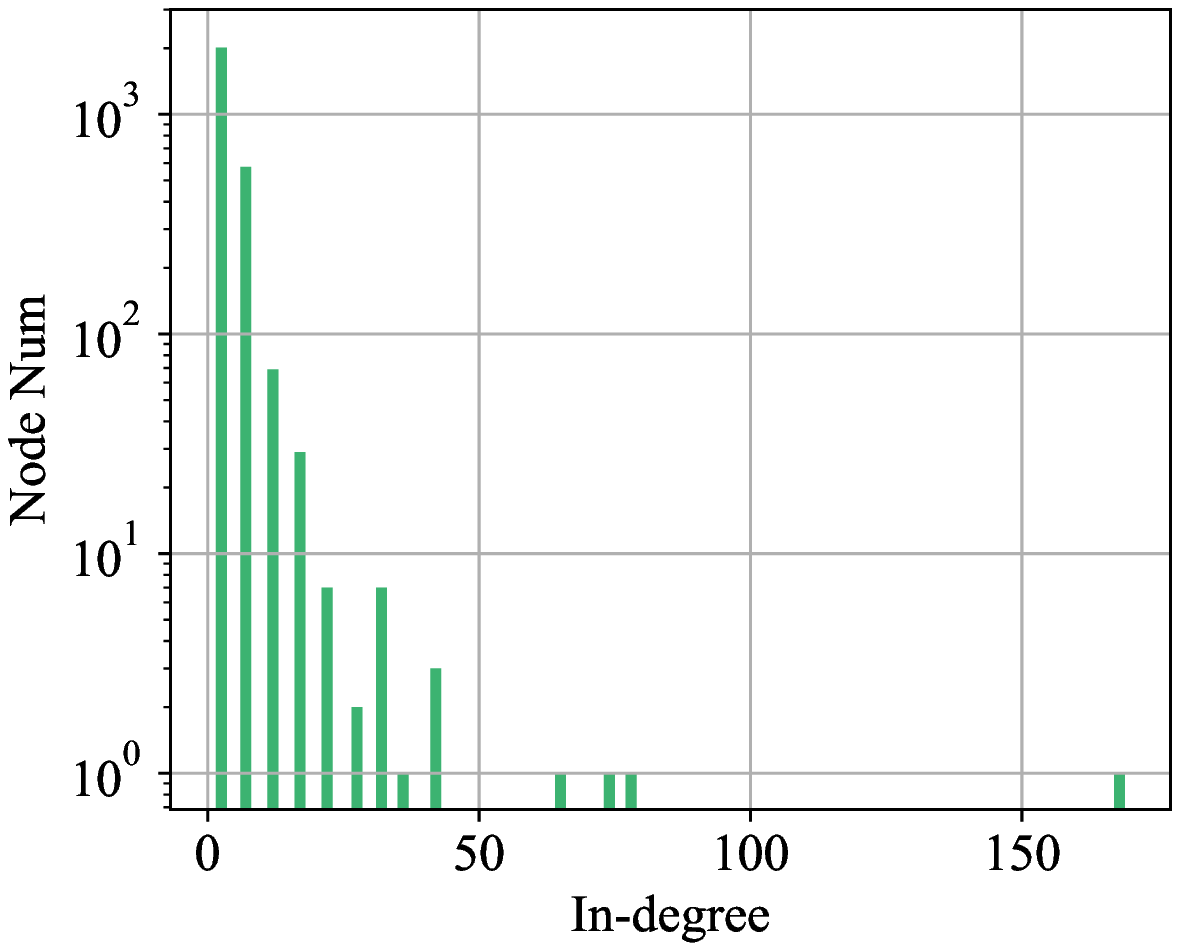}
         % \caption{GCN-CiteSeer}
       %   \label{gcn_cifar_1}
      \end{minipage}
   }
   \subfigure[CiteSeer]{
      \begin{minipage}[t]{0.4\textwidth}
         \centering
         \includegraphics[scale=0.38]{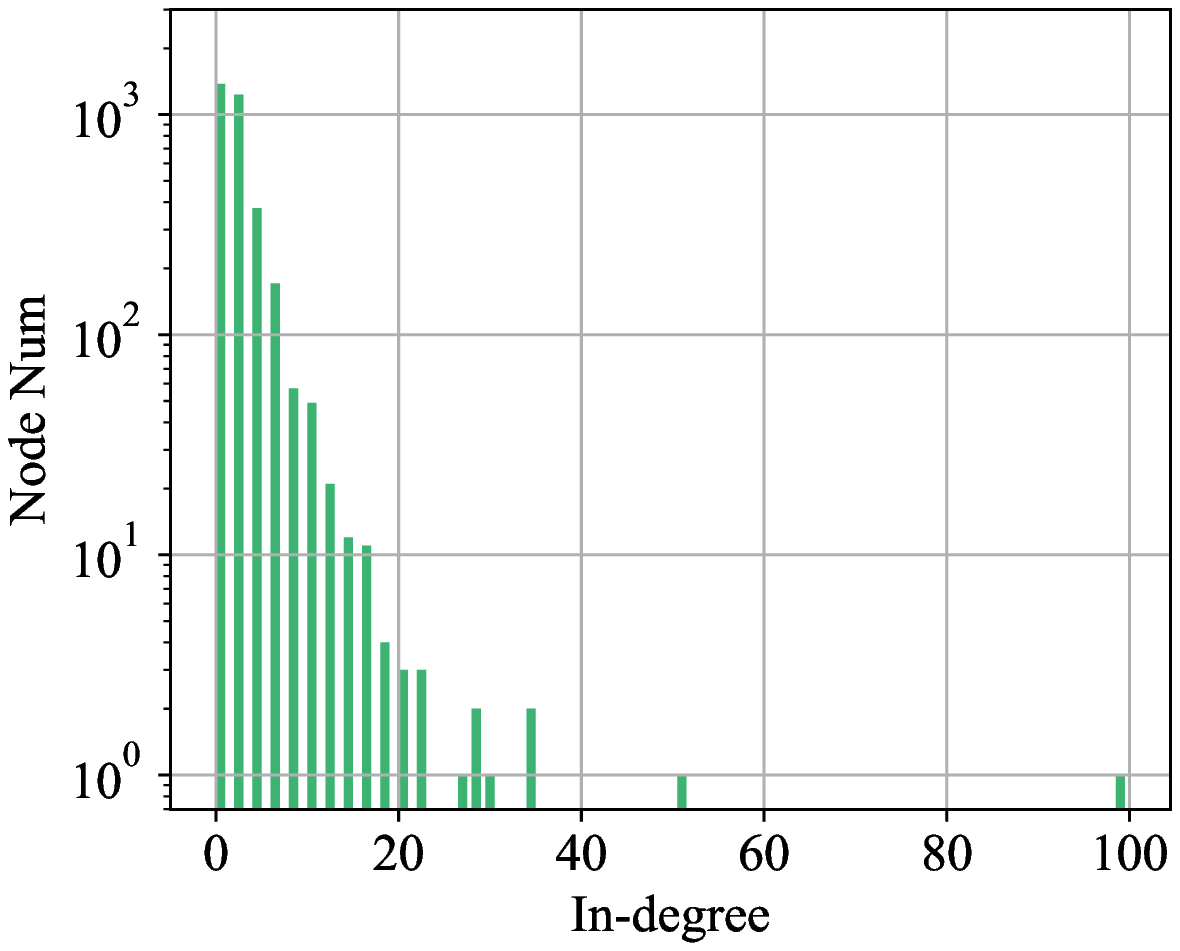}
         % \caption{GIN-CiteSeer}
       %   \label{gcn_cifar_2}
      \end{minipage}
   }

   \subfigure[PubMed]{
      \begin{minipage}[t]{0.4\textwidth}
         \centering
         \includegraphics[scale=0.38]{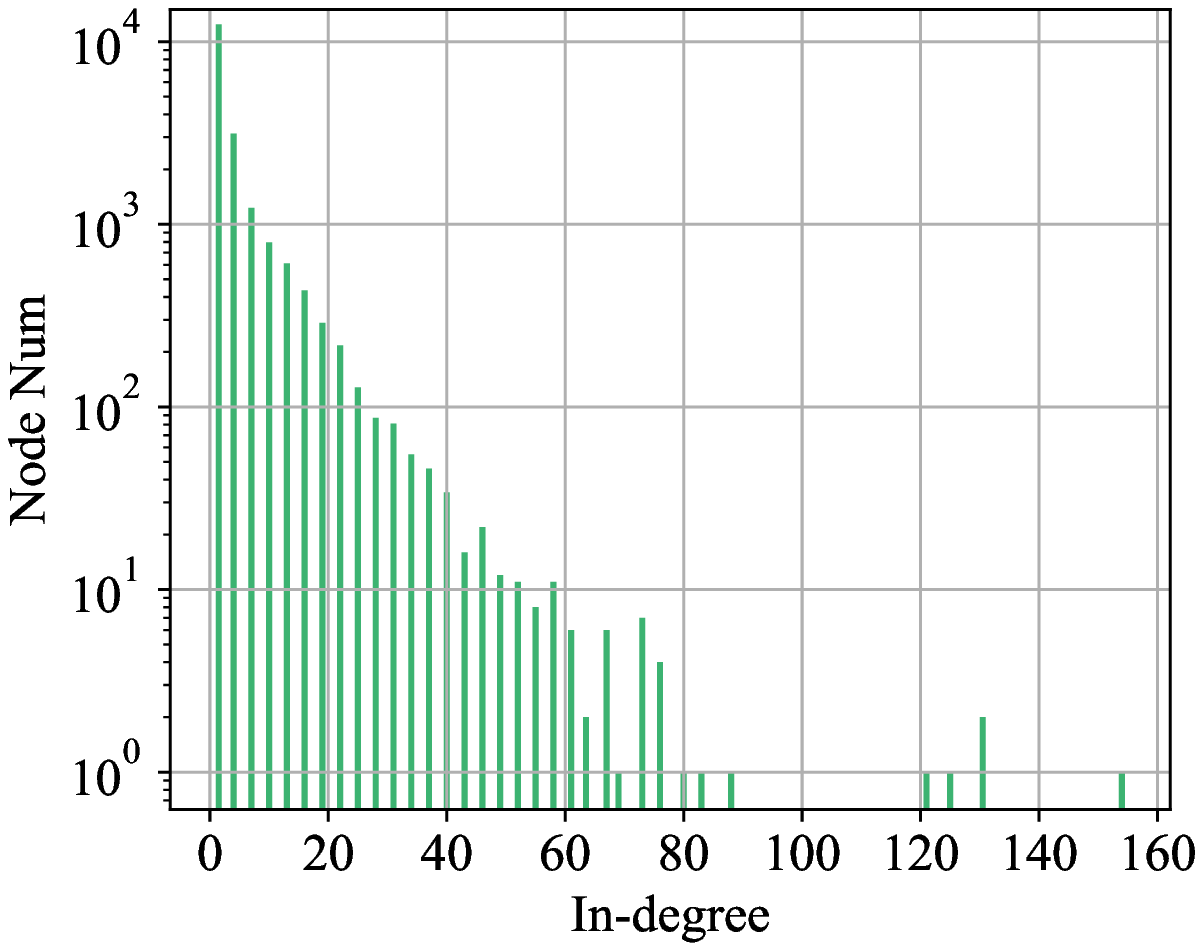}
         % \caption{GAT-CiteSeer}
       %   \label{gcn_cifar_3}
      \end{minipage}
   }
   \subfigure[ogbn-arxiv]{
      \begin{minipage}[t]{0.4\textwidth}
         \centering
         \includegraphics[scale=0.38]{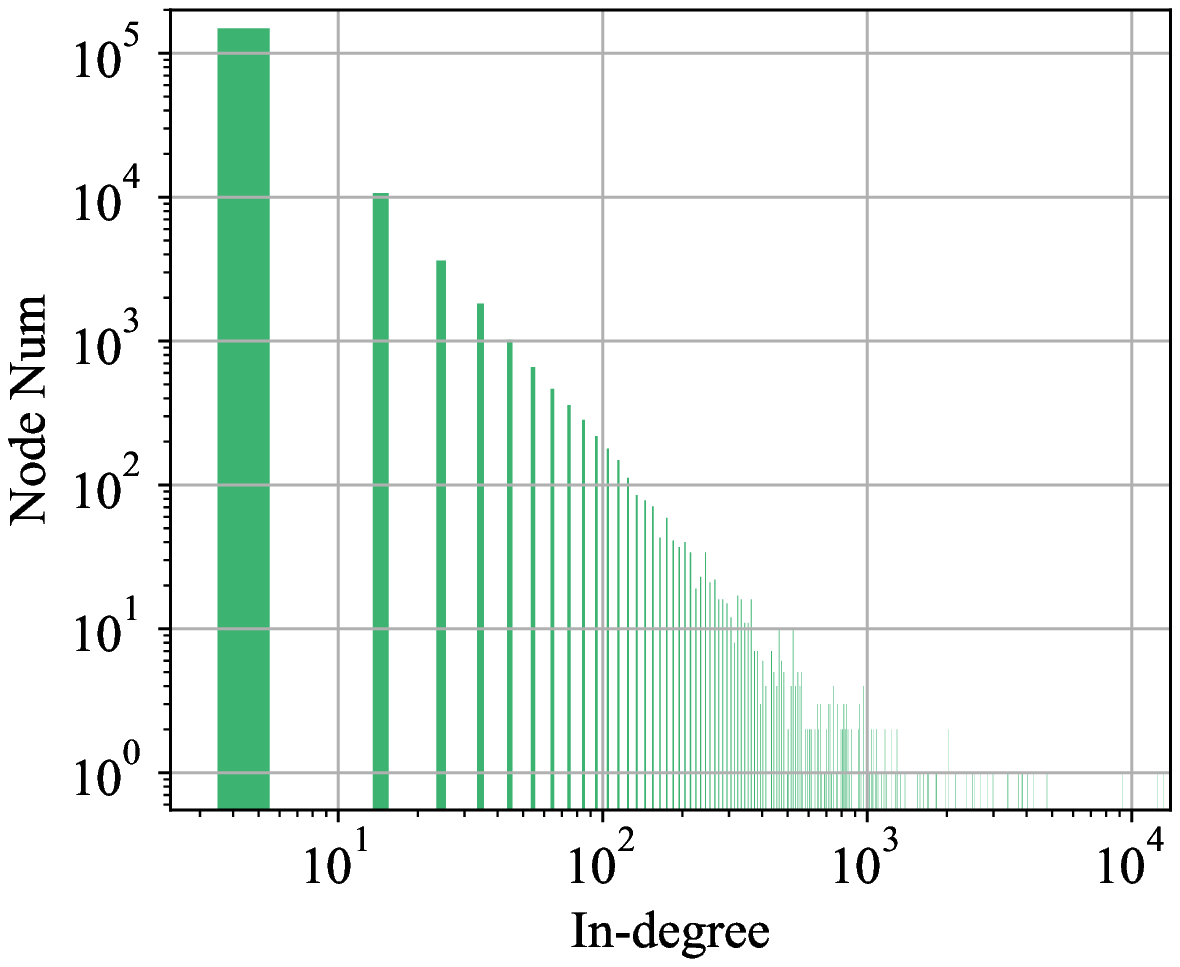}
         % \caption{GAT-CiteSeer}
       %   \label{gcn_cifar_4}
      \end{minipage}
   }

   \subfigure[MNIST]{
      \begin{minipage}[t]{0.4\textwidth}
         \centering
         \includegraphics[scale=0.38]{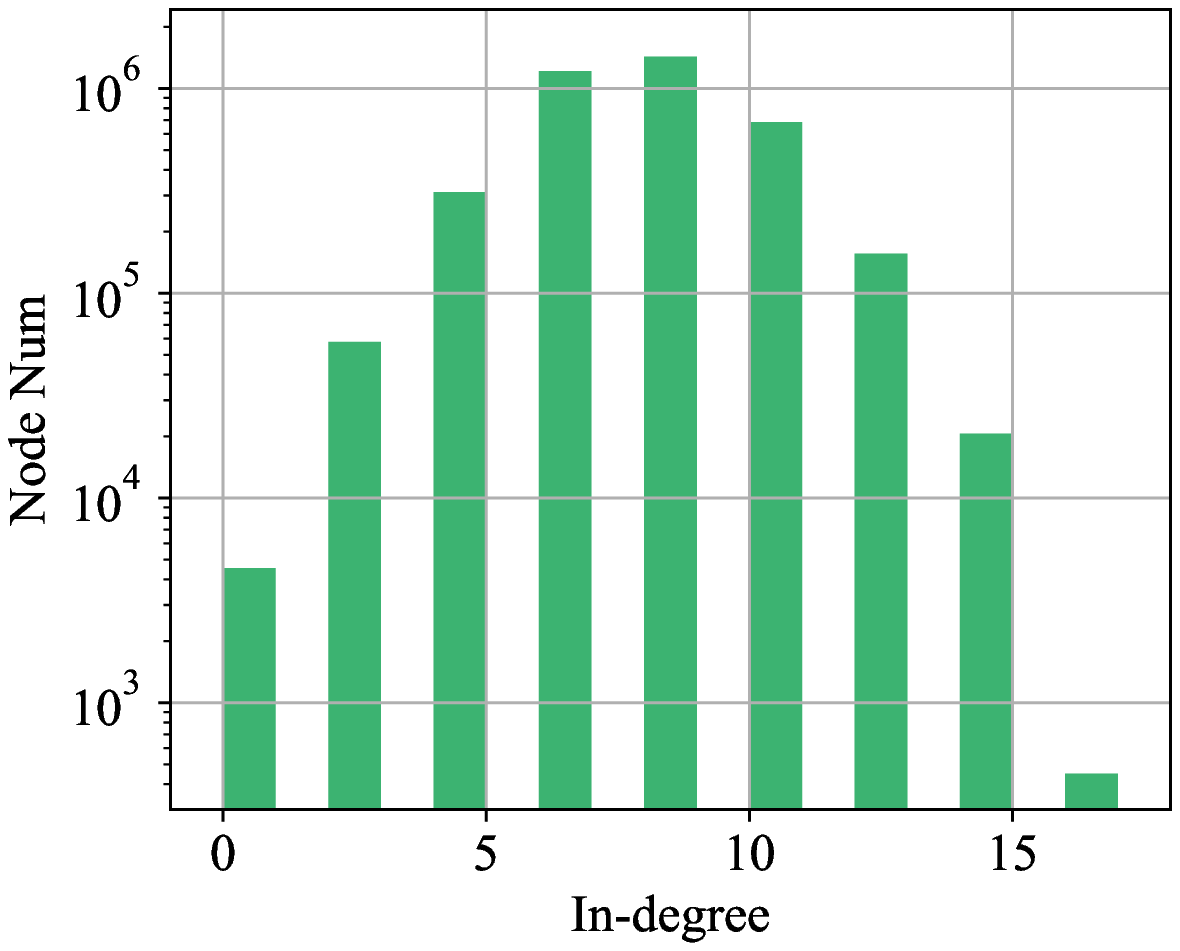}
         % \caption{GCN-CiteSeer}
       %   \label{gcn_cifar_1}
      \end{minipage}
   }
   \subfigure[CIFAR10]{
      \begin{minipage}[t]{0.4\textwidth}
         \centering
         \includegraphics[scale=0.38]{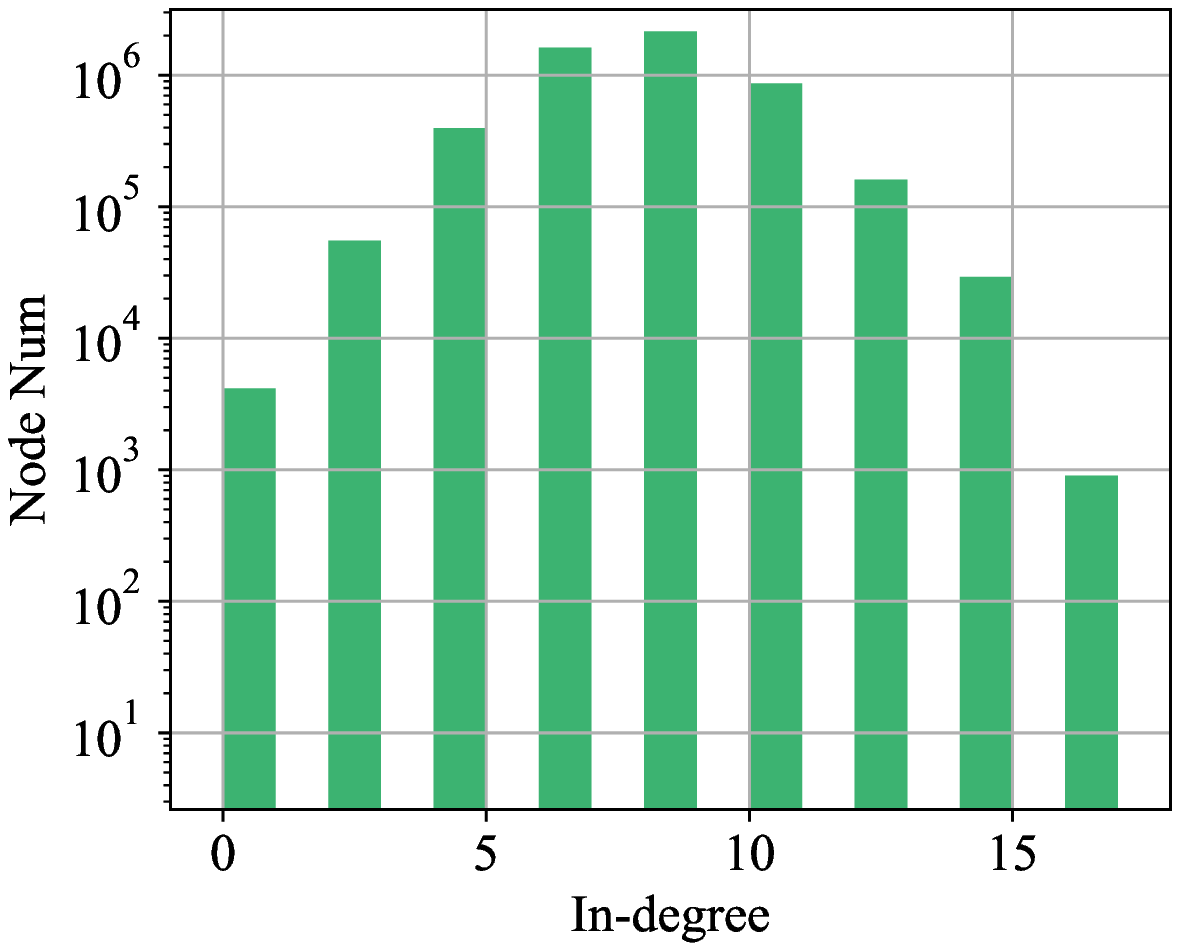}
         % \caption{GIN-CiteSeer}
       %   \label{gcn_cifar_2}
      \end{minipage}
   }
  
   \subfigure[REDDIT-BINARY]{
      \begin{minipage}[t]{0.4\textwidth}
         \centering
         \includegraphics[scale=0.38]{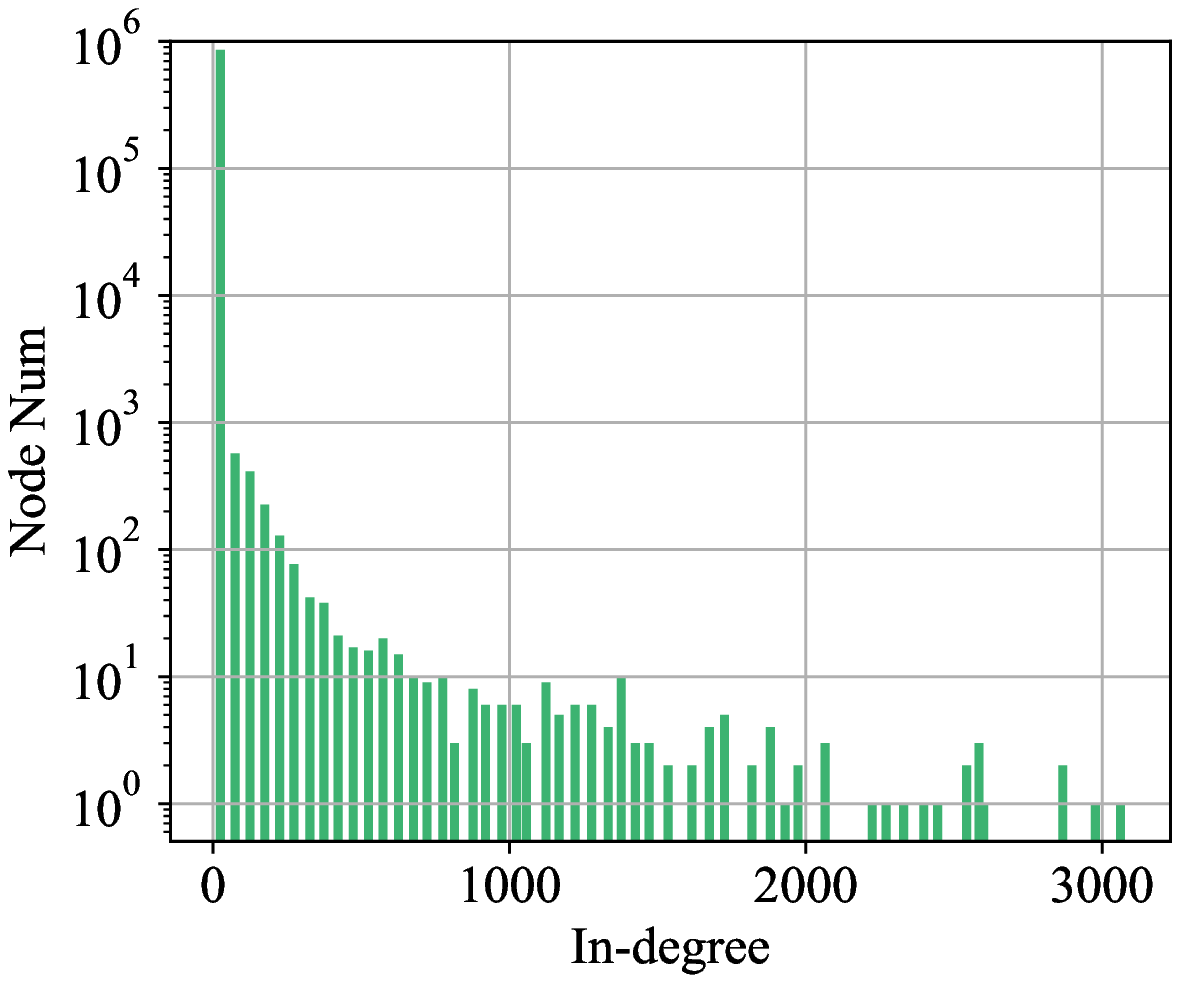}
         % \caption{GAT-CiteSeer}
       %   \label{gcn_cifar_3}
      \end{minipage}
   }
   \subfigure[ZINC]{
      \begin{minipage}[t]{0.4\textwidth}
         \centering
         \includegraphics[scale=0.38]{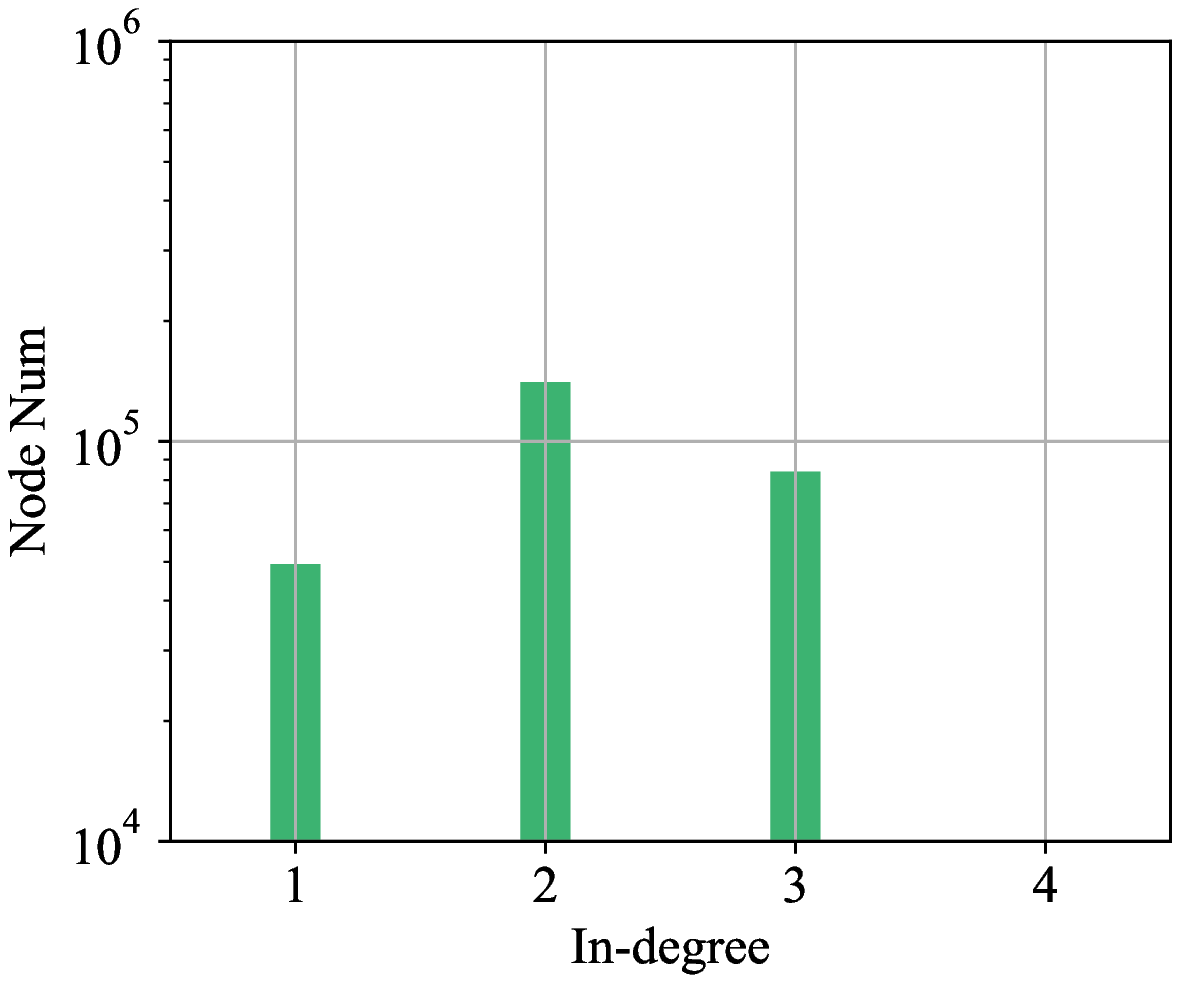}
         % \caption{GAT-CiteSeer}
       %   \label{gcn_cifar_4}
      \end{minipage}
   }
   \caption{The in-degree distribution for each dataset used in this work.}
   \label{deg_node}
\end{figure}

We show the statistics for each dataset used in our work in Table \ref{dataset_sta}.
For datasets in node-level tasks, nodes correspond to documents and edges to citations 
between them. Node
features are a bag-of-words representation of the document. The target is to classify each node in
the graph correctly. The \textbf{Cora}, \textbf{CiteSeer} and \textbf{PubMed} are 
from \citet{yang2016revisiting}. 
The \textbf{ogbn-arxiv}, \textbf{ogbl-mag} and \textbf{ogbn-collab} are from \citet{hu2020open}. 
The \textbf{Flickr} is from \citet{zeng2019graphsaint}. The \textbf{Reddit} 
is from \citet{hamilton2017inductive}.
In graph-level tasks, \textbf{REDDIT-BINARY} \citep{yanardag2015deep} 
is a balanced dataset
where each graph corresponds to an online discussion thread
and the nodes correspond to users. 
There would be an edge between two nodes if at least one of them responded to another's comment.
The task is then to identify
whether a given graph belongs to a question/answer-based
community or a discussion-based community
The \textbf{MNIST} and \textbf{CIFAR-10} datasets \citep{dwivedi2020benchmarking} 
which are often 
used for image classification tasks 
are transformed into graphs in which every
node is represented by their superpixel and location, and the edges are constructed by \citet{achanta2012slic}. 
The task is to classify the
image using its graph representation. The \textbf{ZINC} \cite{gomez2018automatic} 
dataset contains graphs representing
molecules, where each node is an atom. The task is to regress the penalized $\rm logP$ 
(also called constrained solubility in some works) of a given graph.
In Figure \ref{deg_node}, we show the in-degree distribution for 
all the datasets we use in our paper.
\linespread{1.32}
\begin{table}[h]
    \caption{The statistics for each dataset used in this work.}
   \label{dataset_sta}
   \begin{center}
    \begin{tabular}{ccccccc}
        \hline \toprule[2pt]
        Task                         & Name          & Graphs & Nodes       & Edges       & Features & Classes \\ \midrule[1pt]
        \multirow{8}{*}{Node-level}  & Cora          & 1      & 2708        & 10556       & 1433     & 7       \\
                                        & CiteSeer      & 1      & 3327        & 9104        & 3703     & 6       \\
                                        & PubMed        & 1      & 19717       & 88648       & 500      & 3       \\ 
                                        & ogbn-arxiv     & 1      & 169343      & 1166243     & 128      & 23      \\
                                        & ogbn-mag      & 1      & 1939743        & 25582108    & 128     & 349       \\
                                        & ogbl-collab      & 1      & 235868        & 1285465        & 128     & --       \\
                                        & Reddit      & 1      & 232965        & 11606919        & 602     & 41       \\
                                        & Flickr         & 1      & 89250        & 899756        & 500     & 7       \\\hline
        \multirow{4}{*}{Graph-level} & REDDIT-BINARY & 2000   & $\sim$429.6 & $\sim$995.5 & 0        & 2       \\
                                        & MNIST         & 70000  & $\sim$71    & $\sim$565   & 3        & 10      \\
                                        & CIFAR10       & 60000  & $\sim$117.6 & $\sim$941.2 & 5        & 10      \\
                                        & ZINC          & 12000  & $\sim$23    & $\sim$49.8  & 28       & ---     \\ \bottomrule[2pt]
    \end{tabular}
    \end{center}
\end{table}

\begin{figure}[htbp]
   \centering
   \includegraphics[scale=0.4]{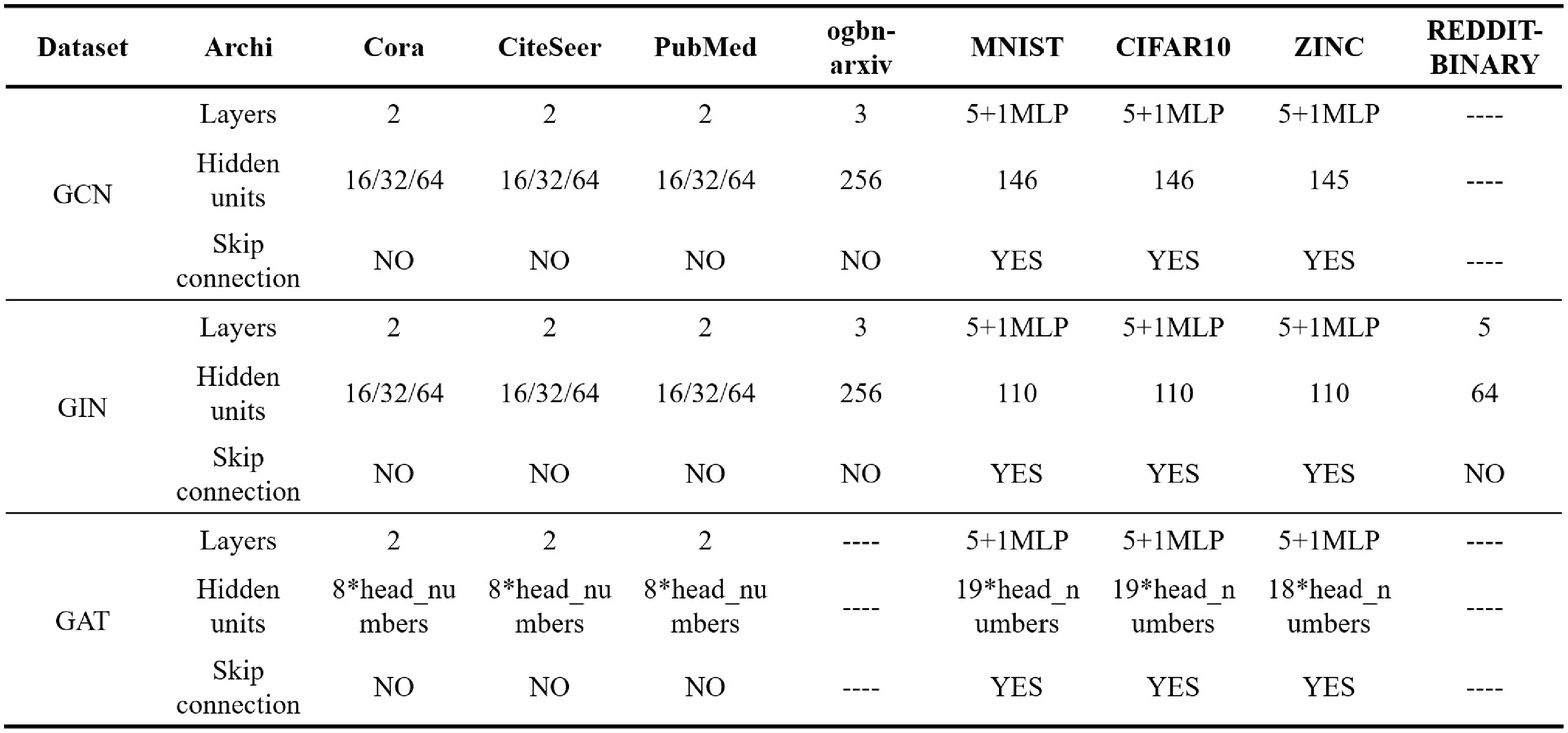}
   \caption{The model architectures used in our evaluations, the head number for all GAT 
   models on different
   tasks are 8.}
   \label{model_information}
\end{figure} 

\subsection{Experimental Setup}
\label{experimental setup}
To make a fair comparison, we adopt the same GNN architectures as \citet{tailor2020degree} 
on every task, and the FP32 baseline
is also the same. For those tasks that \citet{tailor2020degree} does not do, we 
adopt the same architecture as their FP32 version.
For \textbf{ogbn-arxiv} and \textbf{PubMed}, we use 
the architectures and FP32 results reported by \citet{hu2020open} and \citet{kipf2016semi} 
respectively. We use standard splits for MNIST, CIFAR-10, 
and ZINC \citep{dwivedi2020benchmarking}. For \textbf{Cora}, \textbf{CiteSeer} and \textbf{PubMed}, we
use the splits used by \citet{yang2016revisiting}. 
For REDDIT-BINARY, we use 10-fold cross-validation. 
Our data split way is also the same as DQ-INT4. 

Figure \ref{model_information} shows the architectures of the models used in our 
evaluations,
including the layers, the number of hidden units, and whether to use a skip connection.

Our method is implemented using PyTorch Geometric \citep{fey2019fast}. 
We quantize the same parts as the DQ-INT4 in all models except for the normalized adjacency matrix in the GCN model, which we have
proven that the quantization of this matrix is not necessary in Appendix \ref{proof}, \textbf{proof 2.}.
The values in the \textbf{Cora} and \textbf{CiteSeer} are all 0 or 1, therefore, we do not quantize the input 
features for the first layer of the GNNs trained on the two datasets as DQ.
For all quantized GNNs, we train them by Adam optimizer. 
% For \textbf{Cora} and \textbf{CiteSeer}, the values in the datasets are all 0 or 1, therefore, we do not quantize 
% the input features for the first layer of GNNs.    
The learning rate and the learning rate schedule are consistent with their FP32 version. 
In our method, the quantization parameters $(s,b)$ are also learnable, so we 
set the learning rate for them, including the $b$ for features, $s$ for features, 
and $s$ for weights. 
% The learning rate schedule
% is the same for all learnable parameters. 

When initializing, the parameters of the models are initialized as 
their FP32 version, the quantization bits for
all nodes and weight matrixes are initialized by 4bits, 
and the step sizes for node features and weights are initialized by 
$s\in \mathcal{N} (0.01,0.01)$ except for the graph-level tasks on 
GAT, where we initialize the step size 
by $s\in \mathcal{U} (0,1)$. The $ \mathcal{N} $ is 
normal distribution and the $ \mathcal{U} $ is uniform distribution.
And for GAT model trained 
on graph-level datasets, we just learn the quantization bits of the node features, 
while in the attention coefficients computation part, we use the exact 4bit to quantize.
The batch size is 128 in all graph-level tasks. 
The results reported in our work for GNNs on \textbf{Cora}, \textbf{CiteSeer} and 
\textbf{PubMed}
are averaged over 100 runs with different seeds, 
and the results for \textbf{ogbn-arxiv}, \textbf{MNIST}, \textbf{CIFAR-10} and \textbf{ZINC} 
are
averaged over ten runs. 
% with different seeds, but the initial seed is 41. 
The results on \textbf{REDDIT-BINARY} are obtained by 10-fold cross-validation and the 
split seed is 12345, which is the same as DQ-INT4.   
All experiments in our paper ran on RTX 2080Ti GPU driven by Ubuntu 18.04. 
The version of the CUDA and Pytorch are 10.2 and 1.8.0, respectively.

\subsubsection{Experimental setups for the ablation study.}
\label{setup for abalation}

\textbf{The advantage of learning-based mixed-precision quantization:} 
During the experiment of 
comparing the learning bitwidth and bit assignment, we ensure 
the average bits of node features of these two methods are comparable
to verify
the effectiveness of our $\rm A^2Q$ method. 
As an example, if the average bit is 2.2bit when assigning 
the bit to nodes with different in-degrees, we will first sort the 
nodes by their in-degrees and then select the nodes with the top 20\% in-degrees, 
and quantize those by 3bit, and for the remaining nodes, we use 2bit to \
quantize. In the model trained by the bit assignment method, 
the bit is not learnable, and other 
hyperparameters are all consistent with the model using the $\rm A^2Q$
method.
For the ``GCN-Cora-mixed-precision'' and ``GIN-CiteSeer-mixed-precision'' tasks, 
we use 3bit and 5bit to quantize
the GNNs while keeping the average bitwidth at 4bits. 
In particular, we assign 5bits to those nodes with 50\% top 
in-degrees and assign 3bits to others.

\textbf{The power of learning the quantization parameters:} For the 
``no-lr-bit'', we initialize
the bitwidth as 4bits for all nodes features and just train the step size. 
For the ``no-lr-step'',
we initialize the step size as previously mentioned but do not train them.
For the ``no-lr'', we just initialize the bitwidth and the step size, but do not train them.

\textbf{Local Gradient v.s. Global Gradient:} All settings of the model trained 
by global gradient is consistent with the model trained by 
local gradient 
method.

\textbf{The overhead of Nearest Neighbor Strategy:} 
The model, without using the Nearest Neighbor Strategy, selects the quantization parameters 
according
to their in-degrees. Every in-degree has a corresponding group of quantization parameters. 
Those nodes whose in-degrees are larger than 1000 will share the same 
group quantization
parameters. In this way, The quantization parameters used by the nodes features can be 
determined as soon as the graph data is available, without the need for 
selection during the inference process,
and then we can compare the overhead introduced by the selection process.

\begin{table}[t]
   \caption{The results comparison on GCN-PubMed and GIN-ogbn-arxiv.}
   \label{appen_node-level-results}
   \begin{center}
   \begin{tabular}{clllll}
   
   \hline \toprule[2pt]
                                          & \multicolumn{1}{c}{} & Accuracy             & Average bits                                            & Compression Ratio                                            & Speedup                                             \\  \midrule[1pt]
   \multirow{3}{*}{\textbf{PubMed}}       & GCN(FP32)            & 78.9±0.7\%           & 32                                                      & 1x                                                           & --- \\          
                                          & GCN(DQ)              & 62.5±2.4\%           & 4                                                       & 8x                                                           & 1x  \\
                                          & GCN(ours)            & \textbf{77.5±0.1\%}  & \textbf{1.90}                                           & \textbf{16.8x}                                               & \textbf{1.45x} \\ \hline
   \multirow{3}{*}{\textbf{ogbn-arxiv}}   & GIN(FP32)            & 68.8±0.2\%           & 32                                                      & 1x                                                           & --- \\
                                          & GIN(DQ)              & 57.6±2.2\%           & 4                                                       & 8x                                                           & 1x  \\
                                          & GIN(ours)            & \textbf{65.2±0.4\%}  & \textbf{3.82}                                           & \textbf{8.4x}                                               & \textbf{1.02x} \\ \bottomrule[2pt]
   
   \end{tabular}
   \end{center}
   \end{table}

\subsection{More Experiments Results}
\label{more_results}
This section is a complementary part about experiments results to demonstrate that our $\rm A^2Q$ 
quantization method is general and robust.
\linespread{1}
\begin{table}[htbp]
   \caption[]{The results comparison on inductive learning tasks and more graphs.}
   \label{inductive-results}
   \begin{center}
      \begin{tabular}{cccc}
         \hline \toprule[2pt]
         Task                                                                               & Acc(\%)          & Average   bits & Compression   Ratio \\  \midrule[1pt]
         \multirow{2}{*}{GCN-mag}                                                           & 30.8±0.1(FP32)   & 32             & 1x                  \\
                                                                                             & 32.7±0.4(Ours)   & 2.7            & 11.7x               \\ \hline
         \multirow{2}{*}{GCN-collab}                                                          & 44.8±1.1(FP32)  & 32             & 1x                  \\
                                                                                             & 44.9±1.5(Ours)   & 2.5            & 12.7x               \\ \hline
         \multirow{2}{*}{\begin{tabular}[c]{@{}c@{}}GraphSage-\\ REDDIT\end{tabular}} & 95.2±0.1(FP32)   & 32             & 1x                  \\
                                                                                             & 95.3±0.1(Ours)   & 3.9            & 8.1x                \\ \hline
         \multirow{2}{*}{\begin{tabular}[c]{@{}c@{}}GraphSage-\\ Flickr\end{tabular}} & 50.9±1.0(FP32)   & 32             & 1x                  \\
                                                                                             & 50.0±0.5\%(Ours) & 3.8            & 8.4x          \\ \bottomrule[2pt]     
         \end{tabular}
   \end{center}
\end{table}
\begin{table}[H]
   \caption[]{Comparison with more quantization method.}
   \label{more_related_works}
   \begin{center}
      \begin{tabular}{cccc}
         \hline \toprule[2pt]
         Task                              & Acc(\%)         & Average   Bits & Compression   Ratio \\ \midrule[1pt]
         \multirow{2}{*}{GCN-Cora}         & 80.9±0.0(Half-pre) & 16             & 1x                  \\
                                           & 80.9±0.6(Ours)  & 1.7            & 9.40x               \\ \hline
         \multirow{2}{*}{GAT-CiteSeer}     & 68.0±0.1(LPGNAS) & 8              & 1x                  \\
                                           & 71.9±0.7(Ours)  & 1.9            & 4.21x               \\ \hline
         \multirow{2}{*}{GraphSage-Cora}   & 74.3±0.1(LPGNAS) & 12             & 1x                  \\
                                           & 74.5±0.2(Ours)  & 2.7            & 4.44x               \\ \hline
         \multirow{2}{*}{\begin{tabular}[c]{@{}c@{}}GraphSage-\\ Flickr\end{tabular}} & 49.7±0.3(LPGNAS) & 8              & 1x                  \\
                                           & 50.0±0.5(Ours)  & 3.8            & 2.11x               \\ \bottomrule[2pt]
         \end{tabular}
   \end{center}
   
   \end{table}
\begin{figure}[H]
\subfigure[GCN-Cora]{
    \begin{minipage}[t]{0.31\textwidth}
        \centering
        \includegraphics[width=1\linewidth,height=0.85\linewidth]{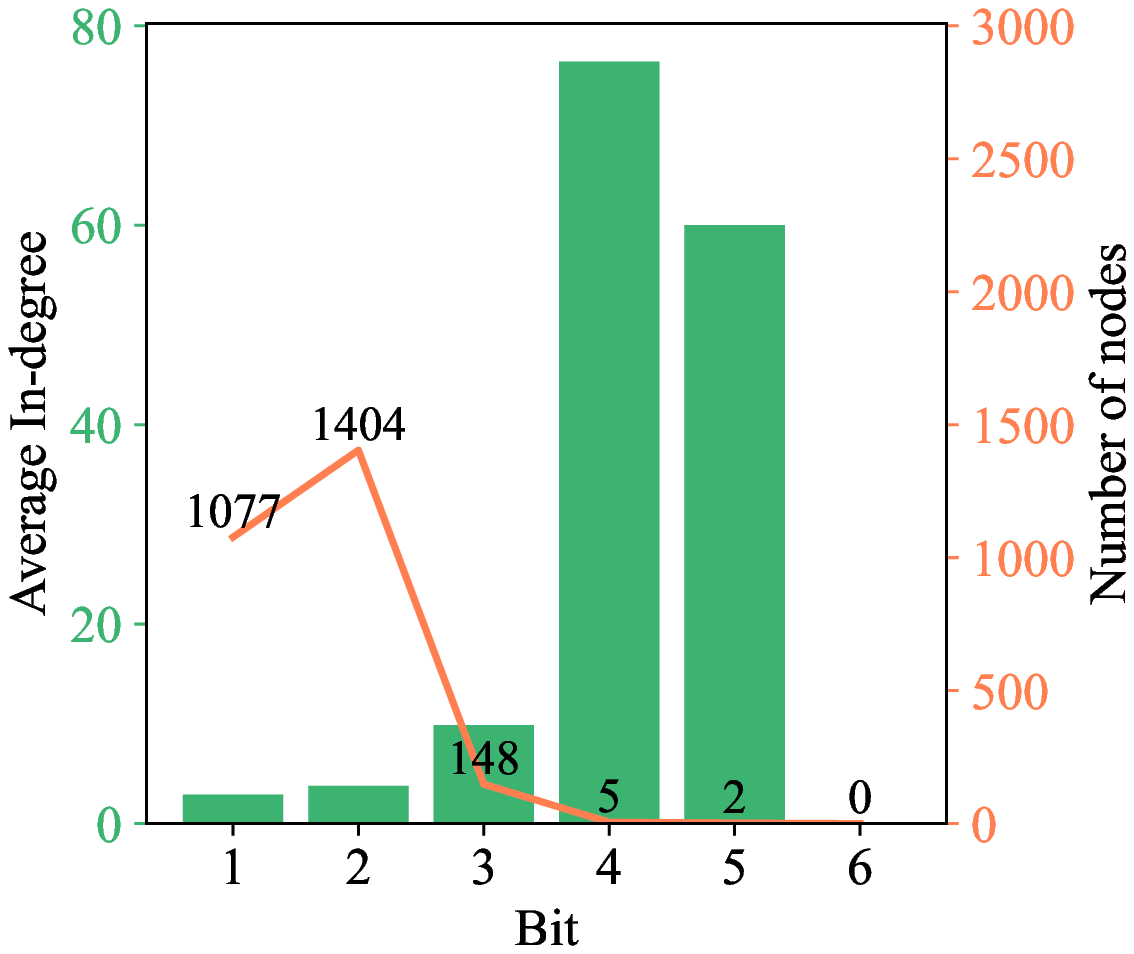}
        % \caption{GCN-CiteSeer}
        \label{gcn_cora_bit_deg}
    \end{minipage}
}
\subfigure[GIN-Cora]{
    \begin{minipage}[t]{0.31\textwidth}
        \centering
        \includegraphics[width=1\linewidth,height=0.85\linewidth]{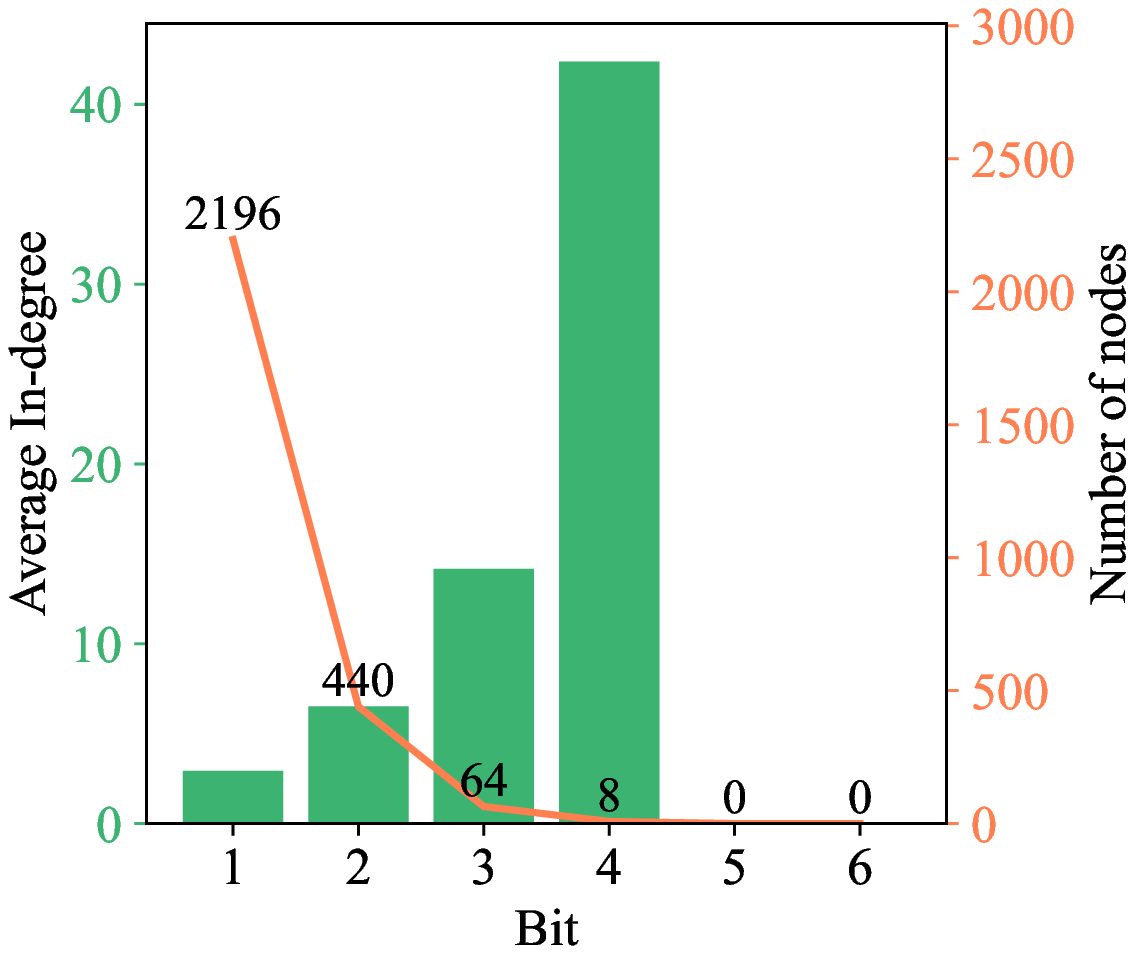}
        % \caption{GIN-CiteSeer}
        \label{gin_cora_bit_deg}
    \end{minipage}
}
\subfigure[GAT-Cora]{
    \begin{minipage}[t]{0.31\textwidth}
        \centering
        \includegraphics[width=1\linewidth,height=0.85\linewidth]{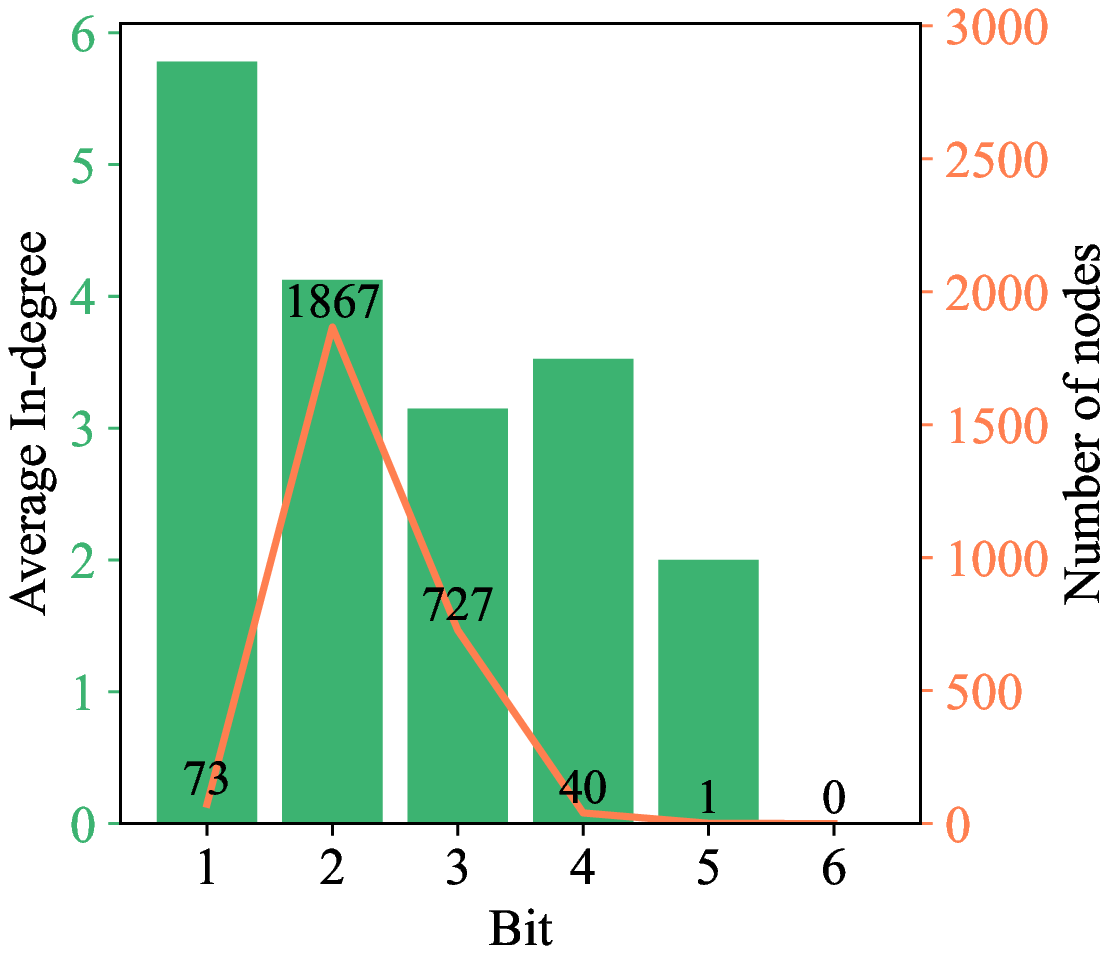}
        % \caption{GAT-CiteSeer}
        \label{gat_cora_bit_deg}
    \end{minipage}
}

\subfigure[GCN-PubMed]{
       \begin{minipage}[t]{0.23\linewidth}
          \centering
          \includegraphics[width=1\linewidth,height=1\linewidth]{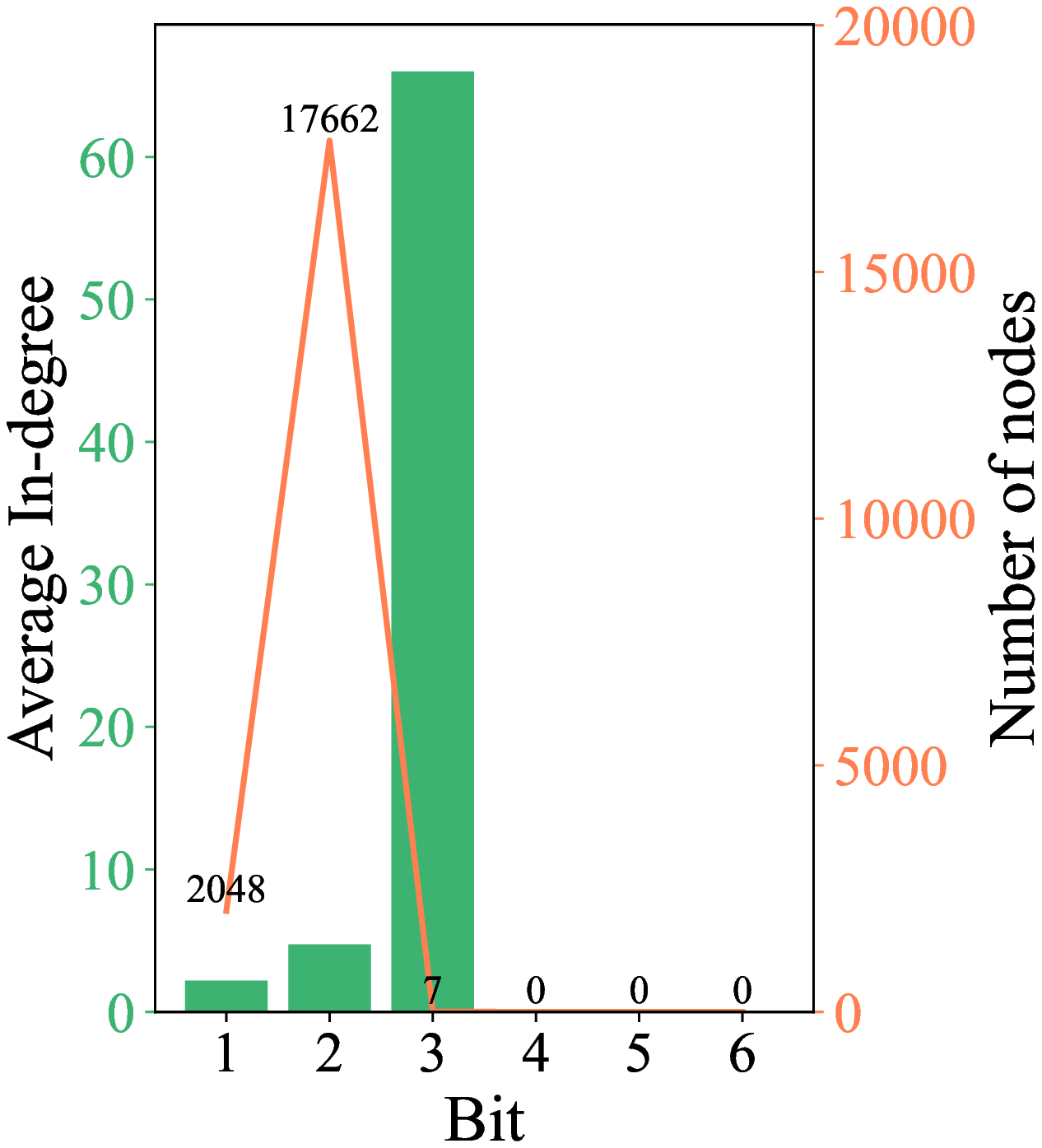}
          % \caption{GCN-CiteSeer}
          \label{gcn_pub_bit_deg}
       \end{minipage}
    }
\subfigure[GAT-PubMed]{
   \begin{minipage}[t]{0.23\linewidth}
      \centering
      \includegraphics[width=1\linewidth,height=1\linewidth]{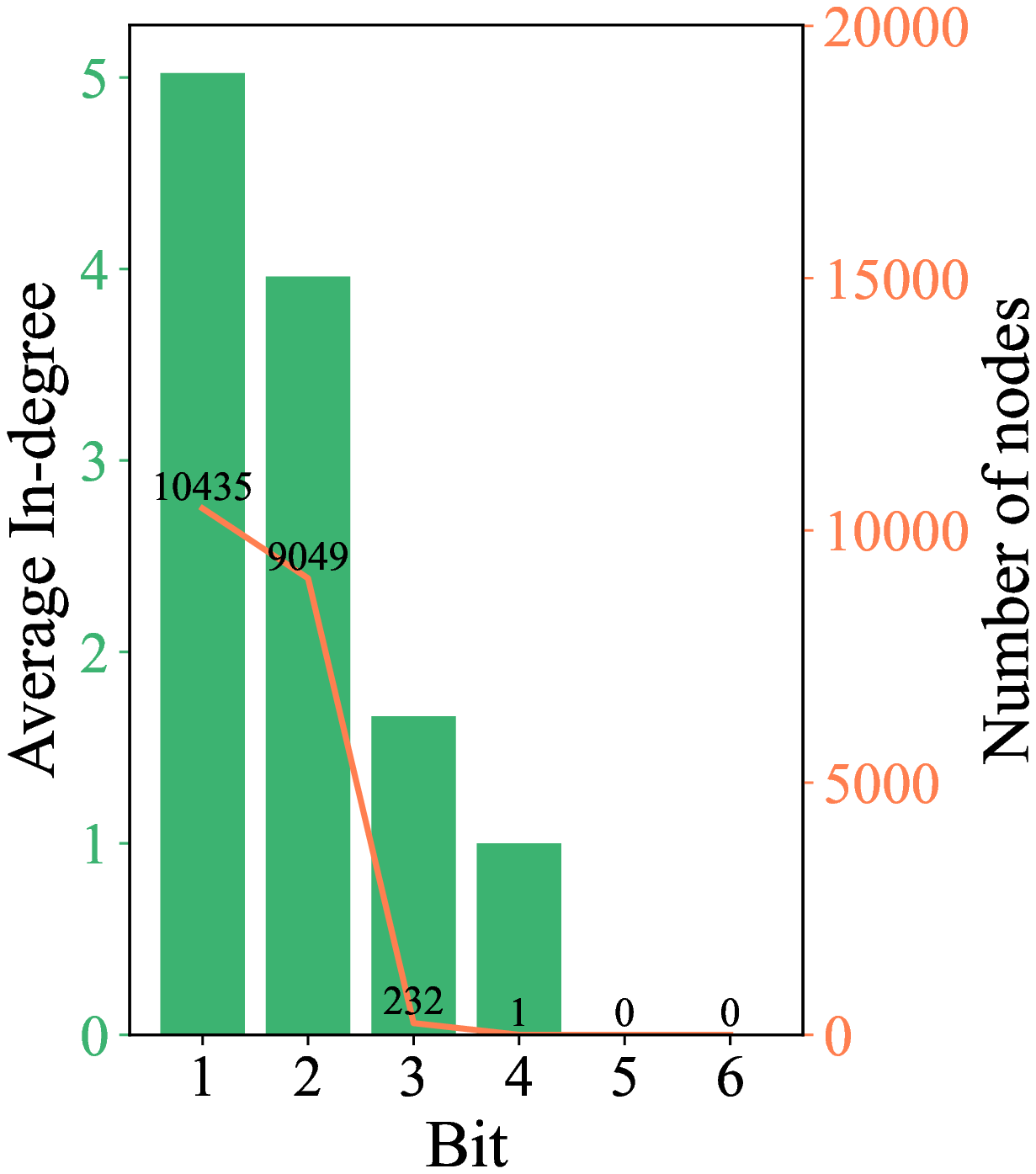}
      % \caption{GAT-CiteSeer}
      \label{gat_pub_bit_deg}
   \end{minipage}
}
\subfigure[GCN-ogbn-arxiv]{
       \begin{minipage}[t]{0.23\linewidth}
          \centering
          \includegraphics[width=1\linewidth,height=1\linewidth]{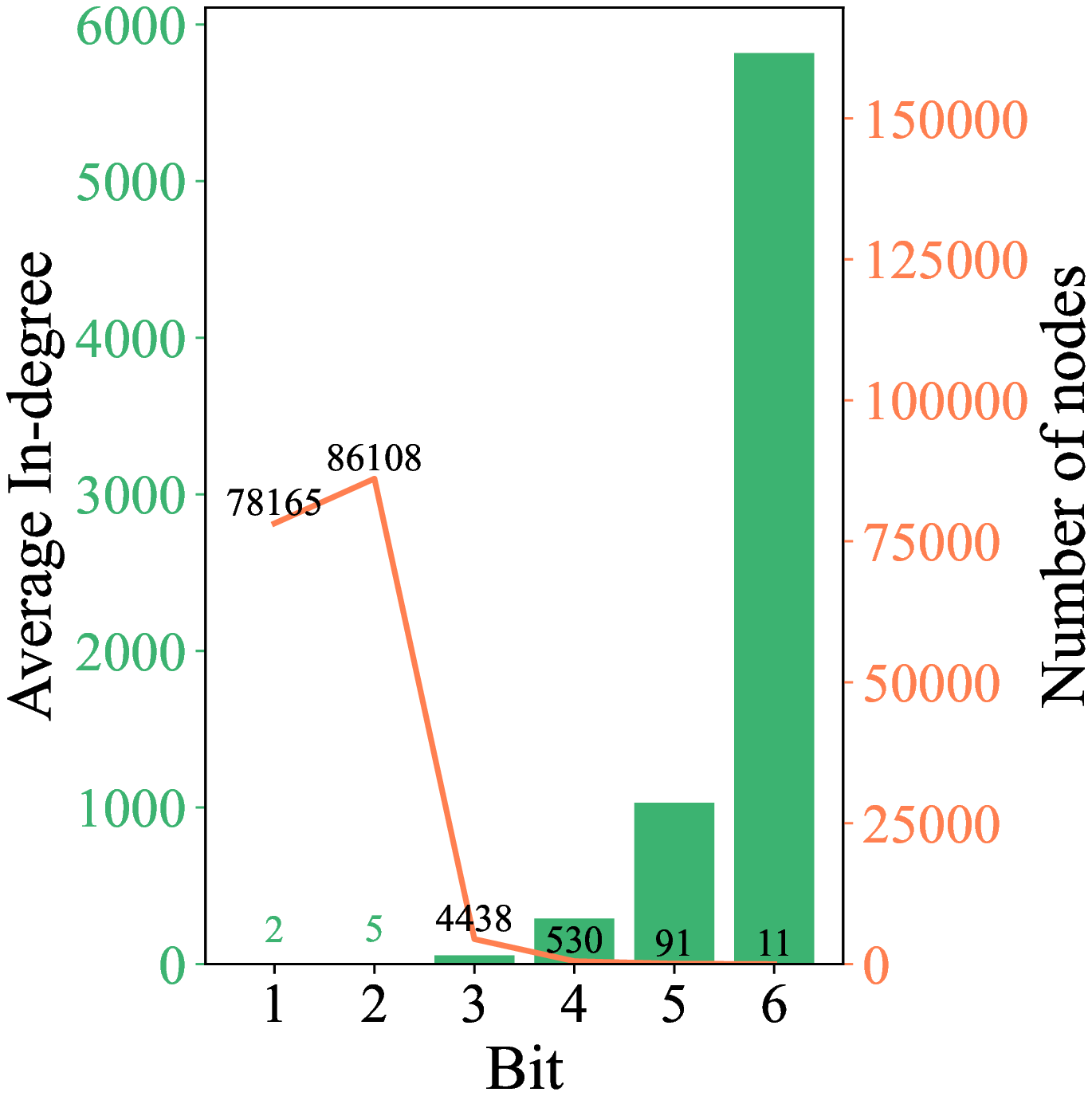}
          % \caption{GCN-CiteSeer
          \label{gcn_arxiv_bit_deg}
       \end{minipage}
    }
    \subfigure[GIN-ogbn-arxiv]{
      \begin{minipage}[t]{0.23\linewidth}
         \centering
         \includegraphics[width=1\linewidth,height=1\linewidth]{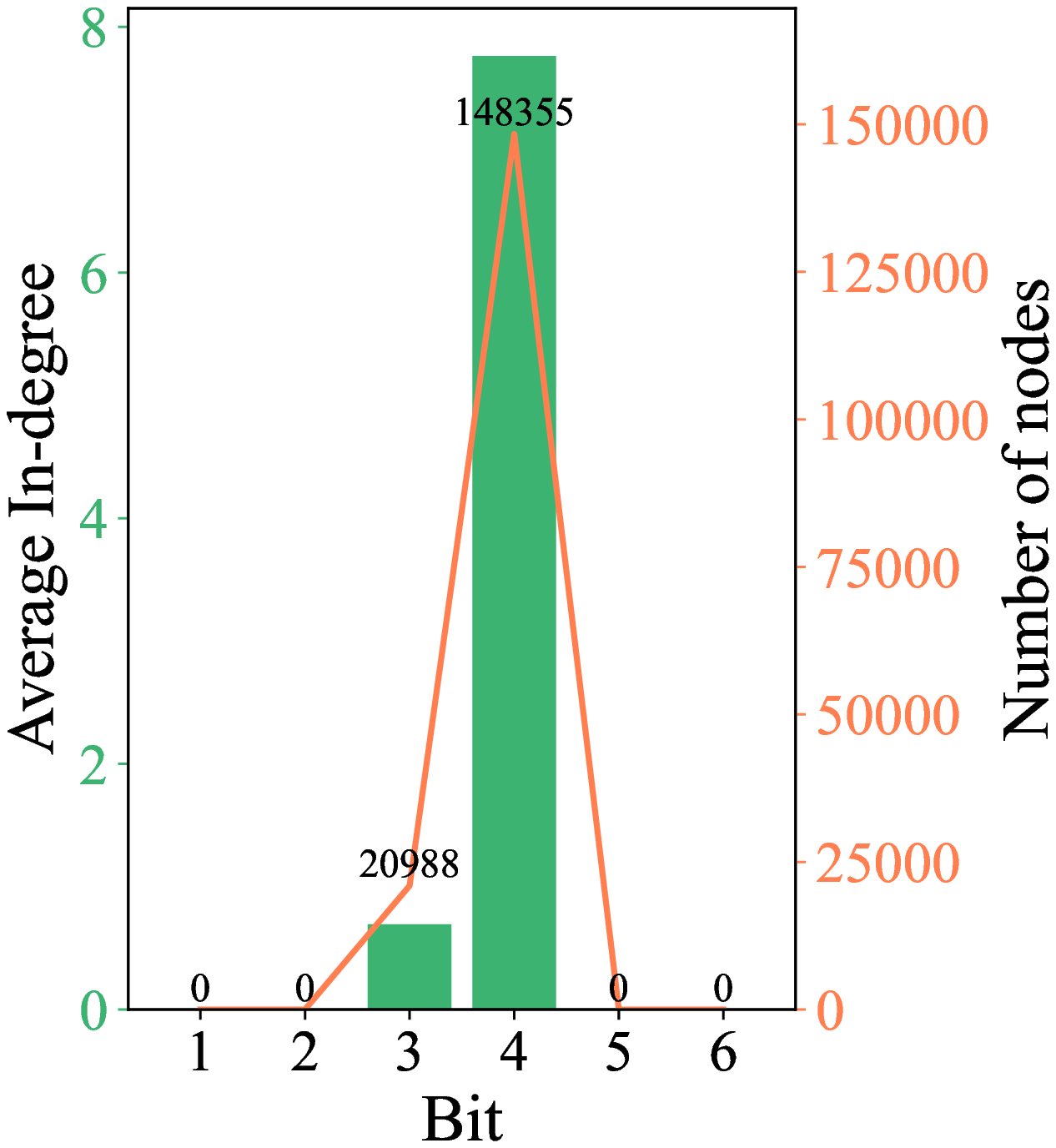}
         % \caption{GIN-CiteSeer}
         \label{gin_pub_bit_deg}
      \end{minipage}
   }
\caption{The relationship between bit and average in-degrees of nodes using the corresponding 
    bitwidth to quantize. 
    (a), (b) and (c) Three GNN models trained on Cora. 
    (d) and (e) GCN and GAT trained on PubMed, respectively. (f) and (g) 
    GCN and GIN trained on ogbn-arxiv, respectively.}
\label{cora_bit_deg}
\end{figure}

\linespread{1}
\begin{table}[H]
   \caption{The effect of \#m on the accuracy of quantized model, using GIN trained on REDDIT-BINARY as 
   an example. The average bitwidth is 4bits.}
   \label{the effect of m}
   \begin{center}
   \begin{tabular}{lllllll}
   \toprule[2pt]
   \multicolumn{3}{c}{GIN(FP32): 92.2± 2.3\%} & \multicolumn{4}{c}{GIN(DQ): 81.3± 4.4\%}                      \\ \midrule[1pt]
   \textbf{m}           & 100           & 400          & \multicolumn{2}{l}{800}        & 1000          & 1500         \\
   \textbf{Accuracy}    & 88.7±3.5\%    & 90.6±3.8\%   & \multicolumn{2}{l}{92.0±2.2\%} & 92.5±1.8\%    & 92.6±1.9\%   \\ \bottomrule[2pt]
   \end{tabular}
   \end{center}
\end{table}
\subsubsection{Node-Level Tasks}
\label{appendix_node_level_results}
In Table \ref{appen_node-level-results}, we show more task results on \textbf{PubMed} and \textbf{ogbn-arxiv}. On the GAT-ogbn-arxiv task, our GPU raised the 
Out Of Memory error, so we do not report the results on the GAT-ogbn-arxiv task.     
The model quantized by our $\rm A^2Q$ method is also
significantly better than DQ-INT4, which shows that our $\rm A^2Q$ is general. 
We do not compare with DQ-INT8 because our results are comparable with the FP32 
baseline with a much larger compression ratio than 
DQ-INT8.

We also show the relationship between bit and average in-degrees of nodes using the corresponding bitwidth to quantize 
on more tasks in Figure \ref{cora_bit_deg}. 
We present the results of the final layer of GNNs. 
The results show that the bitwidth learned by our $\rm A^2Q$ method is also aggregation-aware, 
which means that our method is robust.

We also evaluate 
the inductive model, GraphSage, on some other node-level tasks to 
demonstrate the generality of our method on inductive learning tasks. 
Due to the sampling operation in the GraphSage model, the subgraph input to the model varies, we 
apply our nearest neighbor strategy to these tasks, i.e., GraphSage-Flickr 
and GraphSage-Reddit. In addition, we evaluate our method on more datasets, 
such as the ogbn-mag and ogbl-collab.
ogbn-mag is a heterogeneous graph and the ogbl-collab is used for the link prediction tasks.

The results of our experiments are 
presented in Table \ref{inductive-results}, where we can see that our approach still 
works well and even brings some generalization performance 
improvement while significantly compressing the model size. This also
demonstrates that our Neighbor Nearest Strategy 
generalizes well on inductive models for node-level tasks.

We also compare with more quantization methods on GNNs. \citet{zhao2020learned} uses the
Network Architecture Search (NAS) to search for the best quantization strategy 
for different components
in the GNNs. \citet{brennan2020not} explore the use of half-precision (i.e., FP16) in 
the forward and backward passes of GNNs. Table \ref{more_related_works} presents
the comparison results on various tasks with these two methods. `Half-pre' denotes the method
in \citet{brennan2020not}, and `LPGNAS' denotes the method in \citet{zhao2020learned}.
The results demonstrate that 
our method achieves better accuracy with a smaller quantization bitwidth on all tasks. 

\subsubsection{Graph-Level Tasks}
\label{appendix_graph_level_results}
We propose the Nearest Neighbor Strategy to quantize the node features in graph-level tasks, in which the number of nodes 
input to models is various.
In our Nearest Neighbor Strategy, $\#m$ groups quantization parameters $(s,b)$ 
should be initialized, 
and we explore the effect of the value of m on the performance of 
the quantized model in Table \ref{the effect of m} using the GIN trained on REDDIT-BINARY dataset. 
We can observe that when the value of $m$ is smaller than 800, the accuracy increases
as the value of $m$ increases. When the value of $m$ is higher than 800, the performances of the models 
with different $m$ 
are similar. However, the models
with a larger $m$ are more stable. 

Moreover, the selection of $m$ may be related to the number of nodes input to the model. 
According to our experiments, we finally select $m$ as 1000 for all graph-level tasks.

Table \ref{appen-zinc-results} lists the comparison results on GIN-ZINC and GAT-ZINC. On the regression tasks,
our method is also significantly better than DQ-INT4. Notably, we do not learn different bitwidths for the 
nodes in ZINC datasets due to the similar topology structure between nodes.

\begin{figure}[H]
   \centering
   \subfigure[1-st layer]{
      \begin{minipage}[t]{0.23\textwidth}
         \centering
         \includegraphics[width=1\linewidth,height=0.75\linewidth]{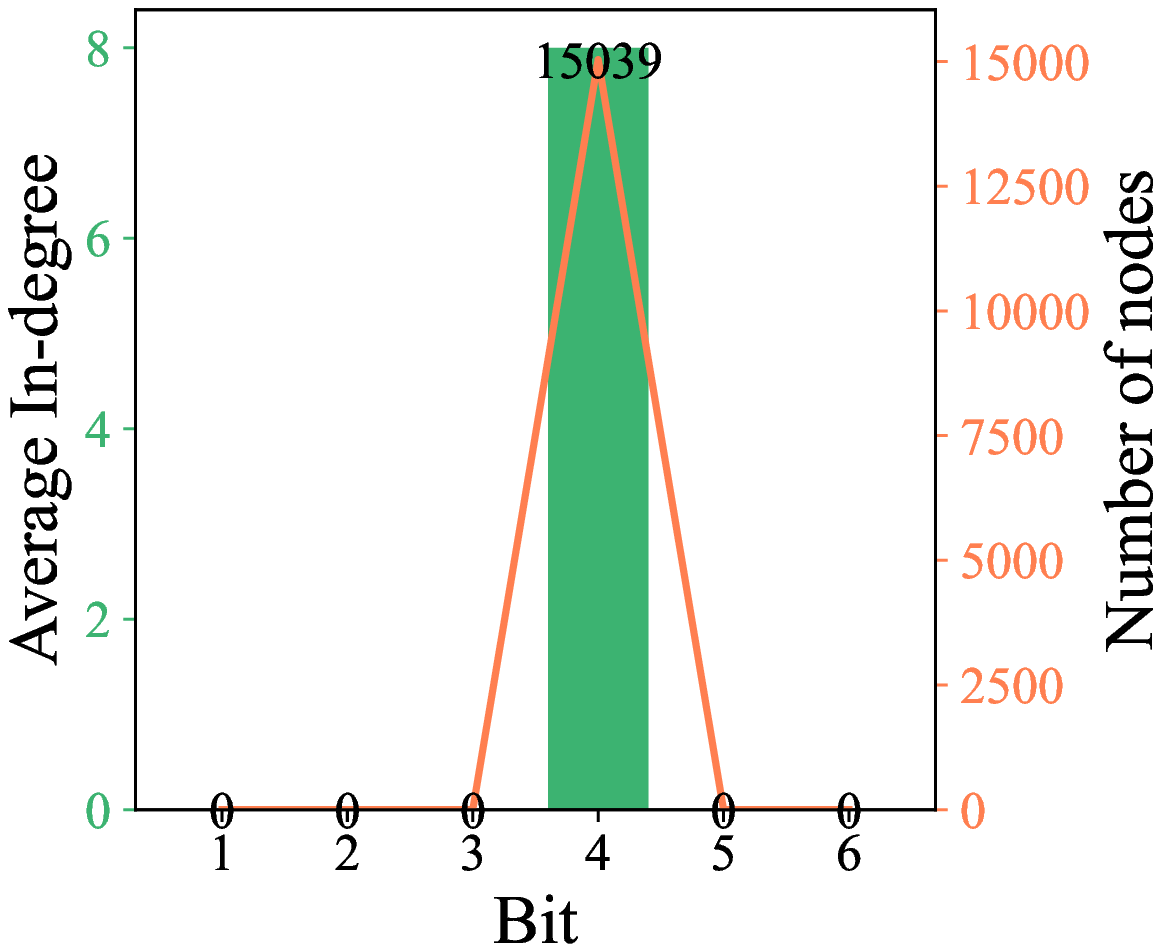}
         % \caption{GCN-CiteSeer}
         \label{gat_cifar_1}
      \end{minipage}
   }
   \subfigure[2-nd layer]{
      \begin{minipage}[t]{0.23\textwidth}
         \centering
         \includegraphics[width=1\linewidth,height=0.75\linewidth]{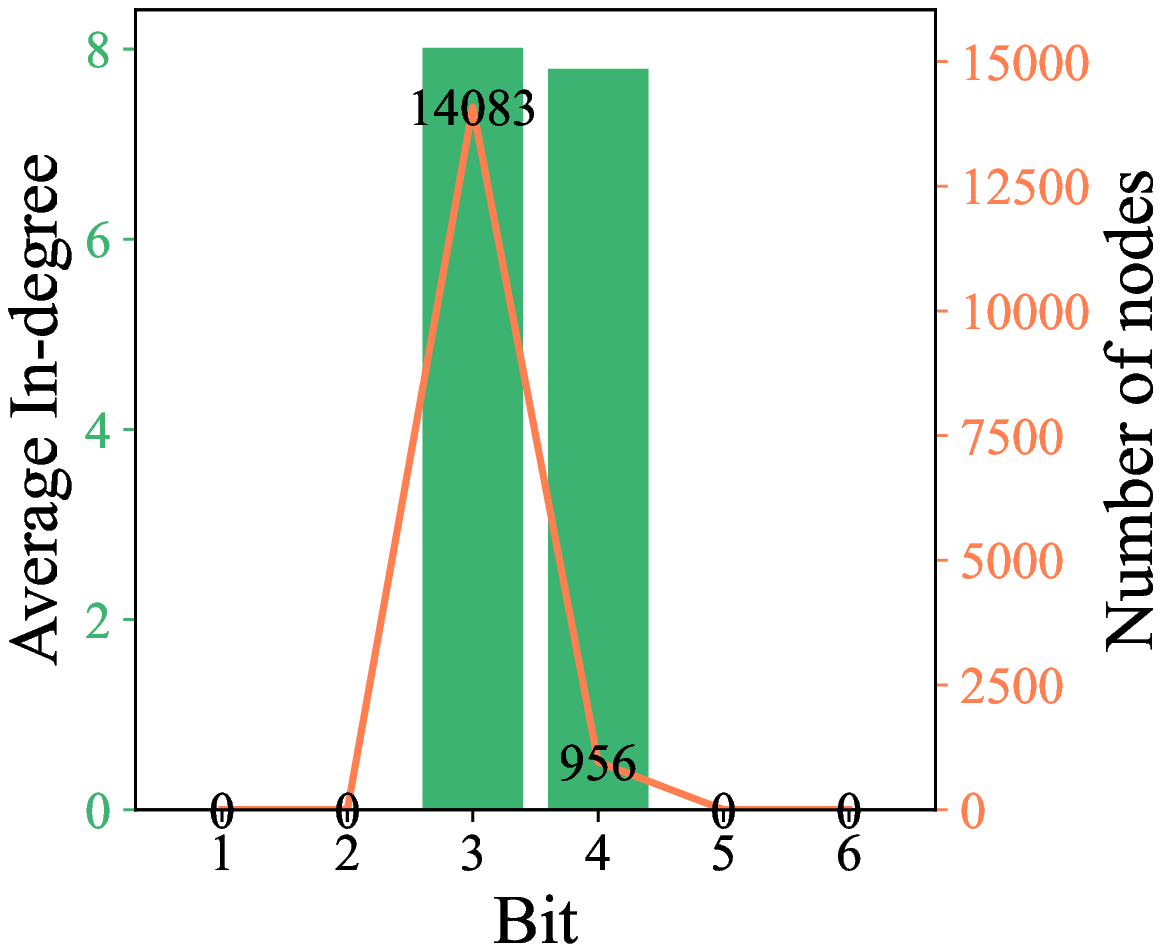}
         % \caption{GIN-CiteSeer}
         \label{gat_cifar_2}
      \end{minipage}
   }
   \subfigure[3-rd layer]{
      \begin{minipage}[t]{0.23\textwidth}
         \centering
         \includegraphics[width=1\linewidth,height=0.75\linewidth]{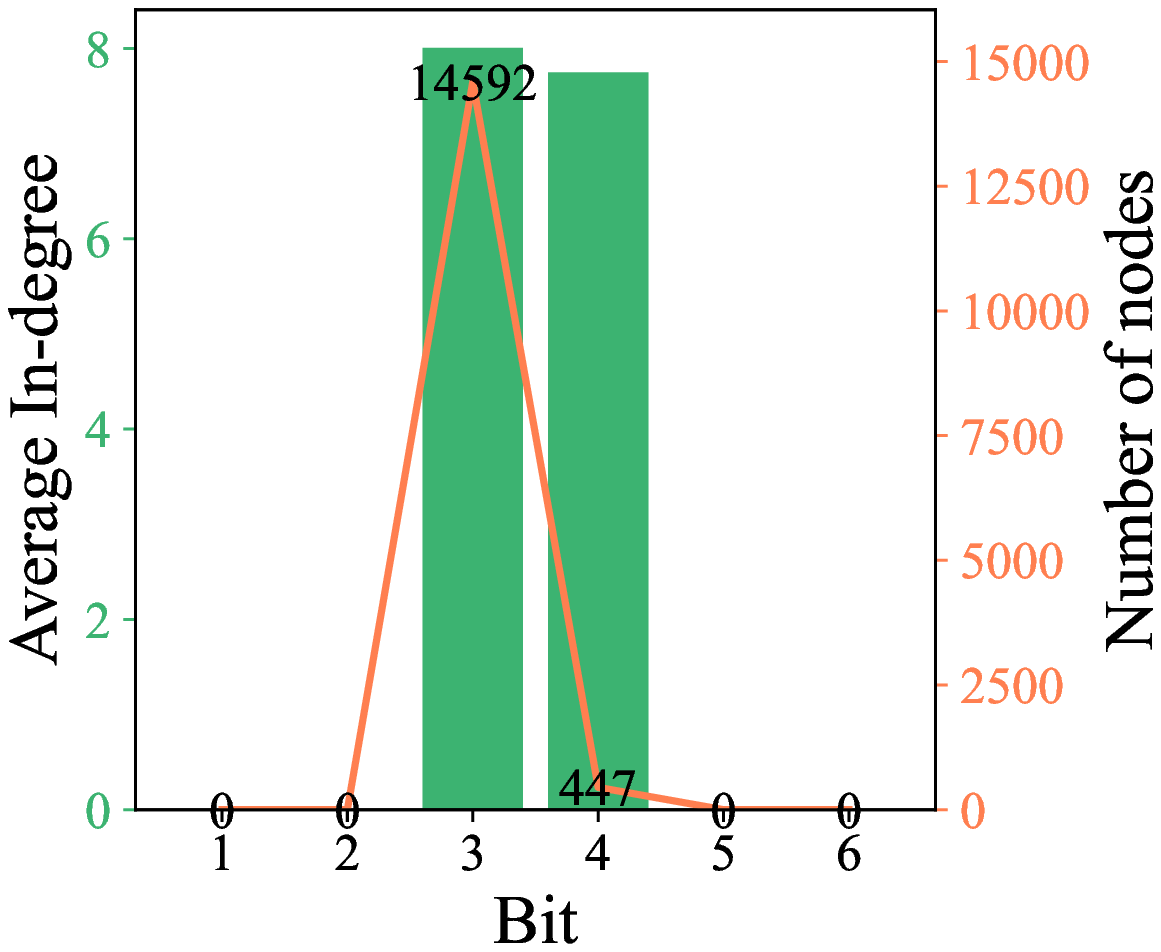}
         % \caption{GAT-CiteSeer}
         \label{gat_cifar_3}
      \end{minipage}
   }
   \subfigure[4-th layer]{
      \begin{minipage}[t]{0.23\textwidth}
         \centering
         \includegraphics[width=1\linewidth,height=0.75\linewidth]{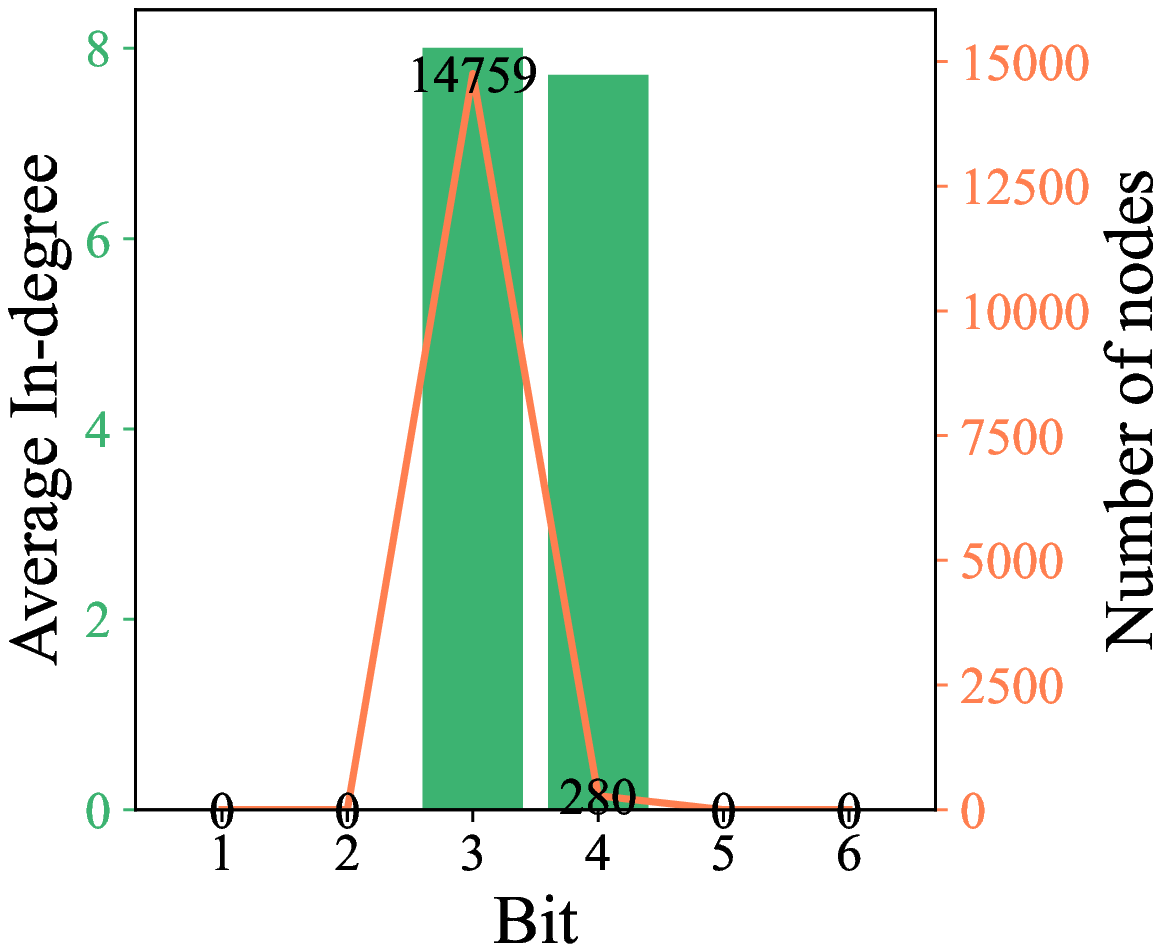}
         % \caption{GAT-CiteSeer}
         \label{gat_cifar_4}
      \end{minipage}
   }
   \caption{The relationship between bit and average in-degrees of nodes using the corresponding bitwidth to quantize 
      on different layers of GAT trained on CIFAR10.}
   \label{gat_cifar_bit_deg}
\end{figure}

\begin{figure}[H]
   \centering
   \subfigure[1-st layer]{
      \begin{minipage}[t]{0.23\textwidth}
         \centering
         \includegraphics[width=1\linewidth,height=0.73\linewidth]{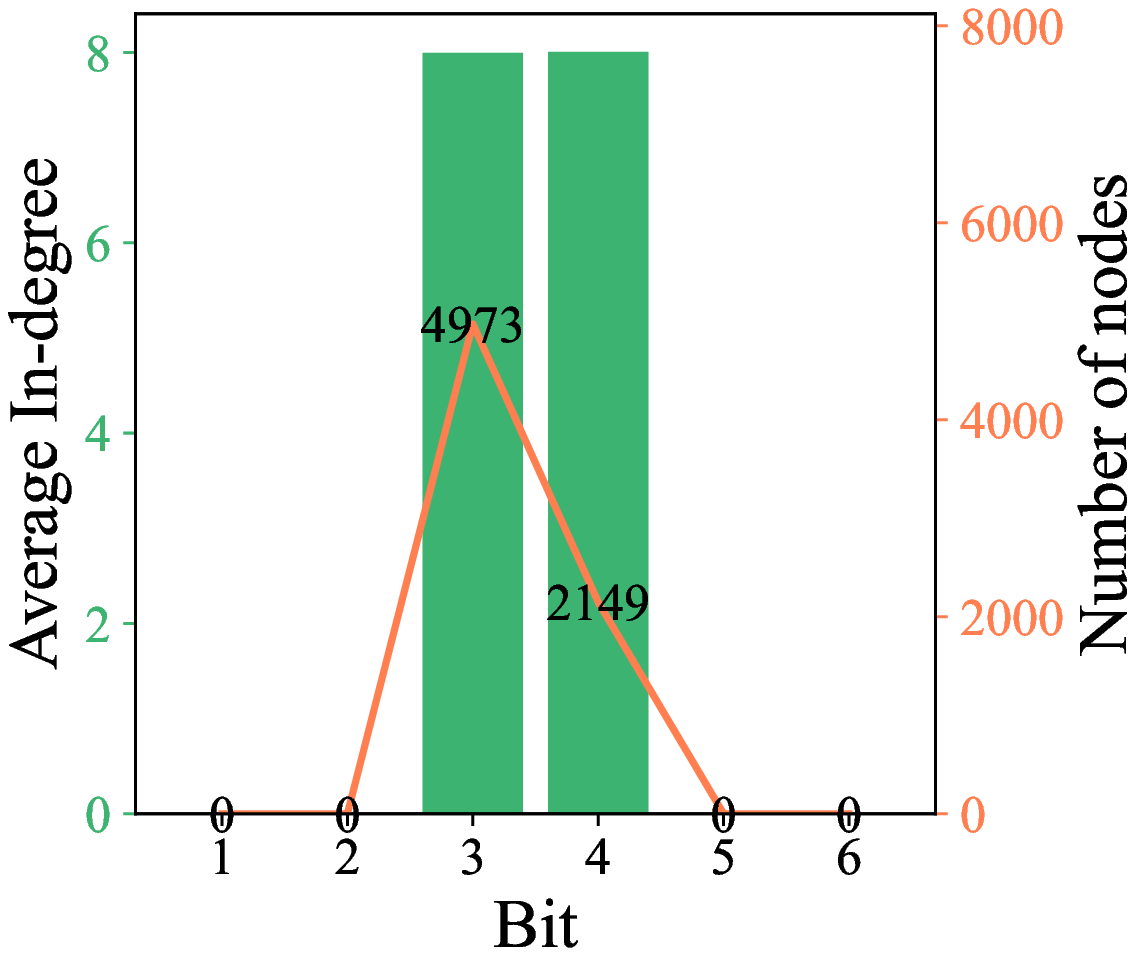}
         % \caption{GCN-CiteSeer}
         \label{gcn_mnist_1}
      \end{minipage}
   }
   \subfigure[2-nd layer]{
      \begin{minipage}[t]{0.23\textwidth}
         \centering
         \includegraphics[width=1\linewidth,height=0.73\linewidth]{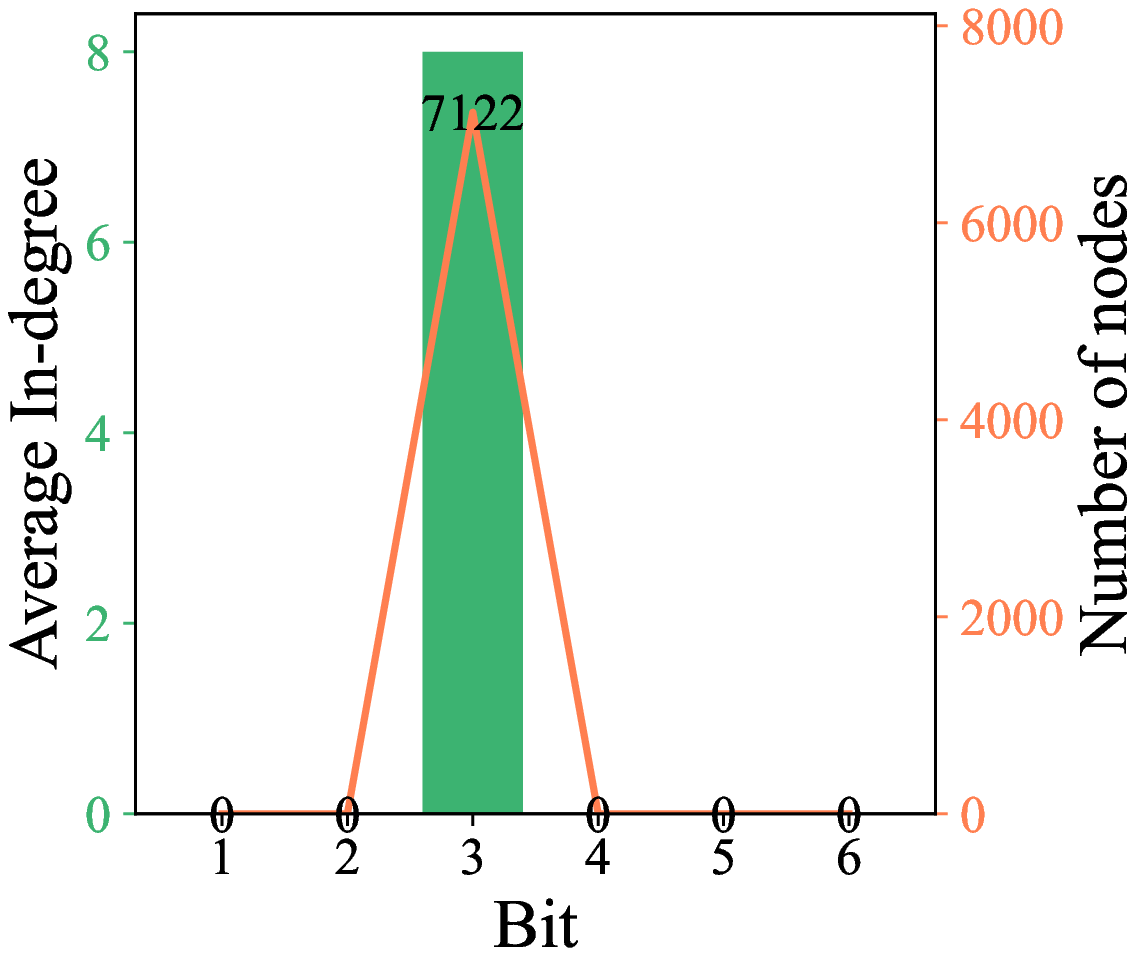}
         % \caption{GIN-CiteSeer}
         \label{gcn_mnist_2}
      \end{minipage}
   }
   \subfigure[3-rd layer]{
      \begin{minipage}[t]{0.23\textwidth}
         \centering
         \includegraphics[width=1\linewidth,height=0.73\linewidth]{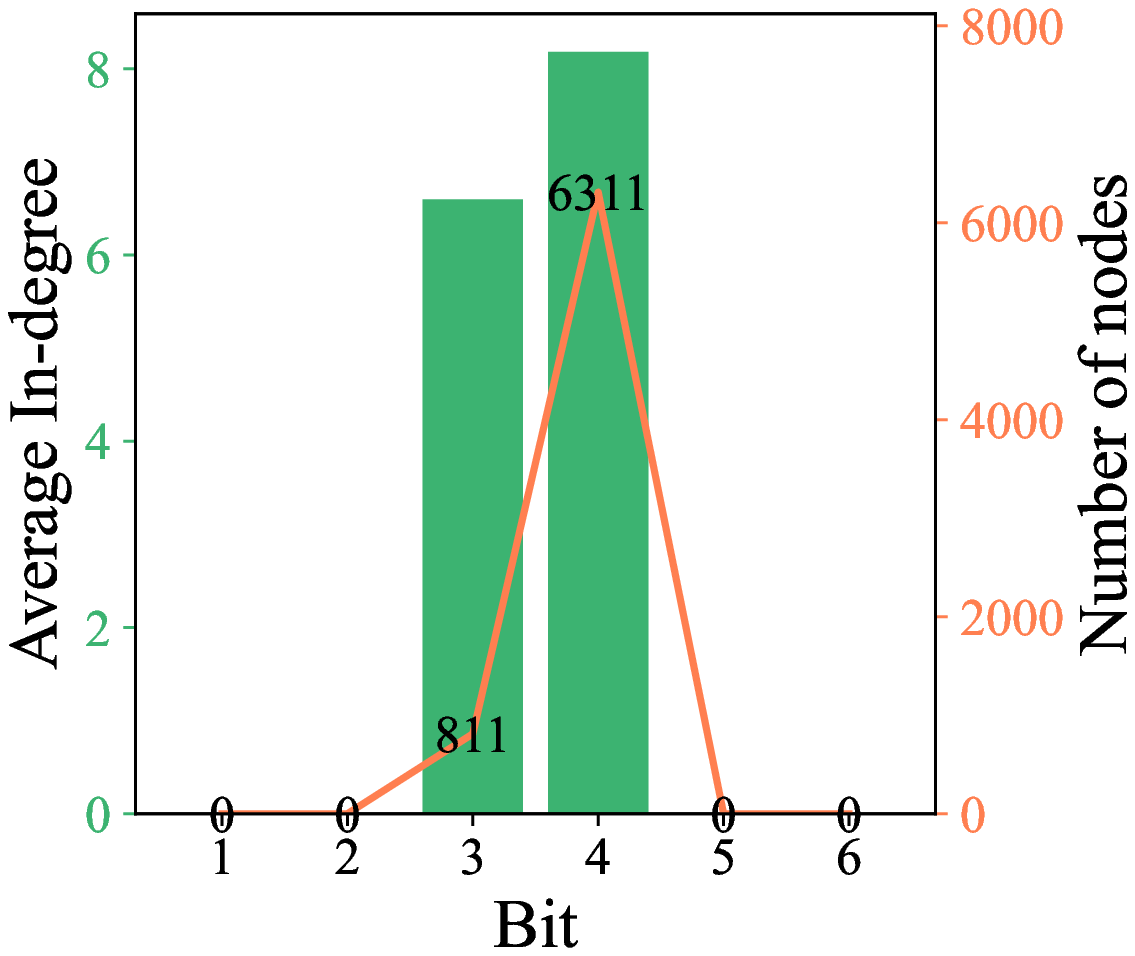}
         % \caption{GAT-CiteSeer}
         \label{gcn_mnist_3}
      \end{minipage}
   }
   \subfigure[4-th layer]{
      \begin{minipage}[t]{0.23\textwidth}
         \centering
         \includegraphics[width=1\linewidth,height=0.73\linewidth]{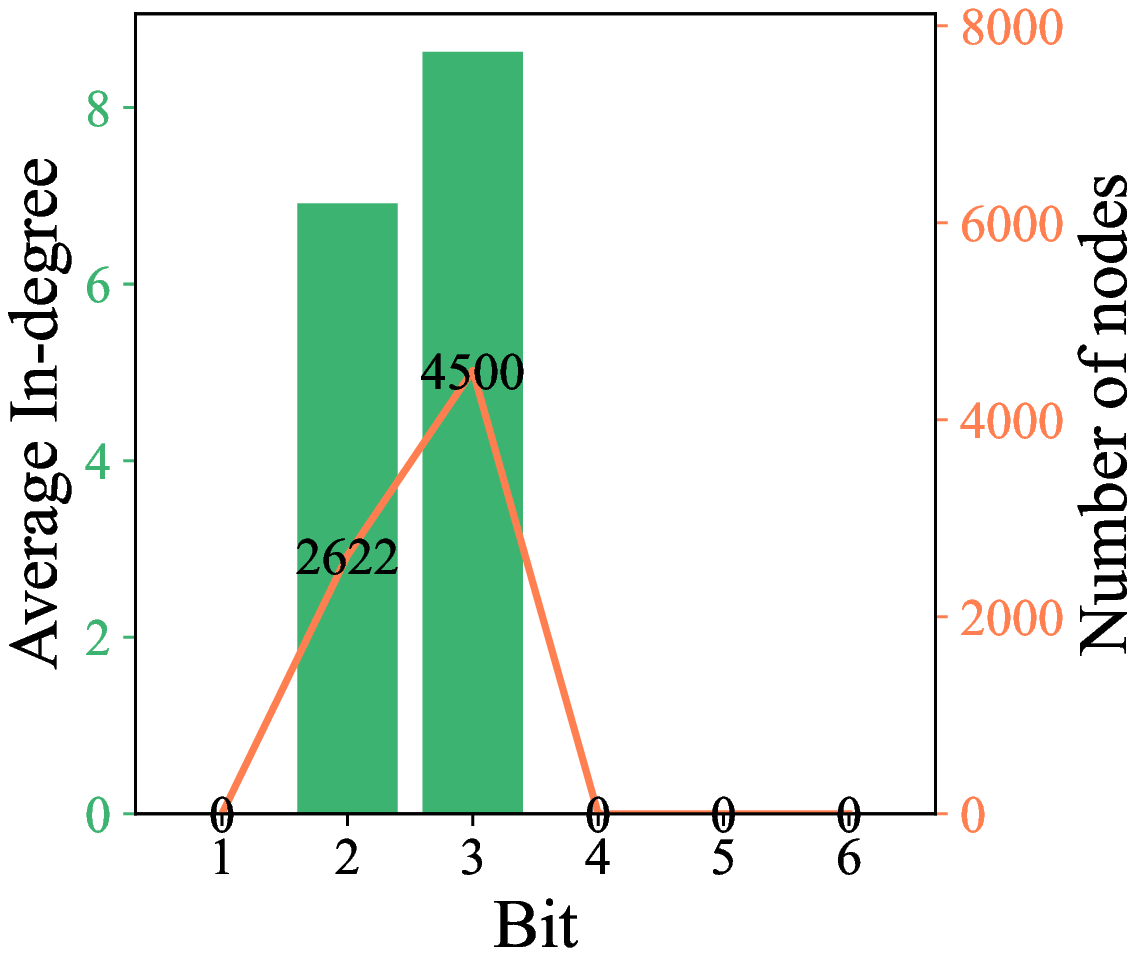}
         % \caption{GAT-CiteSeer}
         \label{gcn_mnist_4}
      \end{minipage}
   }
   \caption{The relationship between bit and average in-degrees of nodes using the corresponding bitwidth to quantize 
      on different layers of GCN trained on MNIST.}
   \label{gcn_mnist_bit_deg}
\end{figure}

\begin{figure}[H]
   \centering
   \subfigure[1-st layer]{
      \begin{minipage}[t]{0.23\textwidth}
         \centering
         \includegraphics[width=1\linewidth,height=0.72\linewidth]{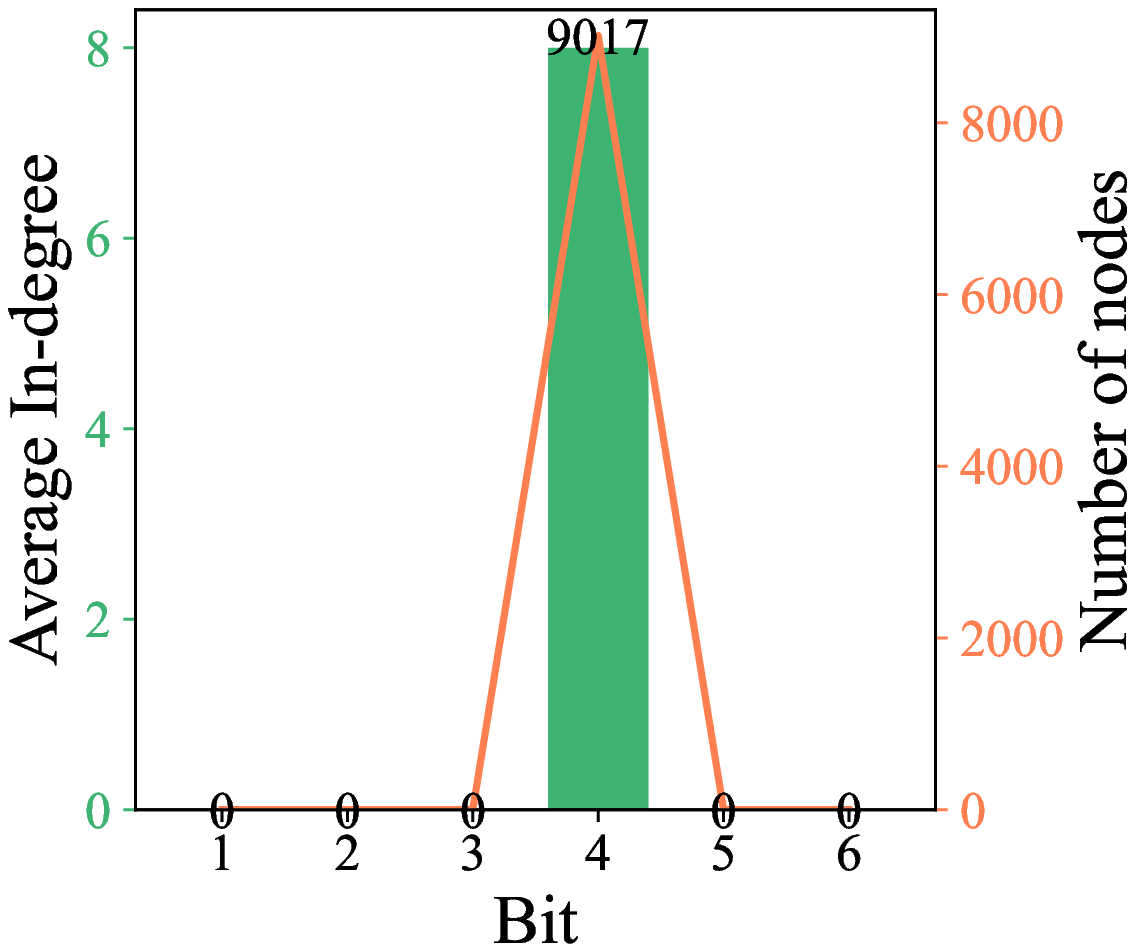}
         % \caption{GCN-CiteSeer}
         \label{gin_mnist_1}
      \end{minipage}
   }
   \subfigure[2-nd layer]{
      \begin{minipage}[t]{0.23\textwidth}
         \centering
         \includegraphics[width=1\linewidth,height=0.72\linewidth]{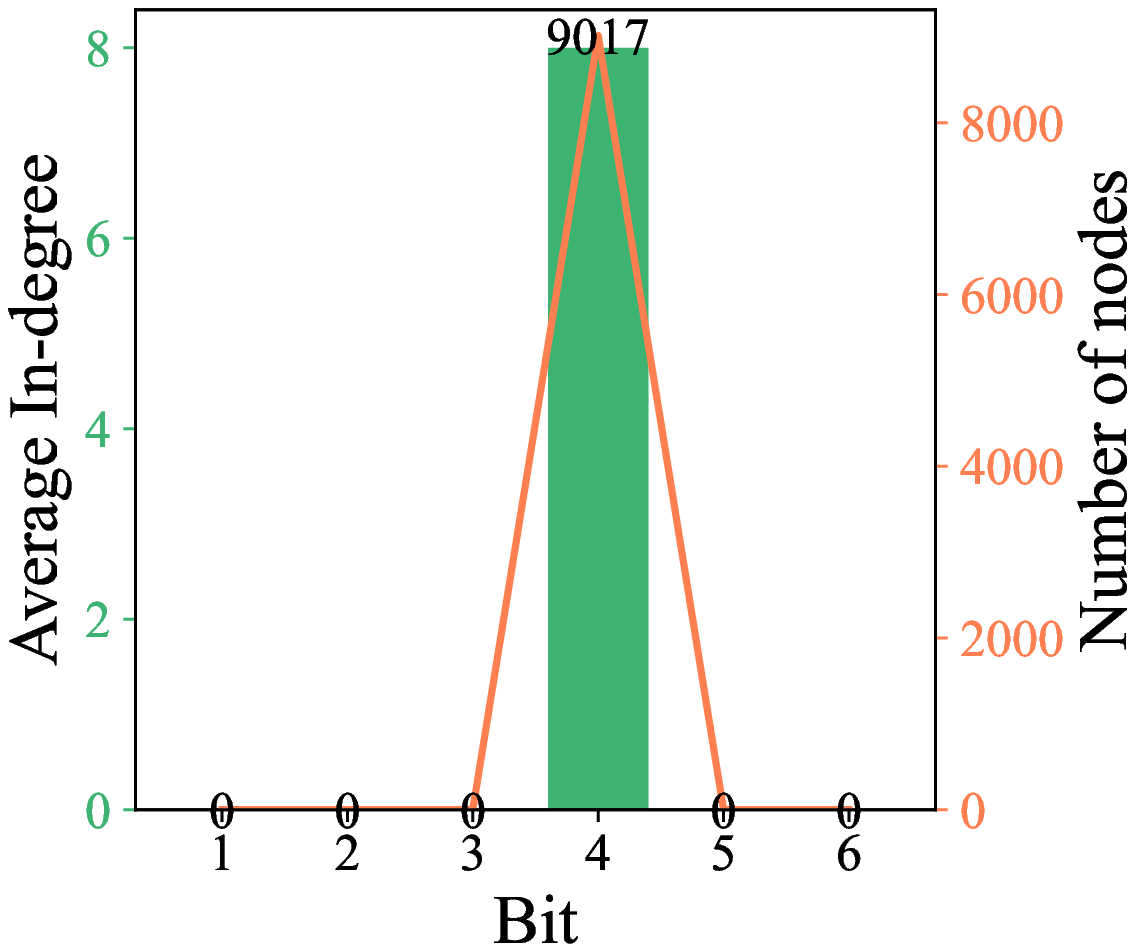}
         % \caption{GIN-CiteSeer}
         \label{gin_mnist_2}
      \end{minipage}
   }
   \subfigure[3-rd layer]{
      \begin{minipage}[t]{0.23\textwidth}
         \centering
         \includegraphics[width=1\linewidth,height=0.72\linewidth]{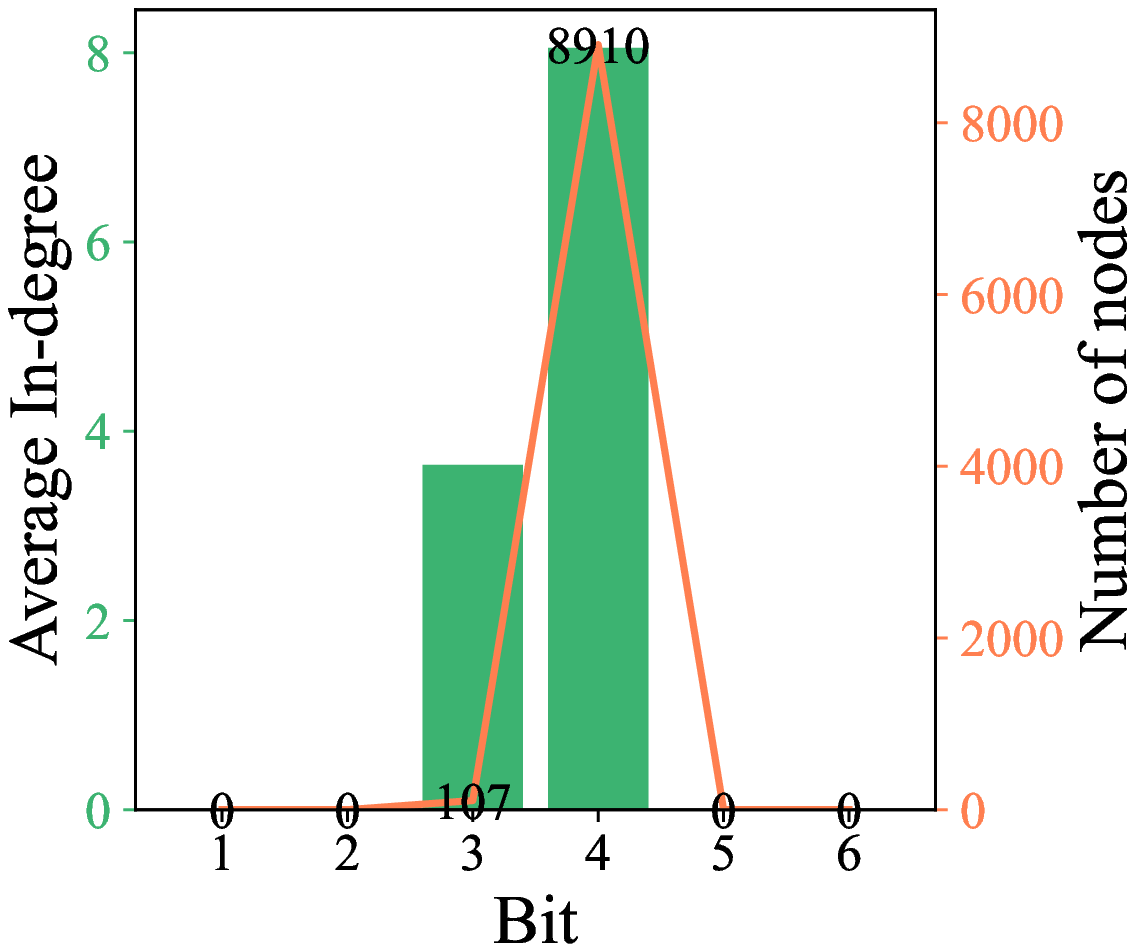}
         % \caption{GAT-CiteSeer}
         \label{gin_mnist_3}
      \end{minipage}
   }
   \subfigure[4-th layer]{
      \begin{minipage}[t]{0.23\textwidth}
         \centering
         \includegraphics[width=1\linewidth,height=0.72\linewidth]{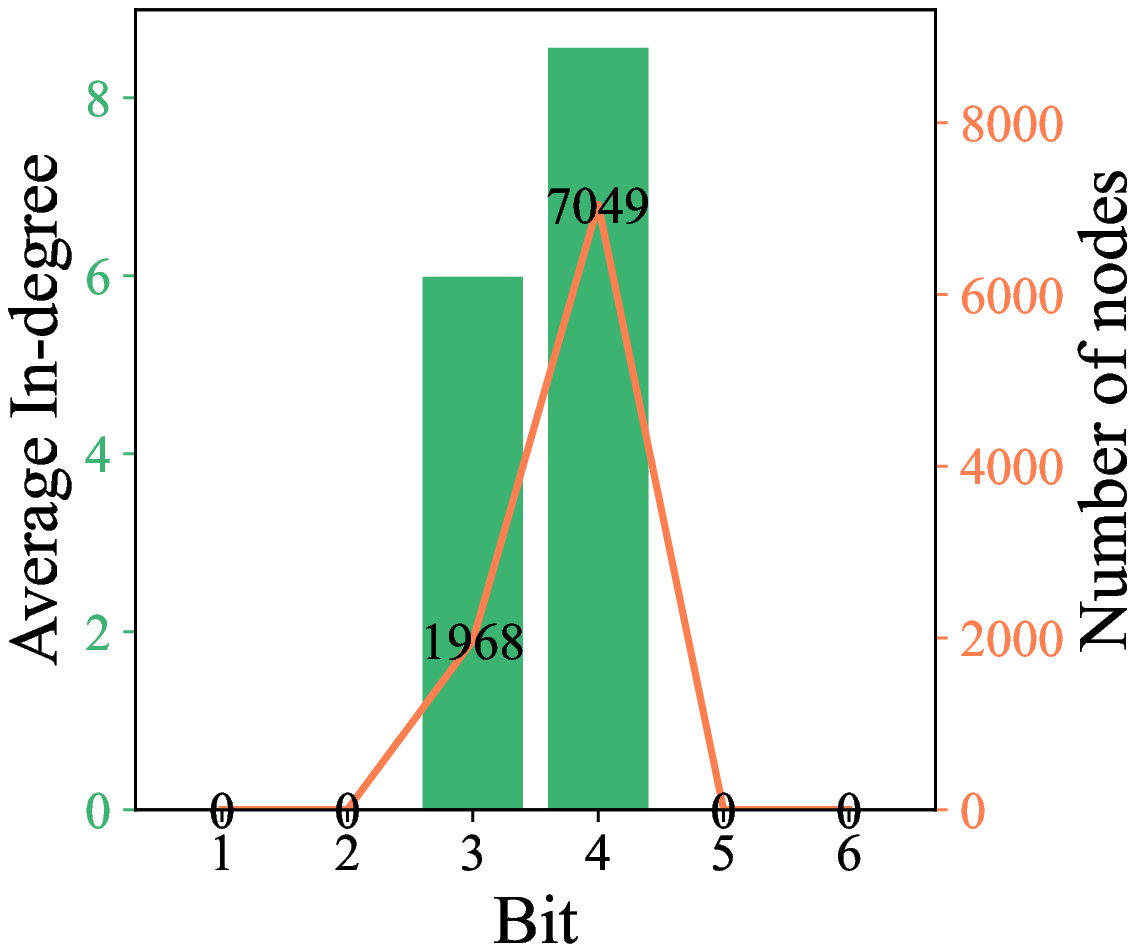}
         % \caption{GAT-CiteSeer}
         \label{gin_mnist_4}
      \end{minipage}
   }
   \caption{The relationship between bit and average in-degrees of nodes using the corresponding bitwidth to quantize 
      on different layers of GIN trained on MNIST.}
   \label{gin_mnist_bit_deg}
\end{figure}

\begin{figure}[H]
   \centering
   \subfigure[1-st layer]{
      \begin{minipage}[t]{0.23\textwidth}
         \centering
         \includegraphics[width=1\linewidth,height=0.73\linewidth]{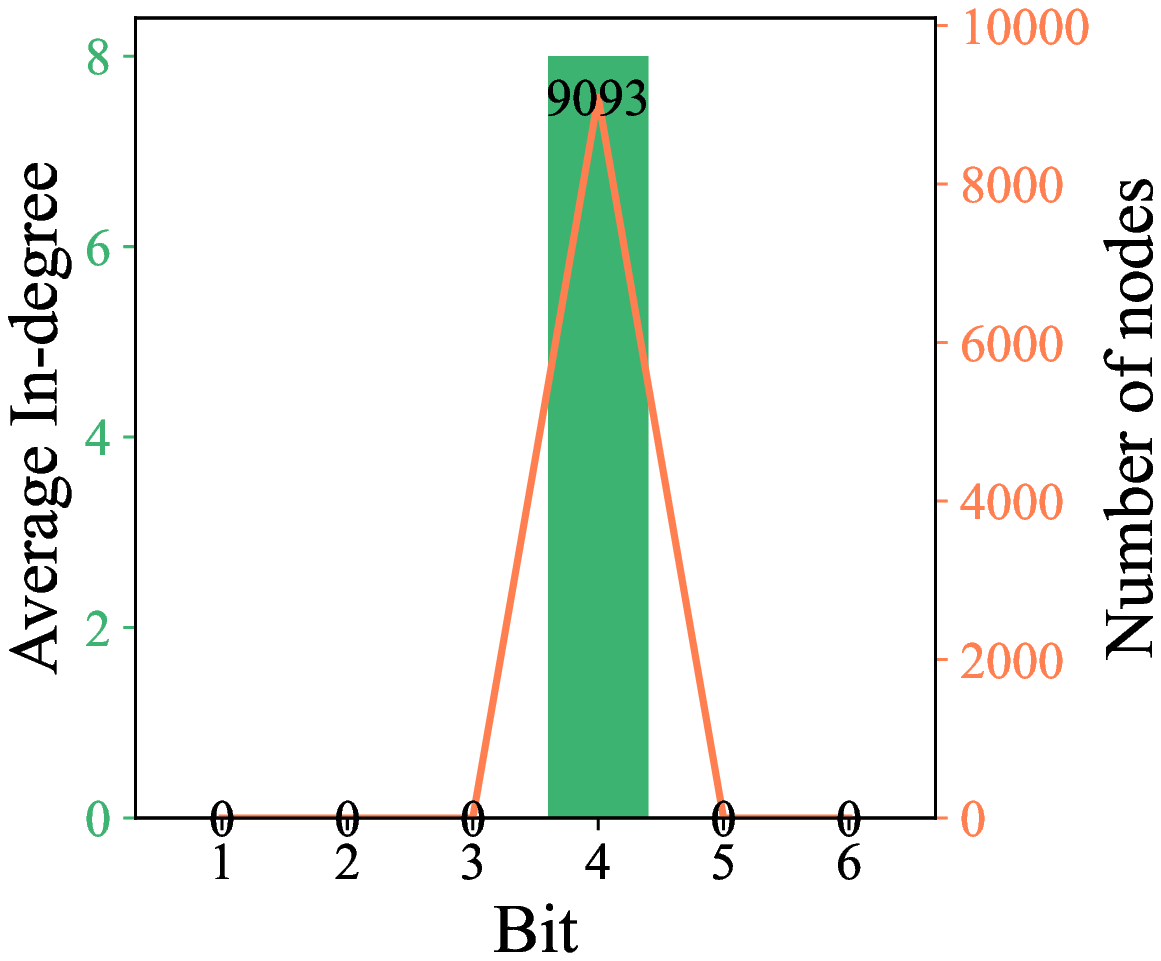}
         % \caption{GCN-CiteSeer}
         \label{gat_mnist_1}
      \end{minipage}
   }
   \subfigure[2-nd layer]{
      \begin{minipage}[t]{0.23\textwidth}
         \centering
         \includegraphics[width=1\linewidth,height=0.73\linewidth]{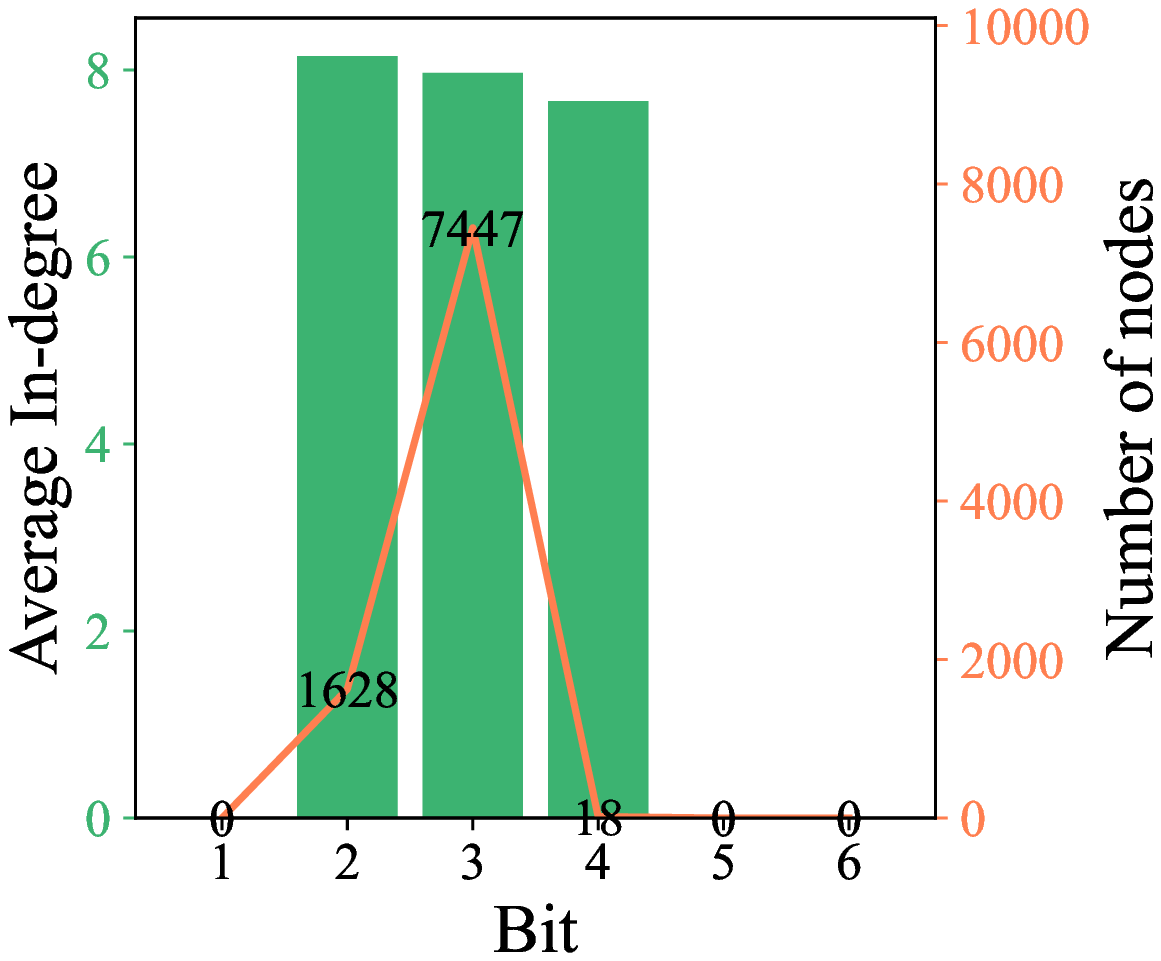}
         % \caption{GIN-CiteSeer}
         \label{gat_mnist_2}
      \end{minipage}
   }
   \subfigure[3-rd layer]{
      \begin{minipage}[t]{0.23\textwidth}
         \centering
         \includegraphics[width=1\linewidth,height=0.73\linewidth]{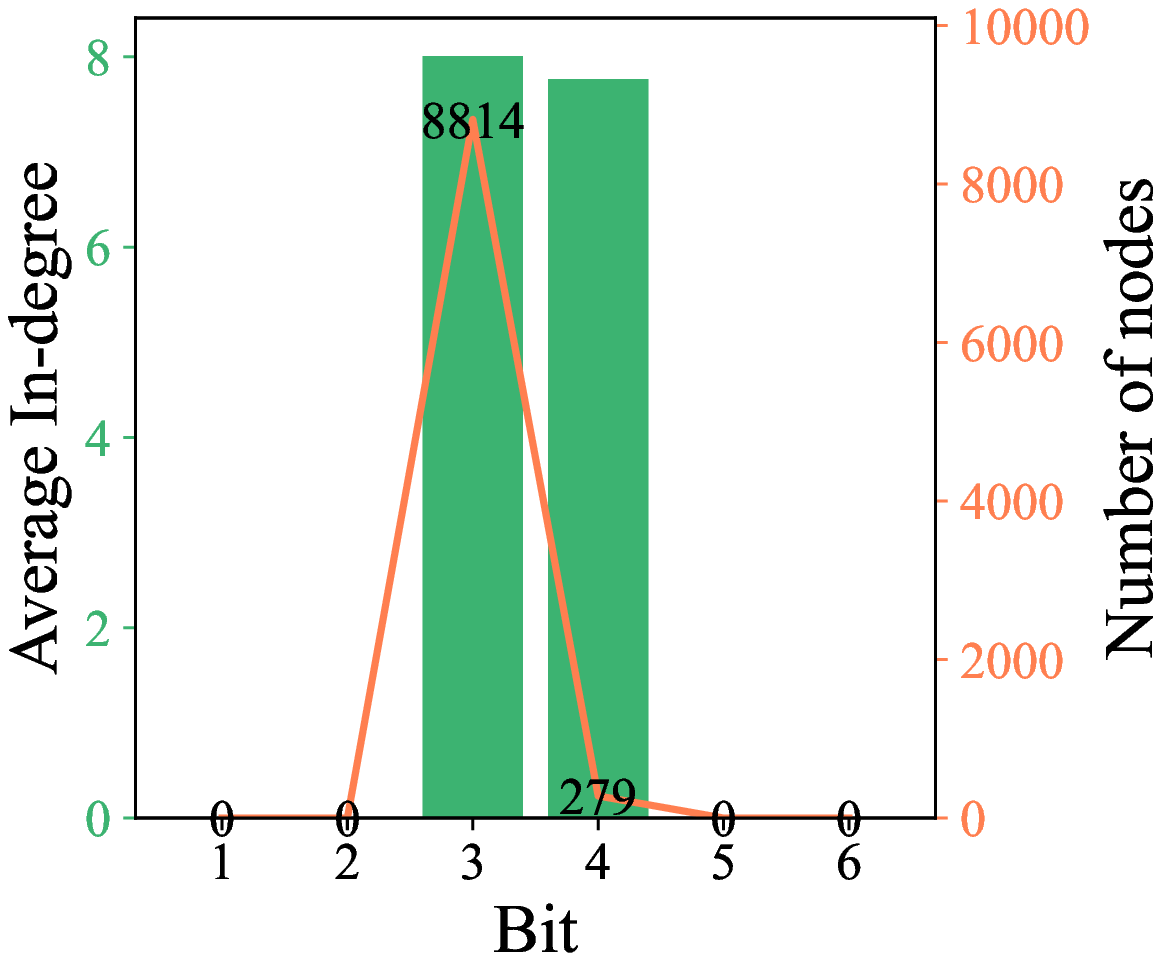}
         % \caption{GAT-CiteSeer}
         \label{gat_mnist_3}
      \end{minipage}
   }
   \subfigure[4-th layer]{
      \begin{minipage}[t]{0.23\textwidth}
         \centering
         \includegraphics[width=1\linewidth,height=0.73\linewidth]{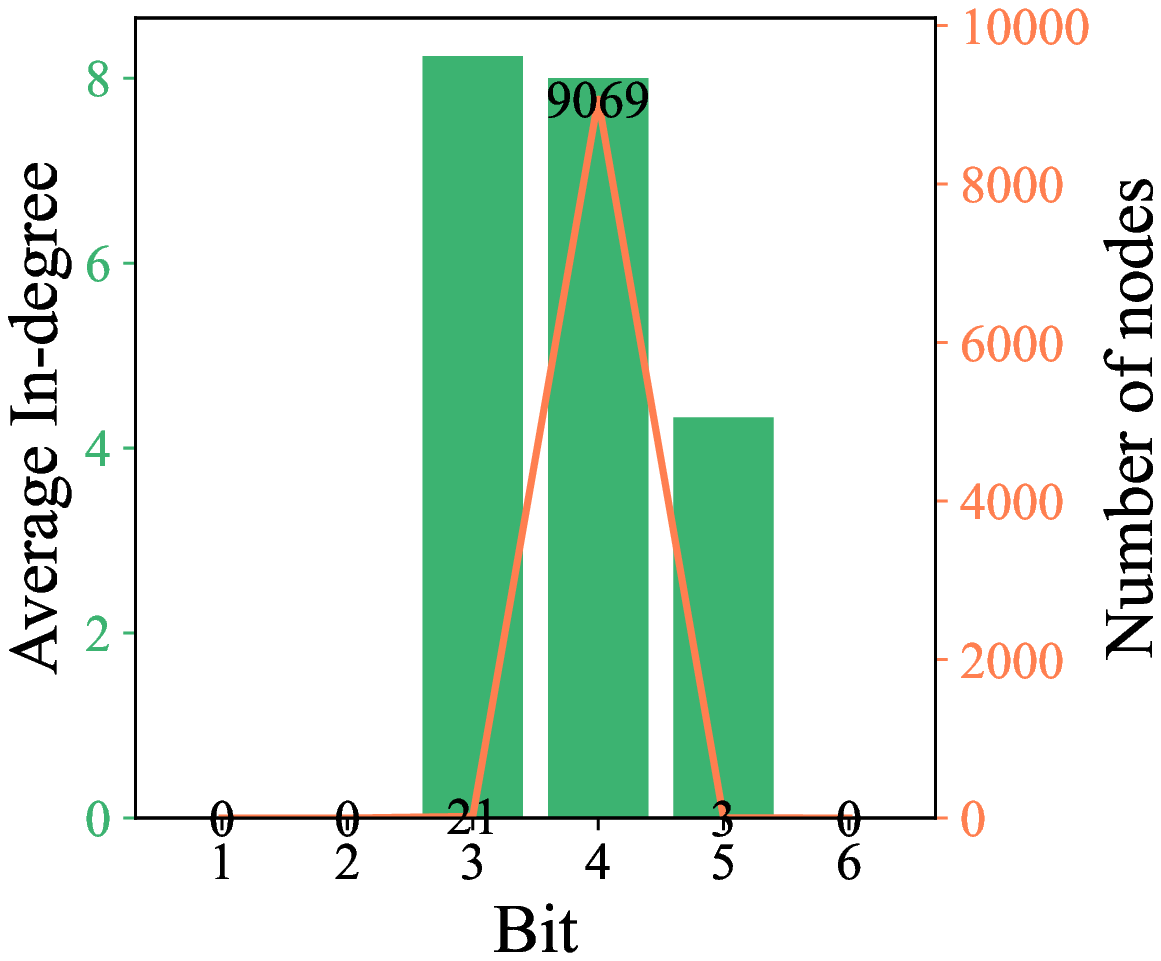}
         % \caption{GAT-CiteSeer}
         \label{gat_mnist_4}
      \end{minipage}
   }
   \caption{The relationship between bit and average in-degrees of nodes using the corresponding bitwidth to quantize 
      on different layers of GAT trained on MNIST.}
   \label{gat_mnist_bit_deg}
   \end{figure}

\linespread{1.5}
   \begin{table}[H]
      \caption{The results comparison on GIN-ZINC and GAT-ZINC.}
      \label{appen-zinc-results}
      \begin{center}
      \begin{tabular}{clllll}
      
      \hline \toprule[2pt]
         Modle                               & Dataset              & Loss↓                 & Average bits                                            & Compression Ratio \\  \midrule[1pt]
      \multirow{6}{*}{\textbf{ZINC}}         & GAT(FP32)            & 0.455±0.006           & 32                                                      & 1x                  \\          
                                             & GAT(DQ)              & 0.520±0.021           & 4                                                       & 8x                \\
                                             & GAT(ours)            & \textbf{0.495±0.006}  & 4                                                       & 8x                \\ \cline{2-5}
                                             & GIN(FP32)            & 0.334±0.024           & 32                                                      & 1x                \\
                                             & GIN(DQ)              & 0.431±0.012           & 4                                                       & 8x                \\
                                             & GIN(ours)            & \textbf{0.380±0.022}  & 4                                                       & 8x                \\ \bottomrule[2pt]
      
      \end{tabular}
      \end{center}
      \end{table}
      \begin{figure}[H]
         \centering
         \subfigure[1-st layer]{
            \begin{minipage}[t]{0.23\textwidth}
               \centering
               \includegraphics[width=1\linewidth,height=0.85\linewidth]{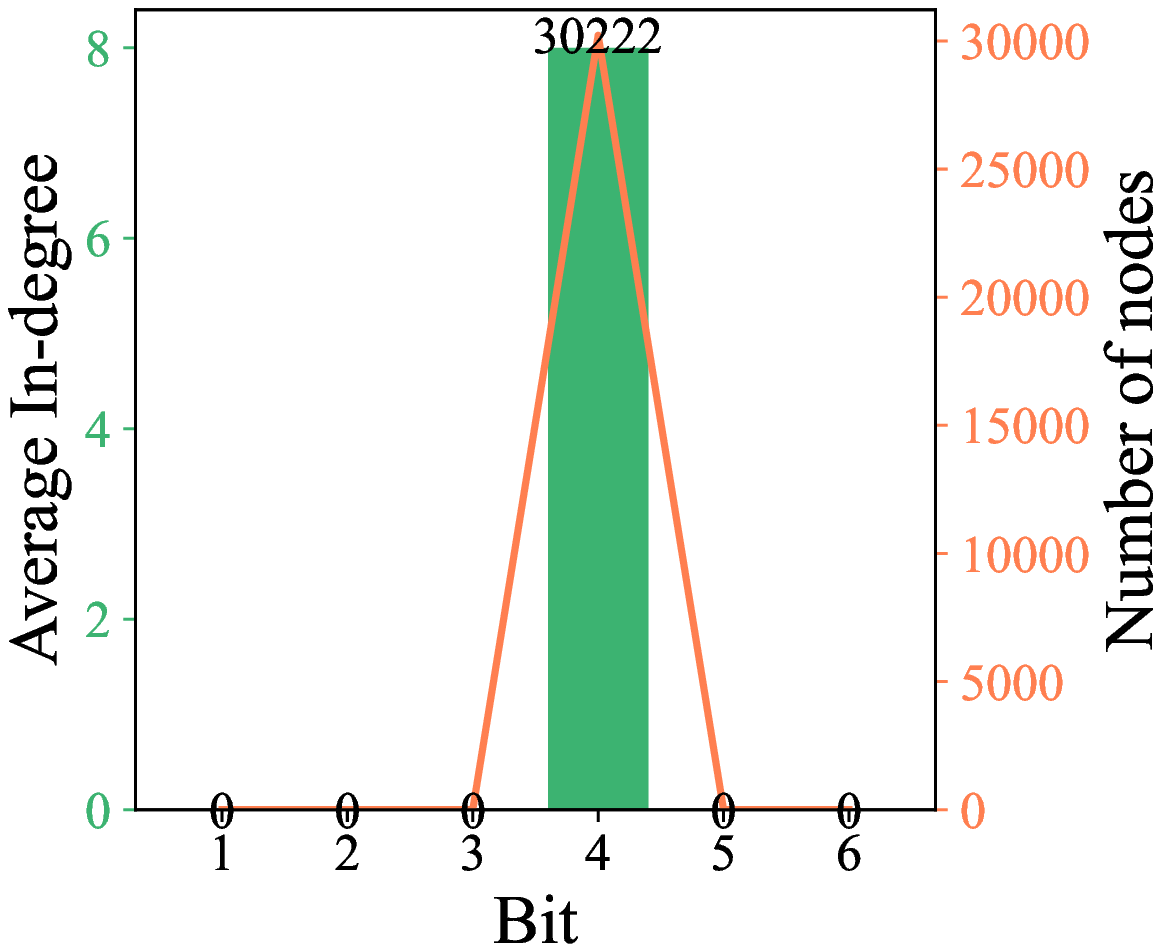}
               % \caption{GCN-CiteSeer}
               \label{gcn_cifar_1}
            \end{minipage}
         }
         \subfigure[2-nd layer]{
            \begin{minipage}[t]{0.23\textwidth}
               \centering
               \includegraphics[width=1\linewidth,height=0.85\linewidth]{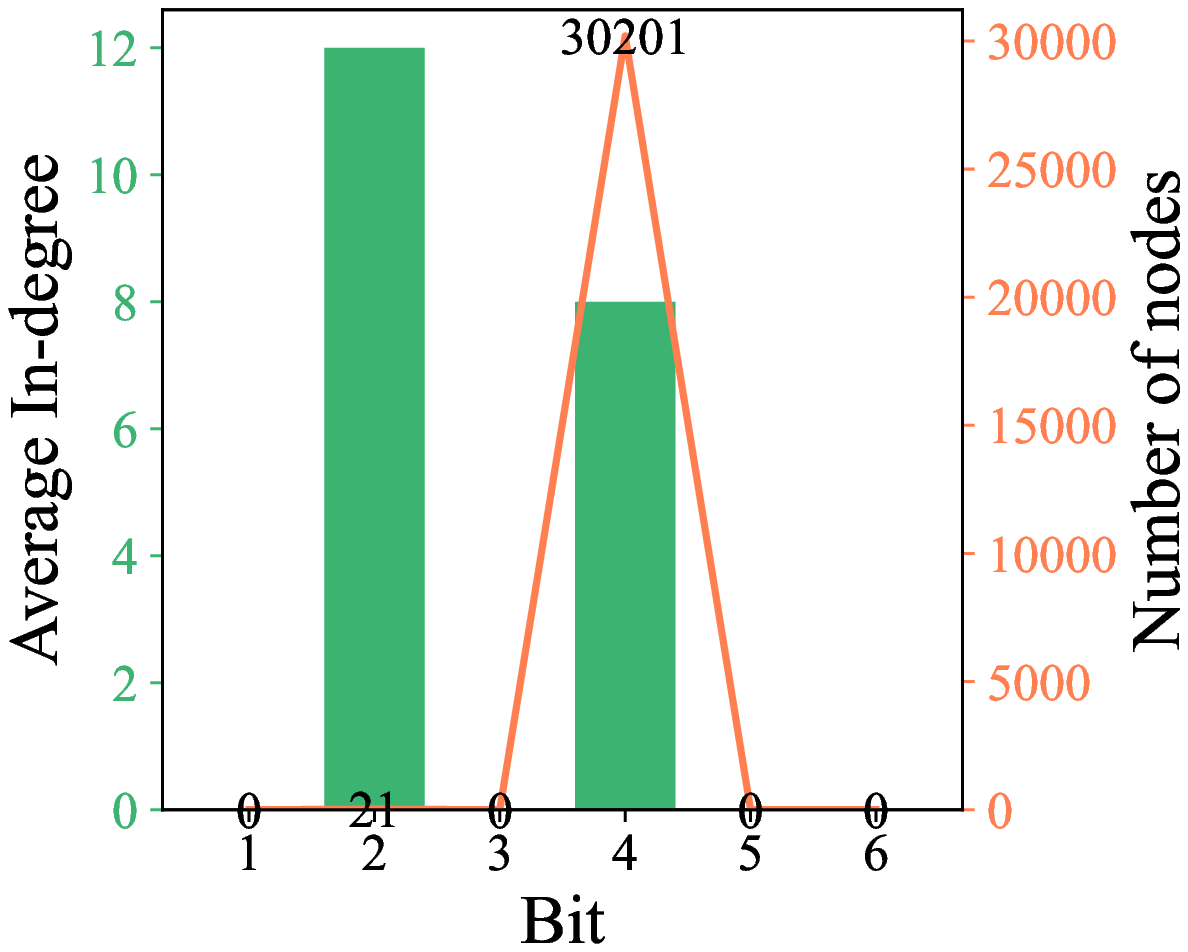}
               % \caption{GIN-CiteSeer}
               \label{gcn_cifar_2}
            \end{minipage}
         }
         \subfigure[3-rd layer]{
            \begin{minipage}[t]{0.23\textwidth}
               \centering
               \includegraphics[width=1\linewidth,height=0.85\linewidth]{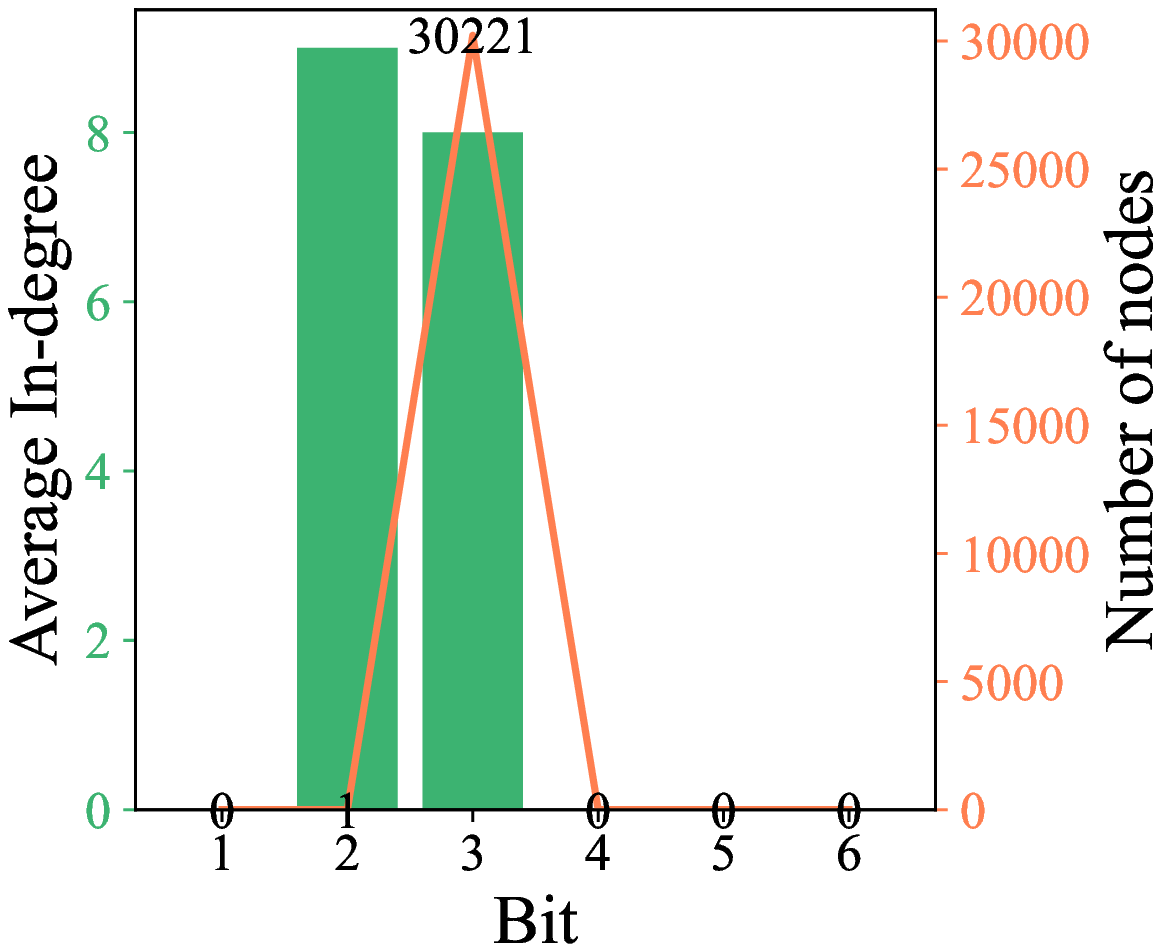}
               % \caption{GAT-CiteSeer}
               \label{gcn_cifar_3}
            \end{minipage}
         }
         \subfigure[4-th layer]{
            \begin{minipage}[t]{0.23\textwidth}
               \centering
               \includegraphics[width=1\linewidth,height=0.85\linewidth]{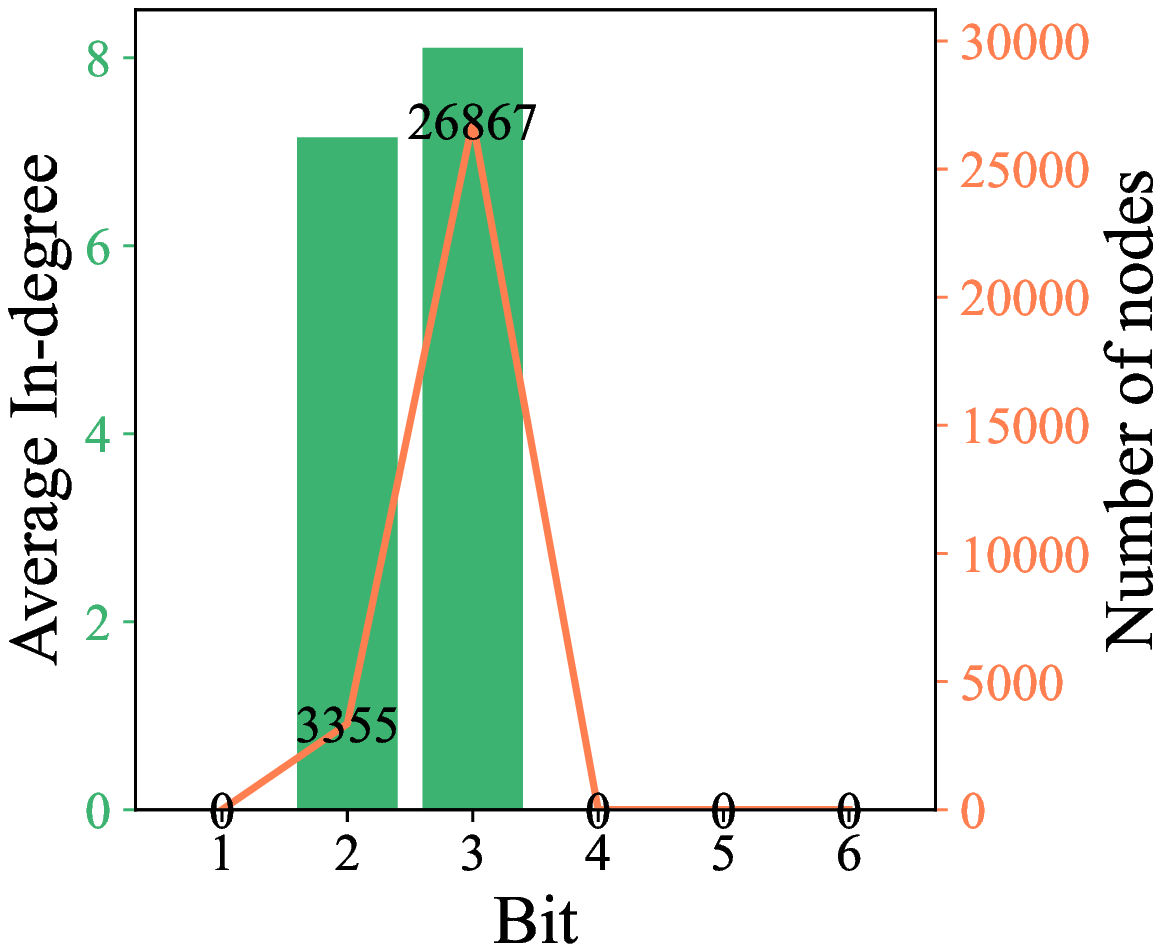}
               % \caption{GAT-CiteSeer}
               \label{gcn_cifar_4}
            \end{minipage}
         }
         \caption{The relationship between bit and average in-degrees of nodes using the corresponding bitwidth to quantize 
            on different layers of GCN trained on CIFAR10.}
         \label{gcn_cifar_bit_deg}
      \end{figure}
      \begin{figure}[H]
         \centering
         \subfigure[1-st layer]{
            \begin{minipage}[t]{0.23\textwidth}
               \centering
               \includegraphics[width=1\linewidth,height=0.85\linewidth]{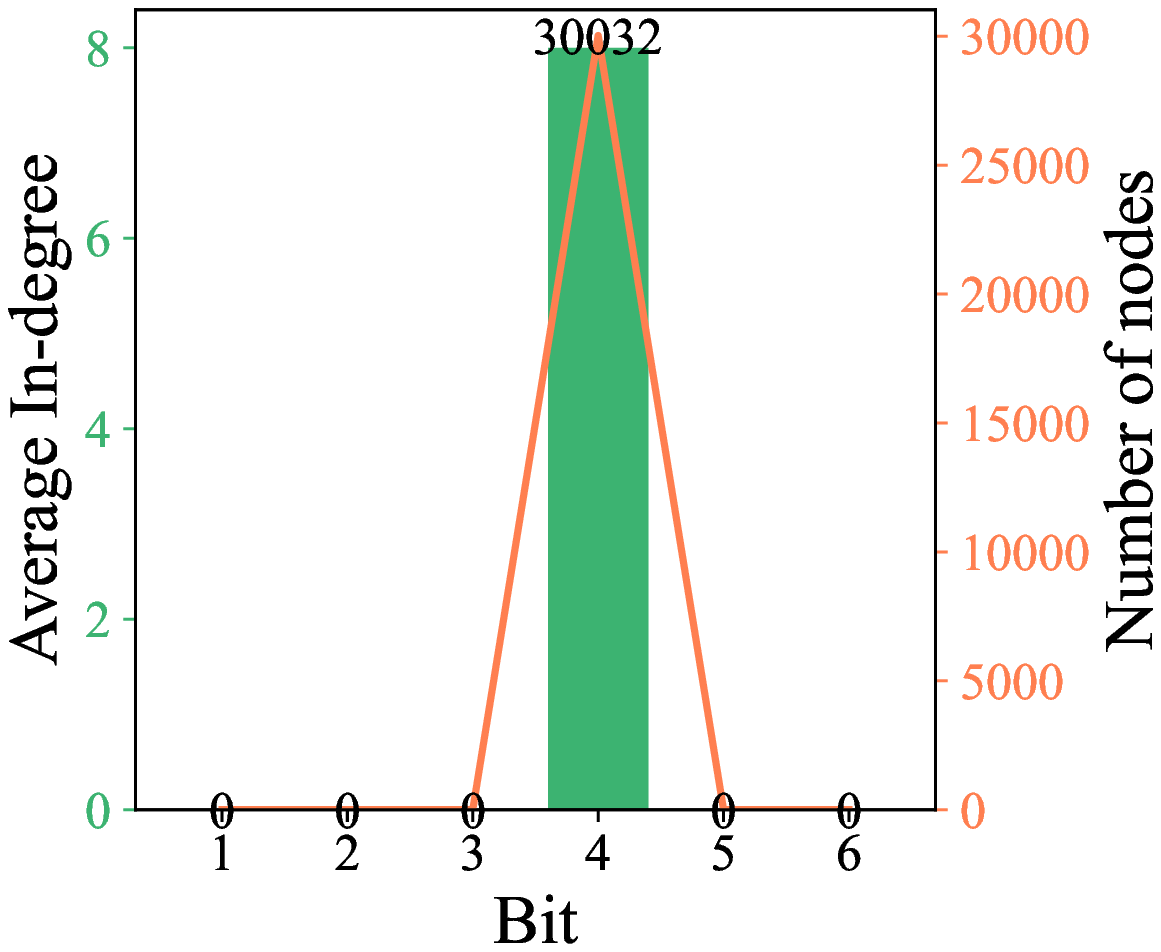}
               % \caption{GCN-CiteSeer}
               \label{gin_cifar_1}
            \end{minipage}
         }
         \subfigure[2-nd layer]{
            \begin{minipage}[t]{0.23\textwidth}
               \centering
               \includegraphics[width=1\linewidth,height=0.85\linewidth]{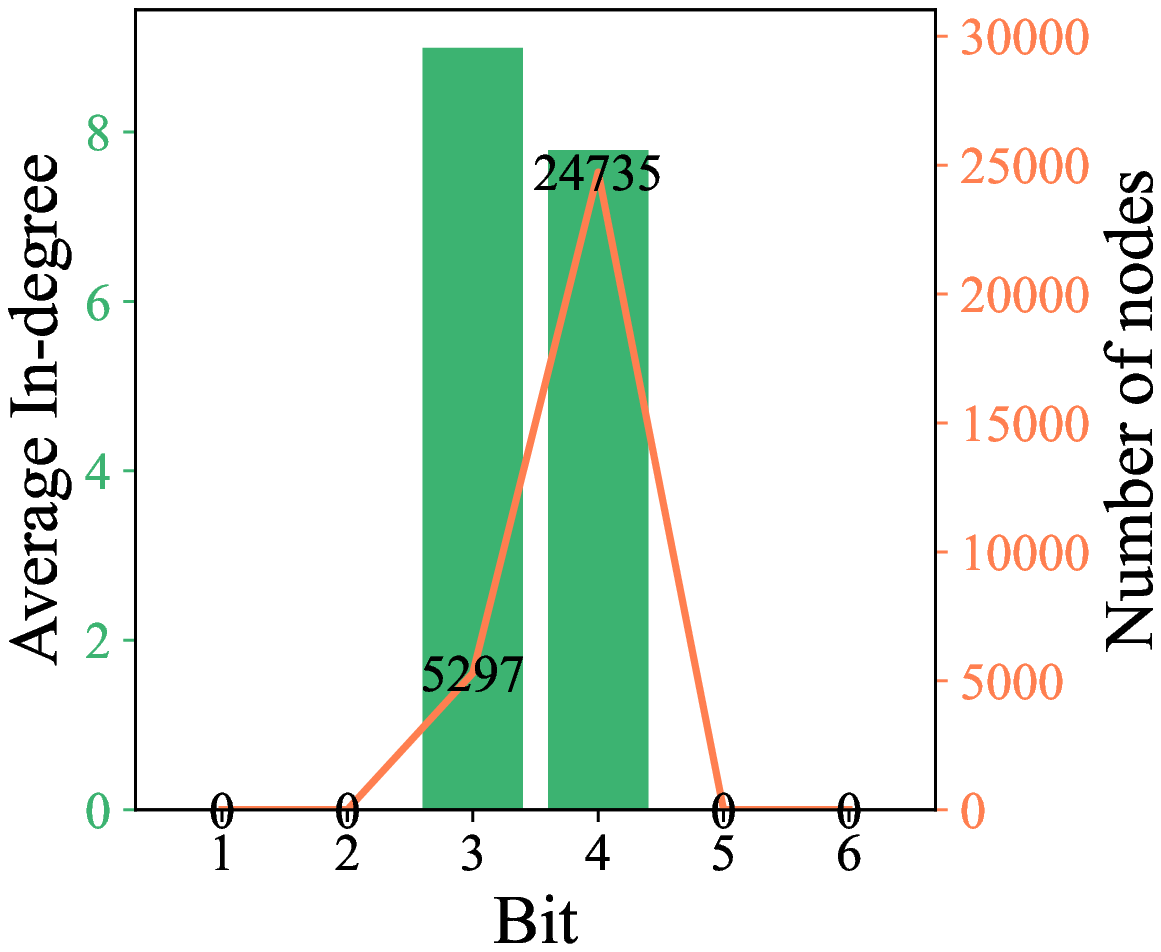}
               % \caption{GIN-CiteSeer}
               \label{gin_cifar_2}
            \end{minipage}
         }
         \subfigure[3-rd layer]{
            \begin{minipage}[t]{0.23\textwidth}
               \centering
               \includegraphics[width=1\linewidth,height=0.85\linewidth]{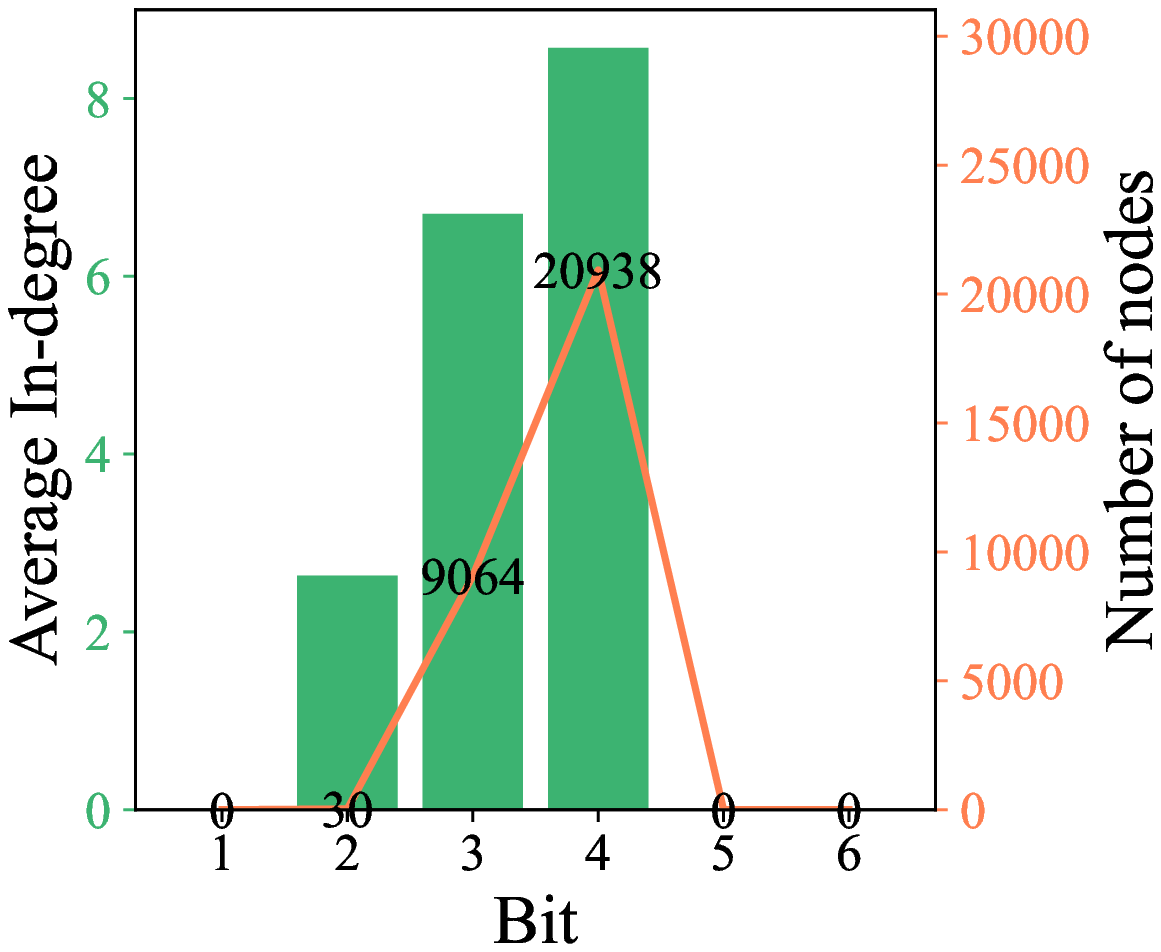}
               % \caption{GAT-CiteSeer}
               \label{gin_cifar_3}
            \end{minipage}
         }
         \subfigure[4-th layer]{
            \begin{minipage}[t]{0.23\textwidth}
               \centering
               \includegraphics[width=1\linewidth,height=0.85\linewidth]{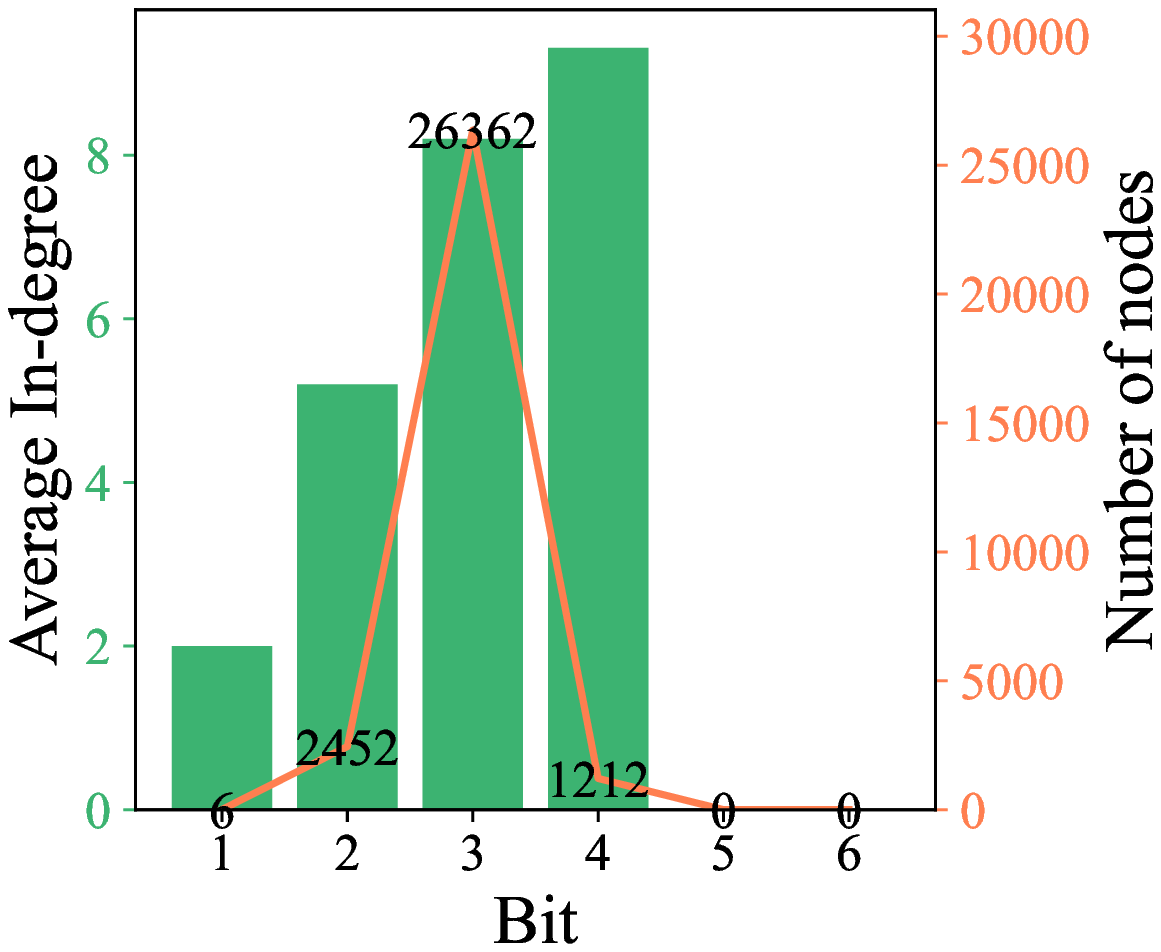}
               % \caption{GAT-CiteSeer}
               \label{gin_cifar_4}
            \end{minipage}
         }
         \caption{The relationship between bit and average in-degrees of nodes using the corresponding bitwidth to quantize 
            on different layers of GIN trained on CIFAR10.}
         \label{gin_cifar_bit_deg}
      \end{figure}
We also show the relationship between bit and average in-degree of nodes using the corresponding bit to quantize for more 
graph-level tasks in different layers immediately after the aggregation 
phase 
in Figure \ref{gcn_cifar_bit_deg}-Figure \ref{gat_mnist_bit_deg}. 
The quantization bitwidths learned for graph-level tasks are also aggregation-aware. 
Because the difference of the in-degrees between different nodes is 
little in the MNIST and CIFAR10 dataset resulting in the 
aggregated features are similar between different nodes, 
the relationship between learned bitwidths and the in-degrees
is irregular in some layers, e.g., the 2-nd layer in GCN trained on MNIST.

\linespread{1.1}
\begin{table}[H]
   \caption[]{The impact of the depth of GNNs on quantization performance.}
   \label{layer_result}
   \begin{center}
      \begin{tabular}{cccccccc}
         \hline \toprule[2pt]
         \multicolumn{2}{c}{Layers}       & \multicolumn{2}{c}{3}                                              & \multicolumn{2}{c}{4}                                                 & \multicolumn{2}{c}{5}                                               \\ \midrule[1pt]
         \multicolumn{2}{c}{Task}         & Accu(\%)  & \begin{tabular}[c]{@{}c@{}}Avarage\\ Bits\end{tabular} & Accu(\%)  & \begin{tabular}[c]{@{}c@{}}Avarage   \\ Bits\end{tabular} & Accu(\%) & \begin{tabular}[c]{@{}c@{}}Avarage  \\ Bits\end{tabular} \\ \hline
         \multirow{2}{*}{GCN-Cora} & FP32 & 80.5±0.6  & 32                                                     & 79.3±0.1  & 32                                                        & 75.8±3.2 & 32                                                       \\
                                   & Ours & 80.2±0.6  & 2.94                                                   & 78.2±0.9  & 3.54                                                      & 75.0±1.2 & 3.61                                                     \\ \cline{2-8} 
         \multirow{2}{*}{GIN-Cora} & FP32 & 49.4±15.8 & 32                                                     & 37.1±13.1 & 32                                                        & ---      & ---                                                      \\
                                   & Ours & 54.5±12.6 & 3.3                                                    & 36.4±11.1 & 3.1                                                       & ---      & ---                                                      \\ \bottomrule[2pt]
         \end{tabular}
   \end{center}
\end{table}

\begin{table}[H]
   \caption[]{The comparison between the model with and without skip connection on GCN-Cora task.}
   \label{skip impact}
   \begin{center}
      \begin{tabular}{cccccc}
         \hline\toprule[2pt]
         \multirow{2}{*}{Layers} & \multirow{2}{*}{GCN-Cora} & \multicolumn{2}{c}{\begin{tabular}[c]{@{}c@{}}Without\\    \\ skip   connection\end{tabular}} & \multicolumn{2}{c}{\begin{tabular}[c]{@{}c@{}}With\\    \\ skip   connection\end{tabular}} \\ \cline{3-6} 
                                 &                           & FP32                                          & Ours                                          & FP32                                         & Ours                                        \\ \midrule[1pt]
         \multirow{2}{*}{3}      & Accu(\%)                  & 80.5±0.6                                      & 80.2±0.6                                      & 82.5±0.5                                     & 82.2±0.7                                    \\
                                 & Bits                      & 32                                            & 2.94                                          & 32                                           & 2.37                                        \\ \hline
         \multirow{2}{*}{4}      & Accu(\%)                  & 79.3±0.1                                      & 78.2±0.9                                      & 81.9±0.7                                     & 81.5±0.3                                    \\
                                 & Bits                      & 32                                            & 3.54                                          & 32                                           & 2.63                                        \\ \hline
         \multirow{2}{*}{5}      & Accu(\%)                  & 75.8±3.2                                      & 75.0±1.2                                      & 81.1±1.1                                     & 80.6±0.6                                    \\
                                 & Bits                      & 32                                            & 3.61                                          & 32                                           & 2.72                                        \\ \hline
         \multirow{2}{*}{6}      & Accu(\%)                  & 73.8±1.6                                      & 73.1±1.9                                      & 80.1±0.8                                     & 80.4±0.7                                    \\
                                 & Bits                      & 32                                            & 4.62                                          & 32                                           & 2.98                                        \\ \bottomrule[2pt]
         \end{tabular}
   \end{center}
   \end{table}

\begin{figure}[ht]
   \centering
   
      \begin{minipage}[t]{0.45\textwidth}
         \centering
         \includegraphics[scale=0.4]{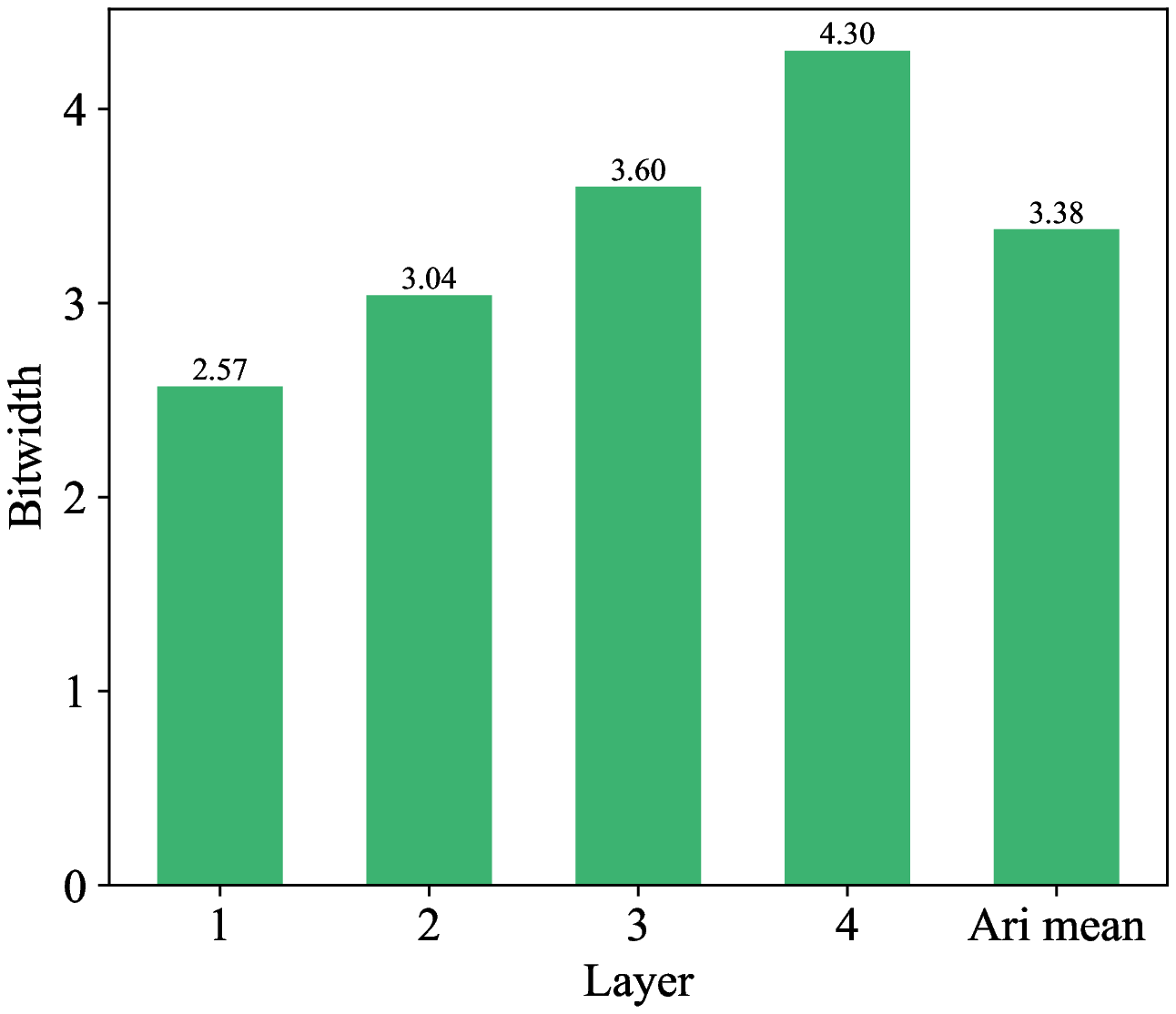}
         \caption[]{The average bitwidth for 2nd-5th layer in five layers GCN.}
         \label{5layer_bit}
      \end{minipage}\quad
      \begin{minipage}[t]{0.45\textwidth}
         \centering
         \includegraphics[scale=0.4]{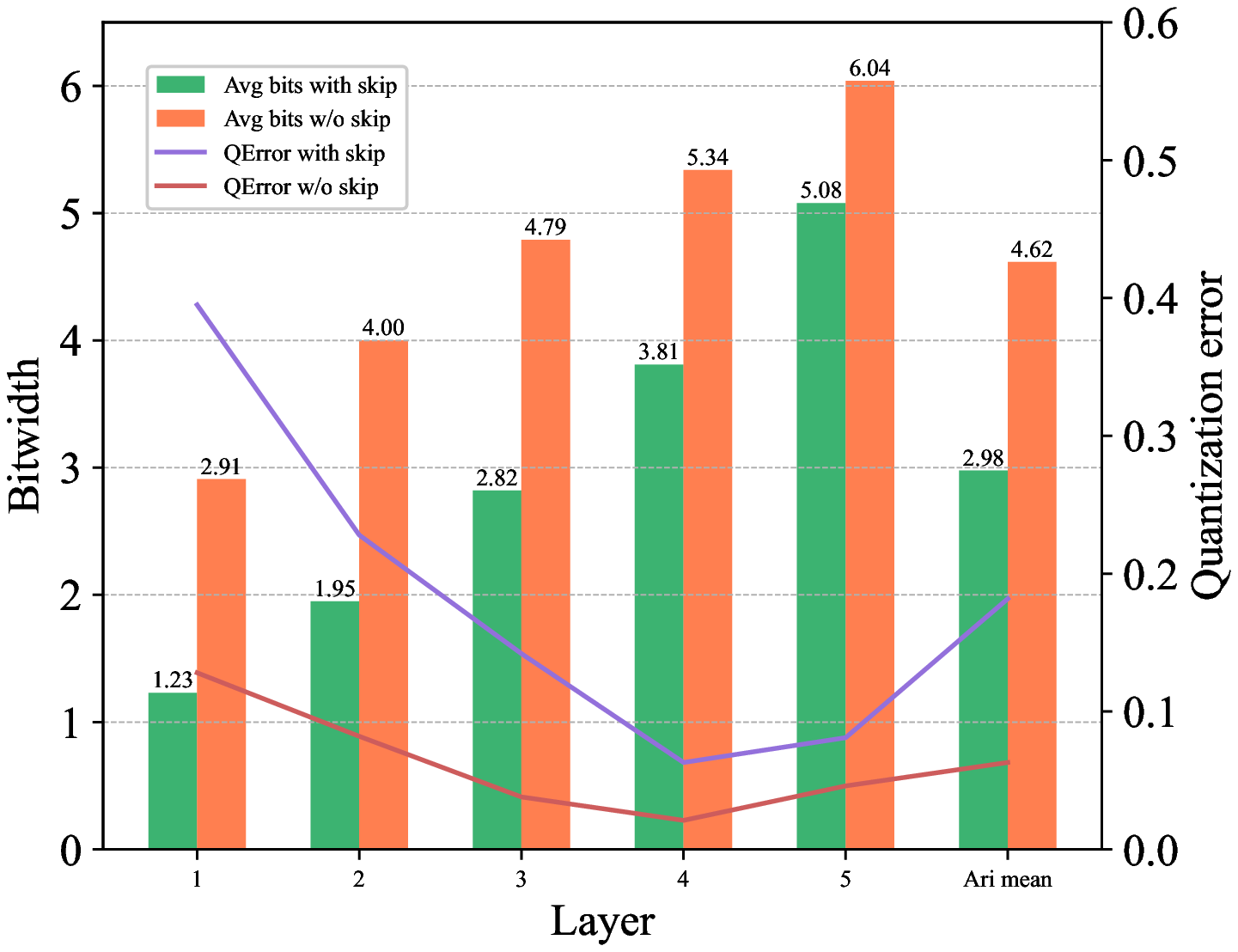}
         \caption[]{The average bitwidth and quantization error for 2nd-6th layer in six layers GCN.}
         \label{skip impact}
      \end{minipage}
   
\end{figure}
\subsubsection{More Ablation Study}
\textbf{The impact of the depth of GNNs on quantization performance:} We explore 
how 
a different number of GNN layers impacts the quantization performance of GCN-Cora and GIN-CiteSeer.
We explore the quantization performance on 3,4,5,6 layers GCN 
model and 
3,4 layers 
GIN model (the GCN and GIN used in Table \ref{node-level-results} are 2 layers).
We did not explore the deeper GNN models 
because the accuracy of the model decreases drastically 
as the number of model layers increases due to the over-smooth phenomenon in GNNs.
As shown in Table \ref{layer_result},
our method can also maintain the 
performance with a high compression ratio for the model with different layers 
compared with the FP32 model.

In addition, we observe
that the learned quantization bitwidth increases with the number of layers.  
We analysis
the average bitwidth used by 2nd to 5th layer for the five layers GCN model in Figure \ref{5layer_bit}. 
Our method learns a higher
bitwidth for the deeper layer. Due to the over-smooth phenomenon that exists in the deep layer, the 
embedding features of different nodes are similar in the deep layer. Therefore, we 
consider
the deeper layer may need a higher quantization 
bitwidth to distinguish the embedding features of different nodes.

\textbf{The impact of skip connection on quantization performance:}
The first column denoted by `Without skip connection' and 
the second column denoted by `With skip connection'  of \ref{skip impact} present the comparison 
results for 
different layers GCN on Cora datasets 
without skip connection and with skip connection, respectively.
For the model with skip connection,
our method is also effective. Our method learns a higher
bitwidth for the deeper layer. Due to the over-smooth phenomenon that exists in the deep layer, 
we 
consider
that the deeper layer may need a higher quantization 
bitwidth to distinguish the embedding features of different nodes. 
and the higher learned quantization bitwidth for deeper layers also alleviate 
quantization error.
And compared to the quantized
model with a skip connection, the learned quantization bitwidths are higher for the quantized
model without skip connection. Figure \ref{skip impact} presents that the quantization errors
of the model with skip connection are always higher than the model without skip connection
in every layer which means that the model without skip connection is more sensitive to 
the quantization error. Therefore, a higher quantization bitwidth is necessary for the model 
without skip connection to maintain the performance. We will add these analyses to the 
appendix in the revision.

\begin{table}
   \caption{The comparison results on other aggregation functions.}
      \label{other aggregation functions}
      \begin{center}
         \begin{tabular}{ccccc}
            \hline\toprule[2pt]
                      & Baseline(FP32) & Ours       & Bit  &Compression Ratio \\ \midrule[1pt]
            GIN\_sum  & 77.6±1.1\%     & 77.8±1.6\% & 2.37 &13.5x \\
            GIN\_mean & 78.8±0.1\%     & 78.5±0.6\% & 2.37 &13.5x  \\
            GIN\_max  & 78.6±1.6\%     & 78.6±0.5\% & 1.97 &16.2x  \\ \bottomrule[2pt]
         \end{tabular}
      \end{center}
\end{table}

\begin{figure}[t]
   \centering
   \includegraphics[scale=0.5]{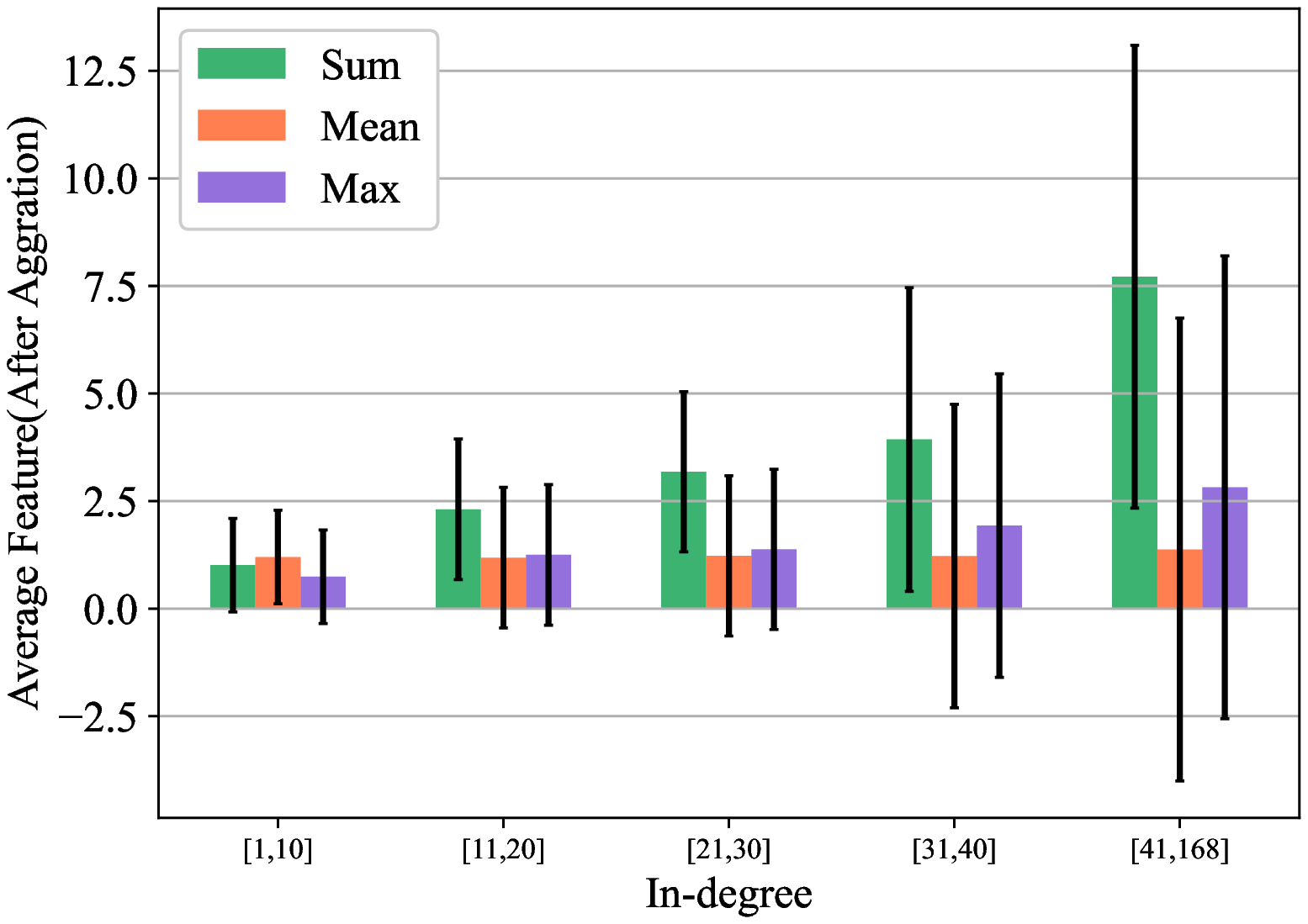}
   \caption{The average aggregated nodes features in different in-degree groups 
   for models with different aggregation functions.}
   \label{other aggregation functions}
\end{figure}

\textbf{Scalability for models that use other aggregation functions:}
To demonstrate that our method is also helpful to the GNNs using other aggregation functions
rather than the sum function, we replace the aggregation function of the GIN model, which is based on the MPNN framework 
with mean and max functions, and we conduct the comparison experiment on the Cora dataset.
As shown in Table \ref{other aggregation functions}, the accuracy degradation is negligible and 
the compression ratio is high 
, indicating that our 
quantization scheme also applies to the GNNs with
mean or max aggregation function. We analyze the average features for different aggregation 
functions in different in-degrees
group in Figure \ref{other aggregation functions}. The average features of the sum and max functions are 
highly dependent on
in-degrees. The other insight is that the variance of the features 
is also highly dependent on in-degrees.

The analysis demonstrates 
the generality of our approach, which can capture differences between nodes introduced 
by topology information of graphs
and compress the model size as much as possible while maintaining the performance.

\linespread{1.1}
\begin{table}[t]
   \caption{The comparison reults with the binary quantization method on Cora and CiteSeer datasets.}
   \label{binary_results}
   \begin{center}
         \begin{tabular}{cllll}
         \toprule[2pt]
         \multicolumn{1}{l}{}      &           & Accuracy            & Average bits & Compression ratio \\ \midrule[1pt]
         \multirow{9}{*}{Cora}     & GCN(FP32) & 81.5±0.7\%          & 32           & 1x                \\
                                 & Bi-GCN    & 81.2±0.8\%          & 1            & 32x               \\
                                 & GCN(ours) & \textbf{81.4±0.7}\%          & 1.61         & 19.9x             \\ \cline{2-5} 
                                 & GIN(FP32) & 77.6±1.1\%         & 32           & 1x                \\
                                 & Bi-GIN    & 33.7±6.6\%          & 1            & 32x               \\
                                 & GIN(ours) & \textbf{77.4±0.8\%} & 1.92         & 16.7x             \\ \cline{2-5} 
                                 & GAT(FP32) & 83.1±0.4\%         & 32           & 1x                \\
                                 & Bi-GAT    & 31.9±0\%            & 1            & 32x               \\
                                 & GAT(ours) & \textbf{82.6±0.5\%} & 2.03         & 15.8x             \\ \hline
         \multirow{9}{*}{CiteSeer} & GCN(FP32) & 71.1±0.7\%         & 32           & 1x                \\
                                 & Bi-GCN    & 70.7±2.4\%          & 1            & 32x               \\
                                 & GCN(ours) & 70.7±0.7\%          & 1.98         & 16.2x             \\ \cline{2-5} 
                                 & GIN(FP32) & 66.1±0.9\%         & 32           & 1x                \\
                                 & Bi-GIN    & 29.1±1.7\%          & 1            & 32x               \\
                                 & GIN(ours) & \textbf{65.6±1.5\%} & 2.39         & 13.4x             \\ \cline{2-5} 
                                 & GAT(FP32) & 72.5±0.7\%          & 32           & 1x                \\
                                 & Bi-GAT    & 20.6±2.6\%          & 1            & 32x               \\
                                 & GAT(ours) & \textbf{71.0±0.7\%} & 2.15         & 14.9x             \\ \bottomrule[2pt]
         \end{tabular}
   \end{center}
\end{table}
\subsubsection{Comparison With Binary Quantization Method}
\label{comp_binary}
In this section, we show the advantages of our method over the binary quantization method for GNNs. 
We select the binary quantization method in \citet{wang2021bi}
as our baseline. We just ran the experiments on the node-level because the binary quantization 
method only supports node-level tasks, which is one of
the drawbacks of the binary quantization method in GNNs. We quantize the same part 
as \citet{wang2021bi} does for a fair comparison.

The comparison results are shown in 
Table \ref{binary_results}. The binary quantization method performs 
well on GCN, where the aggregation and update phases are simple. However, on both models, 
GAT and GIN, the accuracy drops significantly compared with the FP32 baseline, 
which makes the deployment unrealistic. However, our method is
immune to this problem, although it has to use a higher average bit for node features which 
we believe is necessary for GAT and GIN. In summary, our method 
outperforms the binary quantization method in two ways: 

\textbf{1.} Our method can quantize more complex GNN models and ensure the accuracy degradation is 
negligible 
compared with the FP32 baseline while achieving a high compression ratio of 13.4x-19.9x. 

\textbf{2.} Our method can be applied to graph-level tasks. However, the binary quantization 
method can not handle them.
\begin{figure}[t]
   \centering
   \subfigure[]{
      \centering
      \begin{minipage}[b]{0.5\textwidth}
         \includegraphics[scale=0.38]{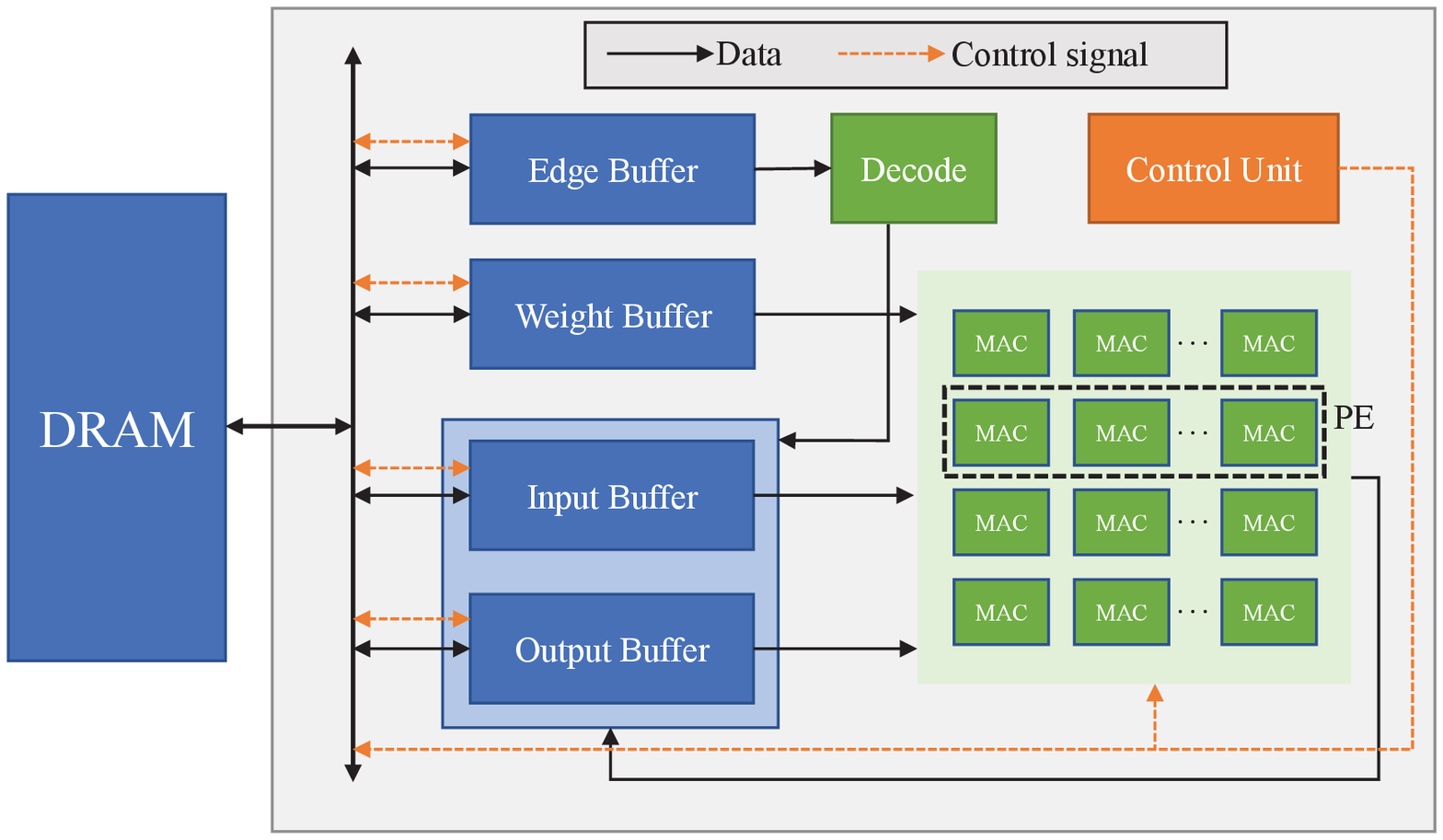}
         % \caption{GIN-CiteSeer}
         \label{archi_overview}
      \end{minipage}
   }
   \subfigure[]{
      \begin{minipage}[b]{0.4\textwidth}
         \centering
         \includegraphics[scale=0.24]{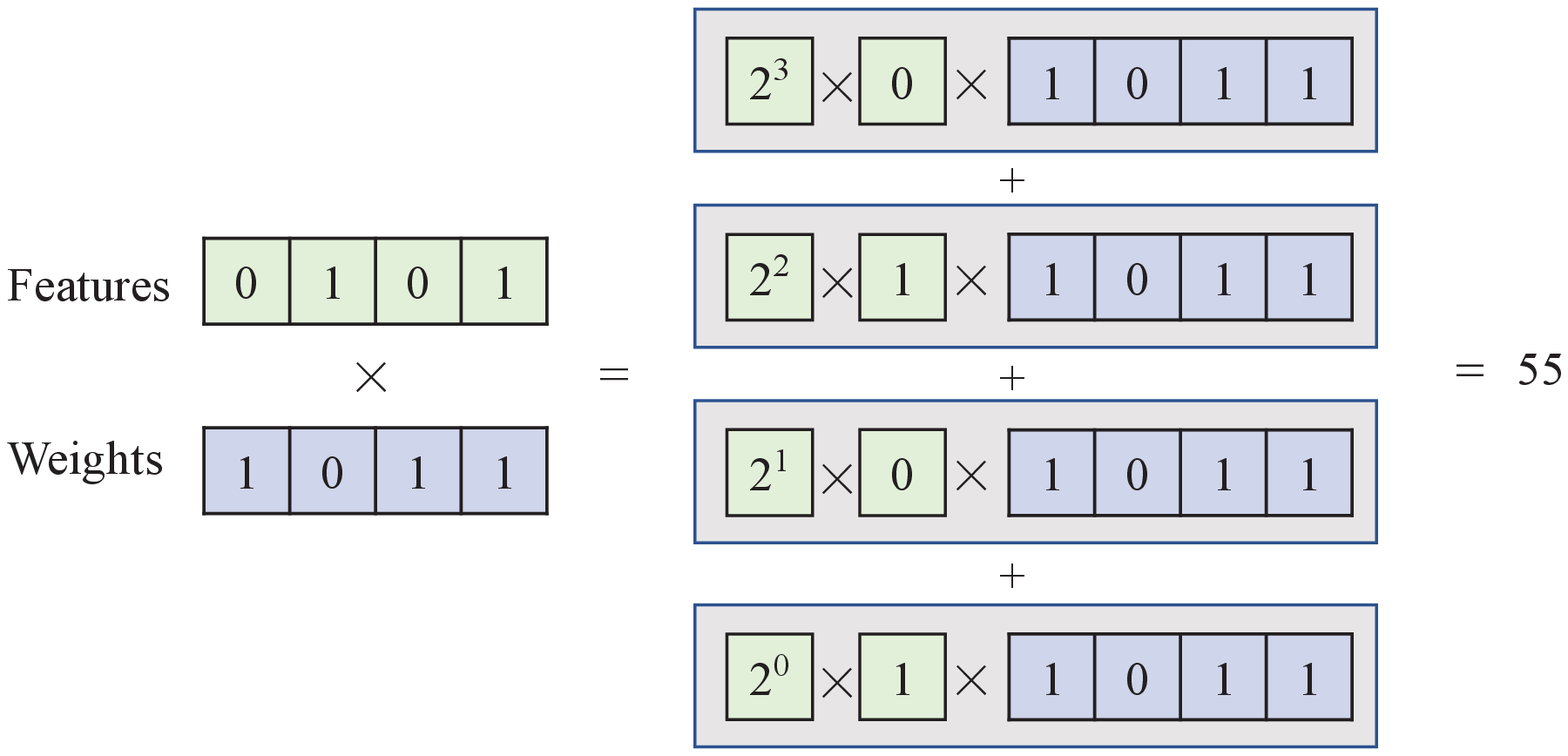}\\
         \includegraphics[scale=0.3]{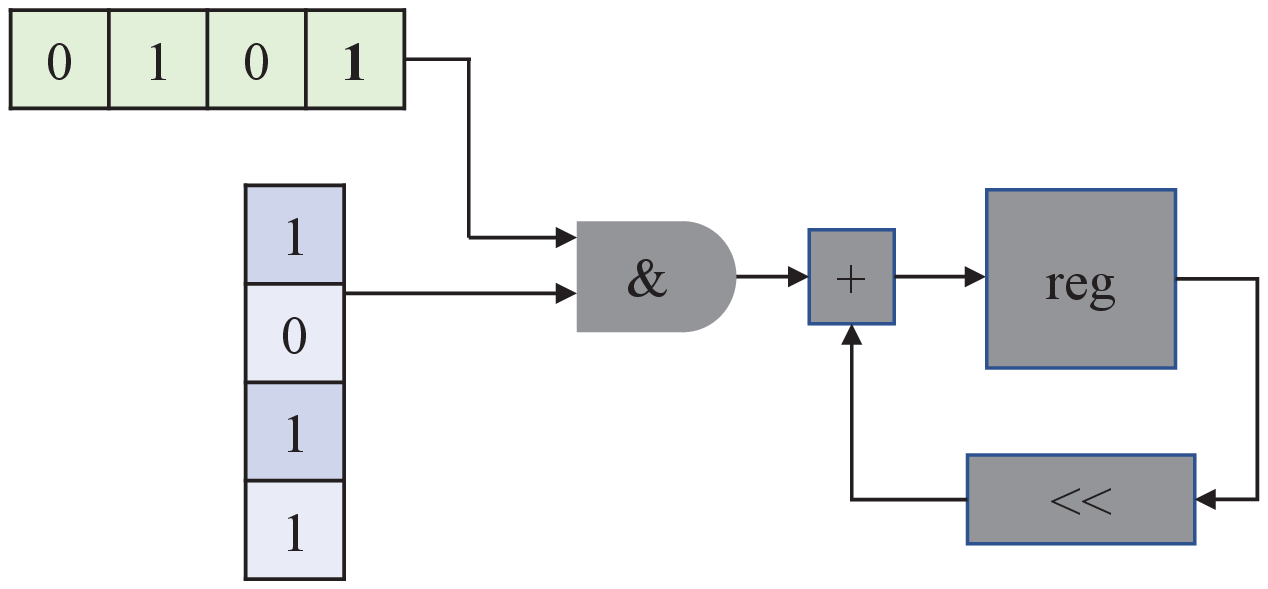}
         % \caption{GCN-CiteSeer}
         \label{bit-serial}
      \end{minipage}
   }
   \caption{(a) The overview of our accelerator architecture. (b) An 
   example of the bit-serial calculation and the architecture of the MAC.}
\end{figure}
\subsubsection{Accelerator Architecture}
\label{architecture}
In this section, we introduce the architecture of our 
hardware accelerator designed for GNN inference.
As presented in Section \ref{aaq}, we quantize each node feature to an 
appropriate precision and fix the weights to 4bits. 
To support mixed-precision computation, we adopt bit-serial multipliers at the core. 
Specifically, we follow the methodology in \cite{judd2016stripes} to only serialize the 
node features. This way, it takes $m$ cycles to complete the 
multiplication between an $m$-bit node feature with a 4bit weight, 
as shown in Figure \ref{bit-serial}. The product
involving $2^n$ is implemented by left-shift, i.e., for $2^n\times a$, 
we can shift $a$ left by $n$ bits to
implement the product.
To increase the computational throughput, we use $256\times 16$ MACs 
which can process 256 16-dimensional features in parallel. 
As shown in Figure \ref{archi_overview}, 
the compute unit is composed of 256 Processing Engines (PEs), each containing a row of 16 MACs.
The architecture of the MAC is shown in Figure \ref{bit-serial}.

The on-chip memory consists of an Edge Buffer, which stores the adjacency 
matrix of graphs, a Weight Buffer, which stores the weight of the GNNs, 
an Input Buffer, and an Output Buffer
to store the input features and the output result, and the register 
of each MAC to store the partial sum. 
To reduce data movement in the memory hierarchy, the 
input buffer and output buffer work in a swapped fashion, 
as the output of the current layer is the input to the next layer.
We set the memory size of Input Buffer, Output Buffer, Edge Buffer, 
and the Weight Buffer 
to 2MB, 2MB, 256KB, and 256KB, respectively. 
The overview of our architecture is shown in Figure \ref{archi_overview}.

To calculate $\mB^{l}=\mX^{l}\mW^{l}$, 256 consecutive rows in $\mX^l$ and a column of $\mW^l$ 
are mapped 
onto the MAC array to compute 256 inner products in each phase. 
To achieve this, a column of $\mW^l$ is broadcast and shared among PEs. 
The results of the inner products are written to the output buffer, which can be reused to reduce
the off-chip DRAM access.
The calculation of $\mX^{l+1}=\mA\mB^{l}$ is also in a inner-product manner. 
In this scenario, $\mA$ is a sparse matrix. We therefore represent 
$\mA$ in the Compressed Sparse Row (CSR) format, where full zero rows or elements of $\mA$ 
are eliminated. 
During inference, consecutive compressed rows of $\mA$ and a column of $\mB^l$ are mapped onto 
the MAC array in each phase.
We also sort the nodes in descending order according to their in-degrees, 
and the nodes with similar in-degrees are processed in 
parallel simultaneously to alleviate the load imbalance problem when performing the 
aggregation operations.

\subsubsection{Energy Efficiency Analysis}

\begin{figure}[ht]
   \centering
      \begin{minipage}[t]{0.45\textwidth}
         \vspace{-4cm}
         \centering
         \includegraphics[scale=0.5]{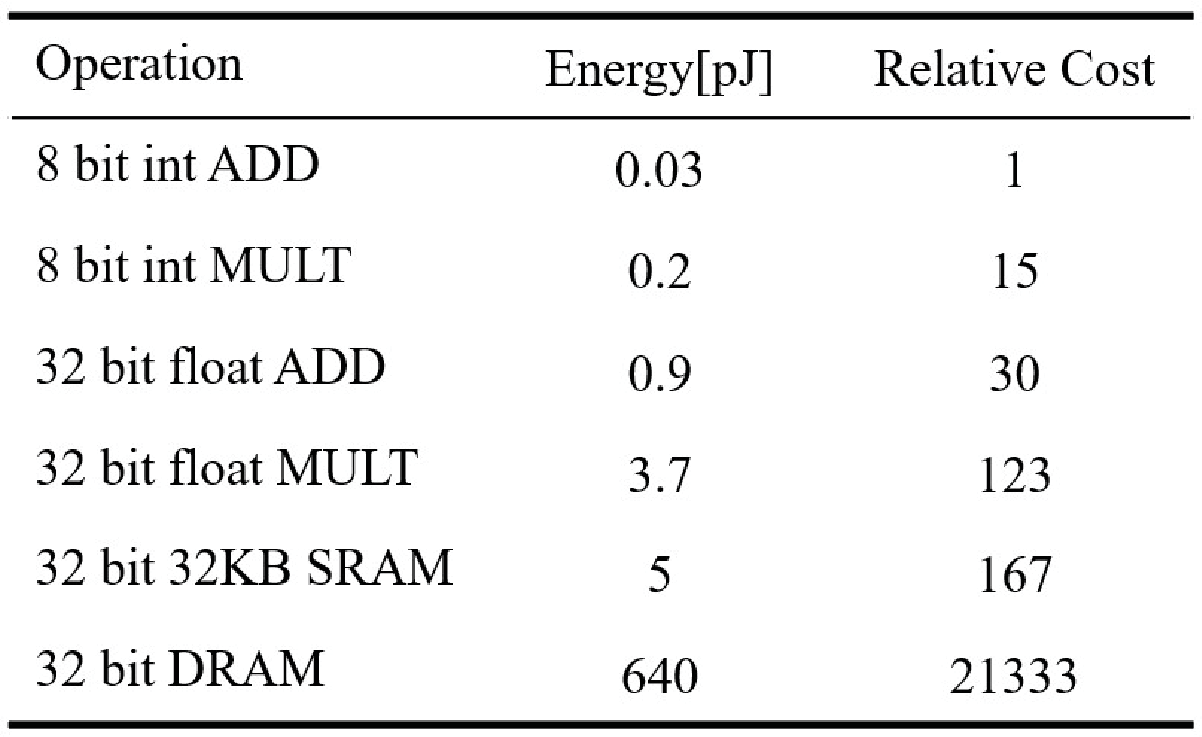}
         \caption[]{The energy table for 45nm technology node\citep{han2016eie,sze2020efficient}.}
         \label{energy table}
      \end{minipage} \ \
      \begin{minipage}[t]{0.45\textwidth}
         \centering
         \includegraphics[scale=0.35]{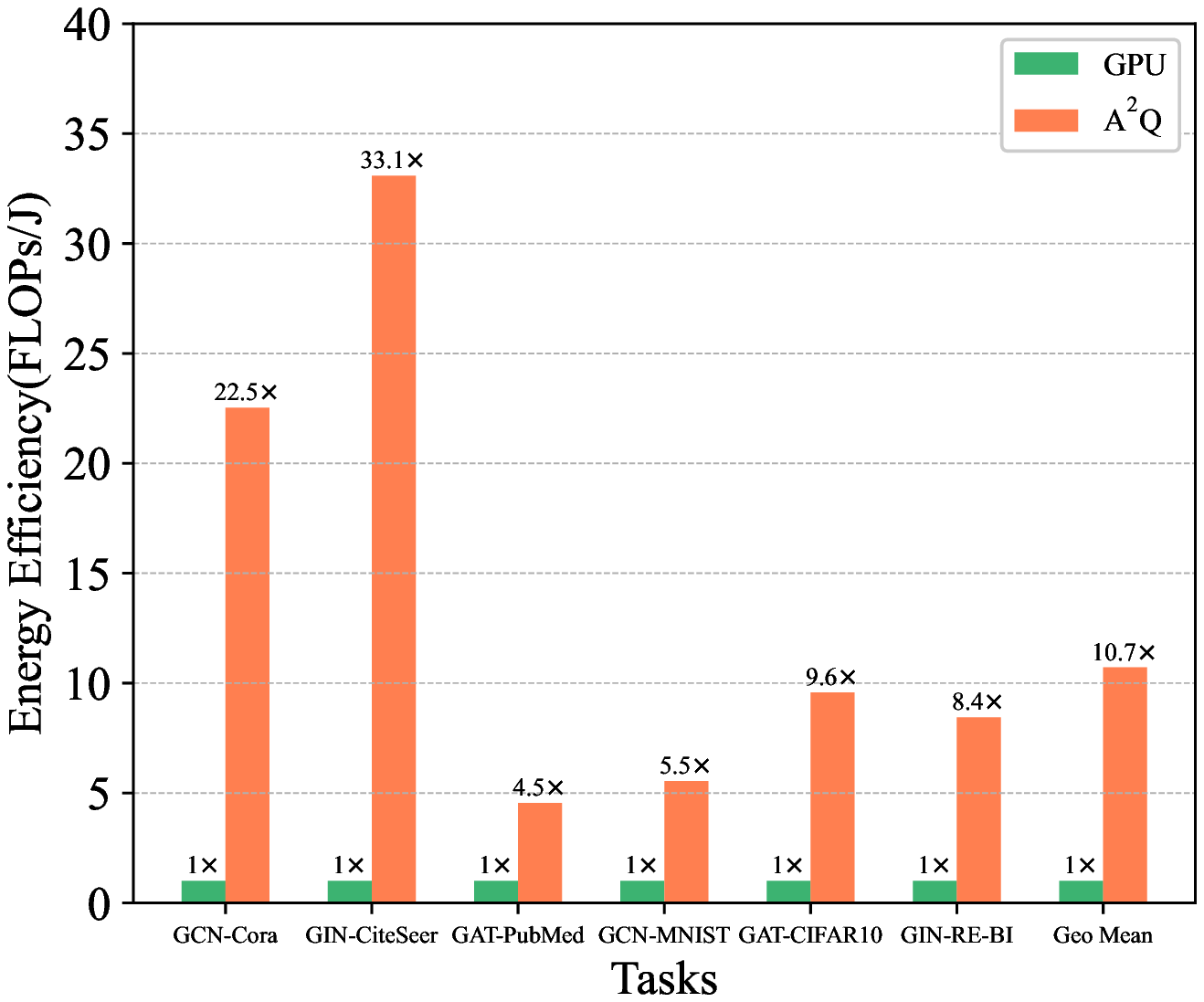}
         \vspace{-0.2cm}
         \caption[]{The energy efficiency compared with 2080Ti GPU on various tasks.}
         \label{energy efficiency}
      \end{minipage}
\end{figure}

Our method can save energy cost significantly from the following two aspects: 

1. By compressing the 
model size as much as possible, e.g., 18.6x compression ratio on GCN-Cora as 
shown in Table \ref{node-level-results}, 
our method can significantly reduce the memory footprints. 
Figure \ref{energy table} presents the energy table for the 45nm technology node.
It shows that memory access consumes further more energy than
arithmetic operations. Therefore, the memory footprints domains the energy cost, and then 
compressing 
the model can save much energy cost. 

2. Through our quantization method and the accelerator, the model can perform inference 
using the fixed-point 
operations instead of float-point operations, which are much more energy-consuming than  
fixed-point operations. As shown in Figure \ref{energy table},
the 32bit float MULT consumes 18.5x energy compared to the 8bit int MULT.
Therefore, our method's energy consumption is much lower than the FP32 model.

To illustrate the advantage of our approach in terms of energy efficiency, we compare our
accelerator with the 2080Ti GPU on various tasks. To estimate the 
energy efficiency of GPU, we use the \textbf{nvidia-smi} to obtain the power of GPU 
when performing the inference and 
measure the inference time by \textbf{time} function provided by Python. Then we can get the energy cost of
GPU. We also model the energy cost of our method on the accelerator.
We use High Bandwidth Memory (HBM) as our off-chip storage. 
Then we count the number of integer operations, 
and floating point operations, and the 
number of accesses to SRAM and HBM when performing the inference process of the quantized models 
on our accelerator. 
Based on the data in Table \ref{energy table}, we 
estimate the energy consumed by fixed-point
operations and floating-point operations. 
The static and dynamic power of SRAM is estimated using CACTI 7.0\citep{balasubramonian2017cacti}.
The energy
of HBM 1.0 is estimated with 7 pJ/bit as in \citep{o2014highlights}.
Figure \ref{energy efficiency}
presents these results, which 
shows that 
the 
the energy efficiency of our method is significantly better than GPU.

\subsection{Complexity Analysis}
In this section, we provide the analysis of the complexity of our proposed $\rm A^2Q$ method, including the 
computational complexity and space complexity.

\textbf{Space Complexity:} When analyzing the space complexity, we use the data size of the 
node features as an 
approximation of the entire loaded data, including weights and features, 
because the node features
account for more than 90\% of the overall memory consumption for a GNN model.
For a GNN has L layers, we assume that the input data to the first layer is 
$\mX\in \mathbb{R}^{N\times F_0}$, and the dimension of the hidden features is $F_1$. Then the dimension 
of the input to the 
2-(L-1) layer is $N\times F_1$. After quantizing the model, the average bits of the feature maps are $b_m$. 
The memory size includes two parts: 1. the nodes features $b_m[NF_0+(L-1)NF_1]$. 
2. the quantization step size (a step size is a float-point number which is 32bit) for each node $32NL$.
Therefore, the space complexity of the overall GNN model is as follows:
\begin{equation}
   \label{memory size}
   M=b_m[NF_0+(L-1)NF_1]+32NL \text{.}
\end{equation}
We can also obtain the ratio of the memory consumption of the step size in overall memory size:
\begin{equation}
   \label{memory ratio}
   r=\frac{32NL}{b_m[NF_0+(L-1)NF_1]} \text{.}
   % =\frac{32L}{b_m[F_0+(L-1)F_1]} \text{.}
\end{equation}
In the node-level tasks, the $F_0$ is usually much larger than 32, e.g., 3703 in the CiteSeer dataset.
Moreover, in the graph-level tasks, we usually
set $m=1000$, which is much smaller than the number of the input nodes to models, i.e., N. 
Therefore, although our method learns the quantization step size for each node 
the
memory overhead introduced by the quantization step size is negligible.

\textbf{Computational Complexity:} The forward pass is divided into the aggregation and update phases according 
to the MPNN framework. The aggregation phase can be represented as $\mH^{l}=\hat{\mA}\mX^{l}$, 
and then the update phase calculates $\mX^{l+1}=\mH^{l}\mW^{l}$. For $\hat{\mA}\in \mathbb{R}^{N\times N}$,
$\mX^{l}\in\mathbb{R}^{N\times F_1}$, and $\mW^{l}\in \mathbb{R}^{F_1\times F_2}$, the 
computational complexity
of the FP32 models is $\mathcal{O}(N^2F_1+NF_1F_2)$, which are all the float-point operations. After quantizing 
the model, the float-point matrix multiplication can be replaced by integer multiplication, and the element-wise 
operation, which calculates the multiplication between integers and float-point numbers according to
the Eq. \ref{mat_mul}. Then the computational
complexity is 
\begin{equation}
   C=\mathcal{O}_{I}(N^2F_1+NF_1F_2)+\mathcal{O}_{E}(NF_2) \text{,}
\end{equation} 
where $\mathcal{O}_{I}$ represents the complexity of the integers multiplication, whose cost is much lower
than the float-point operations, and the $\mathcal{O}_{E}$ represents the complexity of the element-wise 
operations.  
Note that although we quantize each node features by different step size, the complexity of 
element-wise operation involving
float-point is the same as the DQ-INT4 because the number of element-wise operations is equal to the number of the 
elements in a feature map, i.e., $N\times F_2$. 

\end{document}